\documentclass[journal]{IEEEtran}
\usepackage{times}
\usepackage[pdftex]{graphicx}
\usepackage{subfigure}
\usepackage{amsmath,amssymb,amsopn,amstext,amsfonts}
\usepackage{cancel}
\usepackage[space]{cite}
\usepackage{pdfsync}
\usepackage{balance}
\usepackage{color}
\usepackage{mathtools}
\usepackage{algpseudocode}
\usepackage{algorithm} 
\usepackage{bm}

\usepackage{diagbox}
\usepackage{float}
\usepackage{epstopdf}
\usepackage{pifont}
\usepackage{fixltx2e}
\usepackage{amsmath}
\usepackage{multirow}
\usepackage{url}
\usepackage{verbatim}
\usepackage[linkcolor=black,citecolor=black,urlcolor=black,colorlinks=true]{hyperref}

\algnewcommand\algorithmicforeach{\textbf{for each}}
\algnewcommand{\algorithmicgoto}{\textbf{go to}}%
\algnewcommand{\Goto}[1]{\algorithmicgoto~\ref{#1}}%
\algdef{S}[FOR]{ForEach}[1]{\algorithmicforeach\ #1\ \algorithmicdo}

\DeclareGraphicsExtensions{.png,.jpg,.eps,.pdf}

\begin{document}
\title{Teach-Repeat-Replan: \\ A Complete and Robust System for Aggressive Flight in Complex Environments}
\author{Fei~Gao, Luqi~Wang$^*$, Boyu~Zhou$^*$, Luxin~Han, Jie~Pan, 
         and~Shaojie~Shen
\thanks{All authors are with the Department of Electronic and Computer Engineering, Hong Kong University of Science and Technology, Hong Kong, China. {\tt\small $\{$fgaoaa, lwangax, bzhouai, luxin.han, jpanaa, eeshaojie$\}$@ust.hk}.
}
\thanks{$^*$These authors contributed equally to this work}}
\maketitle

\begin{abstract}
In this paper, we propose a complete and robust motion planning system for the aggressive flight of autonomous quadrotors. 
The proposed method is built upon on a classical teach-and-repeat framework, which is widely adopted in infrastructure inspection, aerial transportation, and search-and-rescue. 
For these applications, human's intention is essential to decide the topological structure of the flight trajectory of the drone. 
However, poor teaching trajectories and changing environments prevent a simple teach-and-repeat system from being applied flexibly and robustly. 
In this paper, instead of commanding the drone to precisely follow a teaching trajectory, we propose a method to automatically convert a human-piloted trajectory, which can be arbitrarily jerky, to a topologically equivalent one.
The generated trajectory is guaranteed to be smooth, safe, and kinodynamically feasible, with a human preferable aggressiveness. 
Also, to avoid unmapped or dynamic obstacles during flights, a sliding-windowed local perception and re-planning method are introduced to our system, to generate safe local trajectories onboard.
We name our system as \textit{teach-repeat-replan}. 
It can capture users' intention of a flight mission, convert an arbitrarily jerky teaching path to a smooth repeating trajectory, and generate safe local re-plans to avoid unmapped or moving obstacles.
The proposed planning system is integrated into a complete autonomous quadrotor with global and local perception and localization sub-modules. 
Our system is validated by performing aggressive flights in challenging indoor/outdoor environments.
We release all components in our quadrotor system as open-source ros-packages\footnote{\url{https://github.com/HKUST-Aerial-Robotics/Teach-Repeat-Replan}}. 
\end{abstract}

\IEEEpeerreviewmaketitle

\section{Introduction}
\IEEEPARstart{A}{s} the development of autonomy in aerial robots, Micro Aerial Vehicle (MAV) has been more and more involved in our daily life. 
Among all applications emerged in recent years, quadrotor teach-and-repeat has shown significant potentials in aerial videography, industrial inspection, and human-robot interaction. 
In this paper, we investigate and answer the problem of what is the best way to incorporate a human's intention in autonomous and aggressive flight, and what is a flexible, robust and complete aerial teach-and-repeat system.

\begin{figure}[t]
\begin{center}          
\subfigure[\label{fig:indoor_snapshot} Snapshot of the indoor quadrotor flight.]
{\includegraphics[width=0.92\columnwidth]{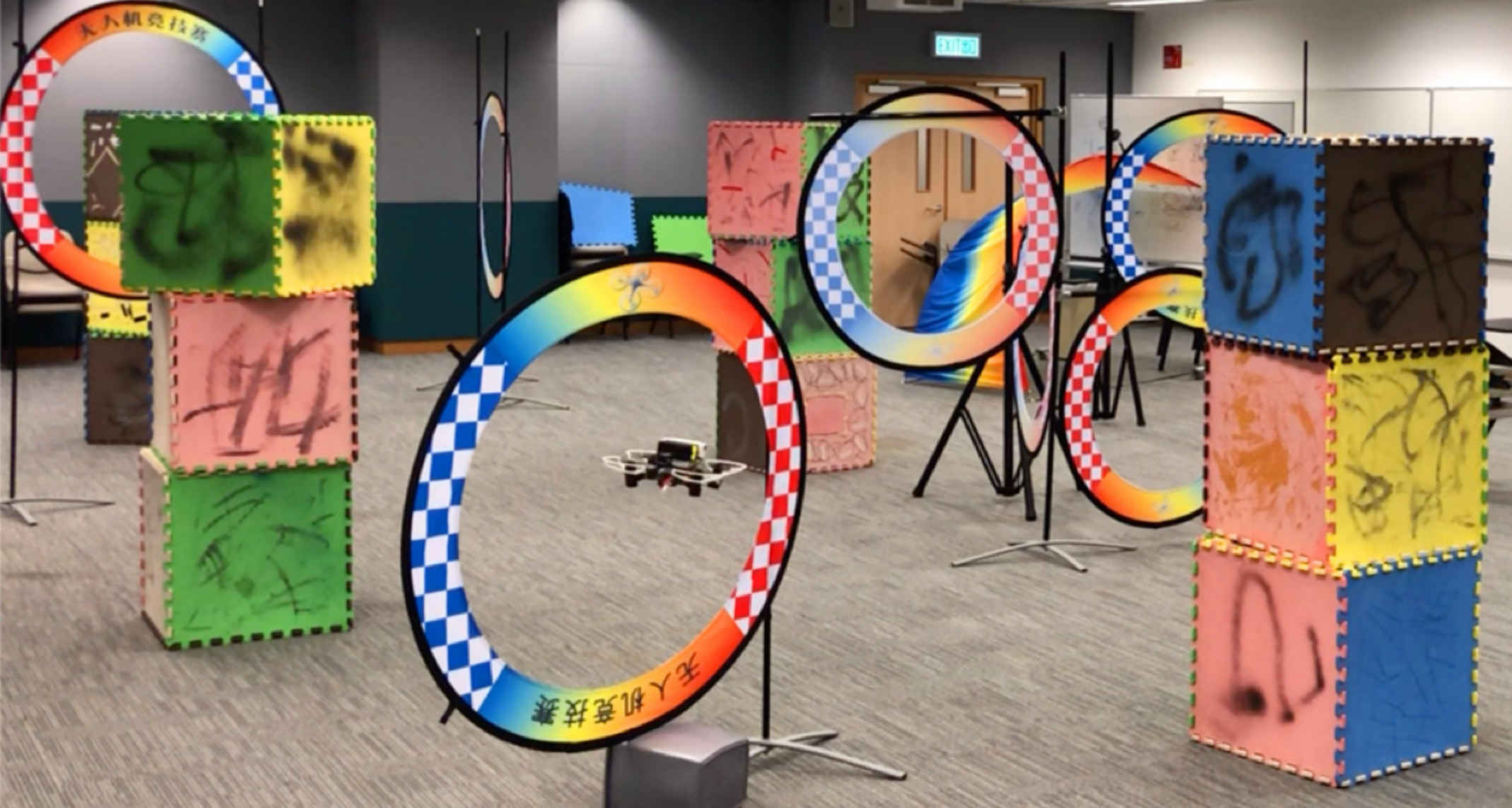}}       
\subfigure[\label{fig:outdoor_snapshot} Snapshot of the outdoor quadrotor flight.]
{\includegraphics[width=0.92\columnwidth]{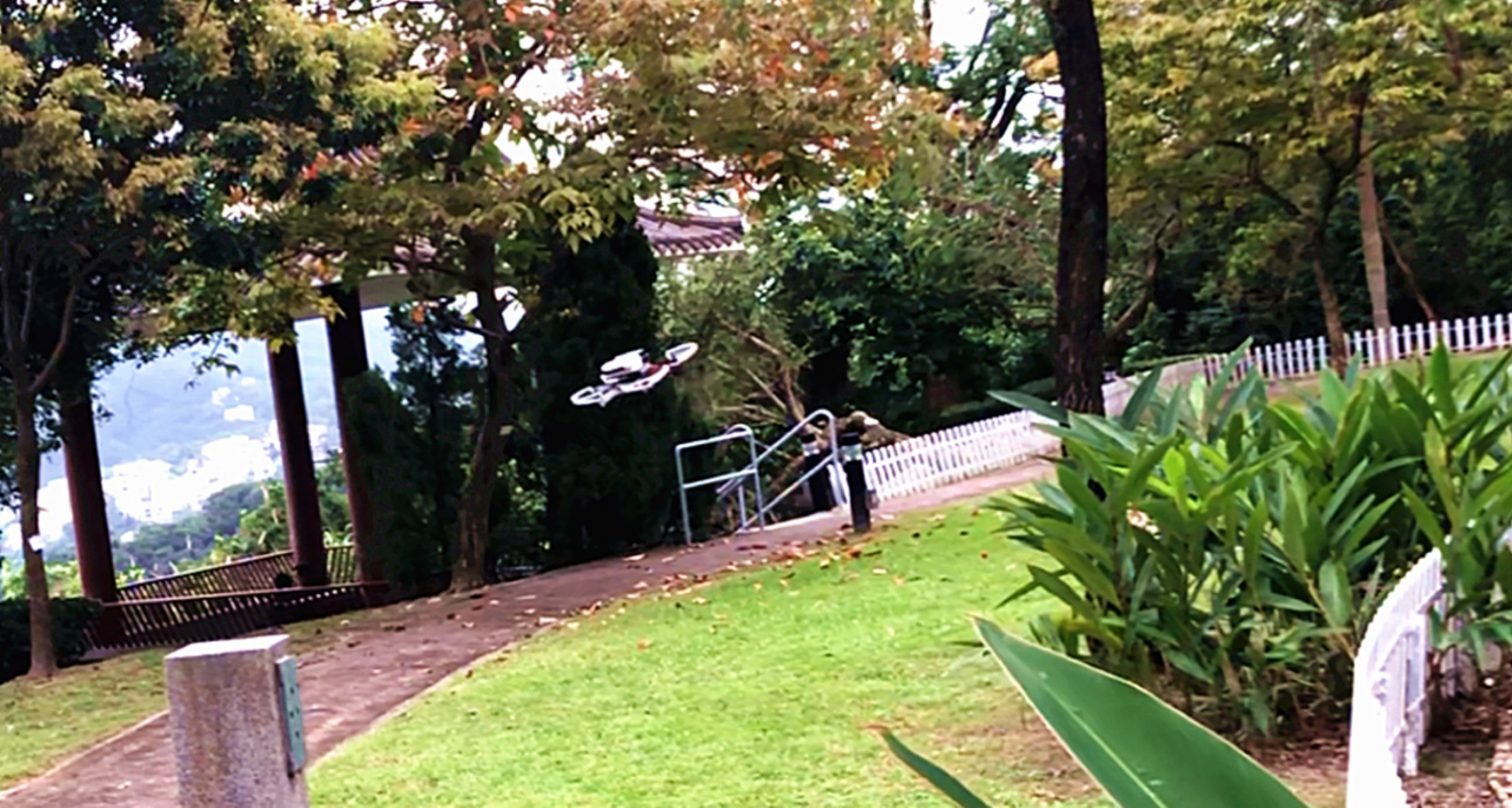}}     
\end{center}
\vspace{-0.5cm}
\caption{\label{fig:exp_snapshot} Experiments in a challenging indoor drone racing site and an outdoor forest. A high-resolution video is available at: \url{https://youtu.be/urEC2AAGEDs}  }
\vspace{-0.75cm}
\end{figure}  

There is a massive market for consumer drones nowadays. 
However, we observe that most of the operators of consumer drones are not professional pilots and would struggle in generating their ideal trajectory for a long time. 
In some scenarios, such as the drone racing or aerial filming, a beginner-level pilot is impossible to control the drone to finish the race safely or take an aerial video smoothly unless months of training. 
Also, there is considerable demand in applying drones to repetitive industrial inspections or search-and-rescue missions, where human provides a preferable routine. 
In these situations, demonstrating a desirable trajectory and letting the drone to repeat it is a common wish.
However, the taught trajectory generated by an unskilled pilot is usually incredibly hard or dynamically infeasible to repeat, especially in some cluttered environments. 
Moreover, most of the vision-based teach-and-repeat applications~\cite{fei2019ral},~\cite{fehr2018visual},~\cite{furgale2010visual}, such as our previous work~\cite{fei2019ral}, are sensitive to changing environments. 
In~\cite{fei2019ral}, even the environment changes very slightly, the global map has to be rebuilt, and the teaching has to be redone.

Based on these observations, in this paper, instead of asking the drone to follow the human-piloted trajectory exactly, we only require the human operator to provide a rough trajectory with an expected topological structure. 
Such a human's teaching trajectory can be arbitrarily slow or jerky, but it captures the rough route the drone is expected to fly. 
Our system then autonomously converts this poor teaching trajectory to a topological equivalent and energy-time efficient one with an expected aggressiveness. 
Moreover, during the repeating flight, our system locally observes environmental changes and re-plans sliding-windowed safe trajectories to avoid unmapped or moving obstacles.
In this way, our system can deal with changing environments.
Our proposed system extends the classical robotics teach-and-repeat framework and is named as \textit{teach-repeat-replan}.
It is complete, flexible, and robust. 

In our proposed system, the surrounding environment is reconstructed by onboard sensors. 
Then the user's demonstrated trajectory is recorded by virtually controlling the drone in the map with a joystick or remote controller. 
Afterward, we find a flight corridor that preserves the topological structure of the teaching trajectory. 
The global planning is decoupled as spatial and temporal planning sub-problems. 
Having the flight corridor, an energy-optimal spatial trajectory which is guaranteed to be safe, and a time-optimal temporal trajectory which is guaranteed to be physically feasible, are iteratively generated. 
In repeating, while the quadrotor is tracking the global spatial-temporal trajectory, a computationally efficient local map~\cite{han2019fiesta} is fused onboard by stereo cameras.
Based on local observations, our proposed system uses a sliding-window fast re-planning method~\cite{boyu2019ral} to avoid possible collisions.
The re-planning module utilizes gradient information to locally wrap the global trajectory to generate safe and kinodynamic feasible local plans against unmapped or moving obstacles.

The concept of generating optimal topology-equivalent trajectories for quadrotor teach-and-repeat was first proposed in our previous research~\cite{fei2019ral}. 
In~\cite{fei2019ral}, once the repeating trajectory is generated, the drone executes it without any other considerations.
In that work~\cite{fei2019ral}, the environment must remain intact during the repeating, and the localization of the drone is assumed to be perfect. 
These requirements are certainly not guaranteed in practice, therefore, prevent the system from being applied widely.
In this paper, we extend the framework of the classical teach-and-repeat and propose several new contributions to make our system complete, robust, and flexible. 
Contributions of this paper are listed as:
\begin{enumerate}
\item We advance our flight corridor generation method. 
The flight corridor we use now provides much more optimization freedom compared to our previous work~\cite{fei2019ral}. 
The improvement of the flight corridor facilitates the generation of more efficient and smooth global trajectories.
Moreover, we propose methods to accelerate the corridor generation on both CPU and GPU.
\item We introduce our previous works on online mapping~\cite{han2019fiesta} and re-planning~\cite{boyu2019ral} into our system, to improve the robustness against errors of global maps, drifts of localization, and environmental changes and moving obstacles.
\item We present a whole set of experiments and comparisons in various scenarios to validate our system.
\item We release all components in the system as open-source packages, which include local/global planning, perception, and localization, and onboard controller.
\end{enumerate}

\begin{figure*}[t]
\centering
\includegraphics[width=1.8\columnwidth]{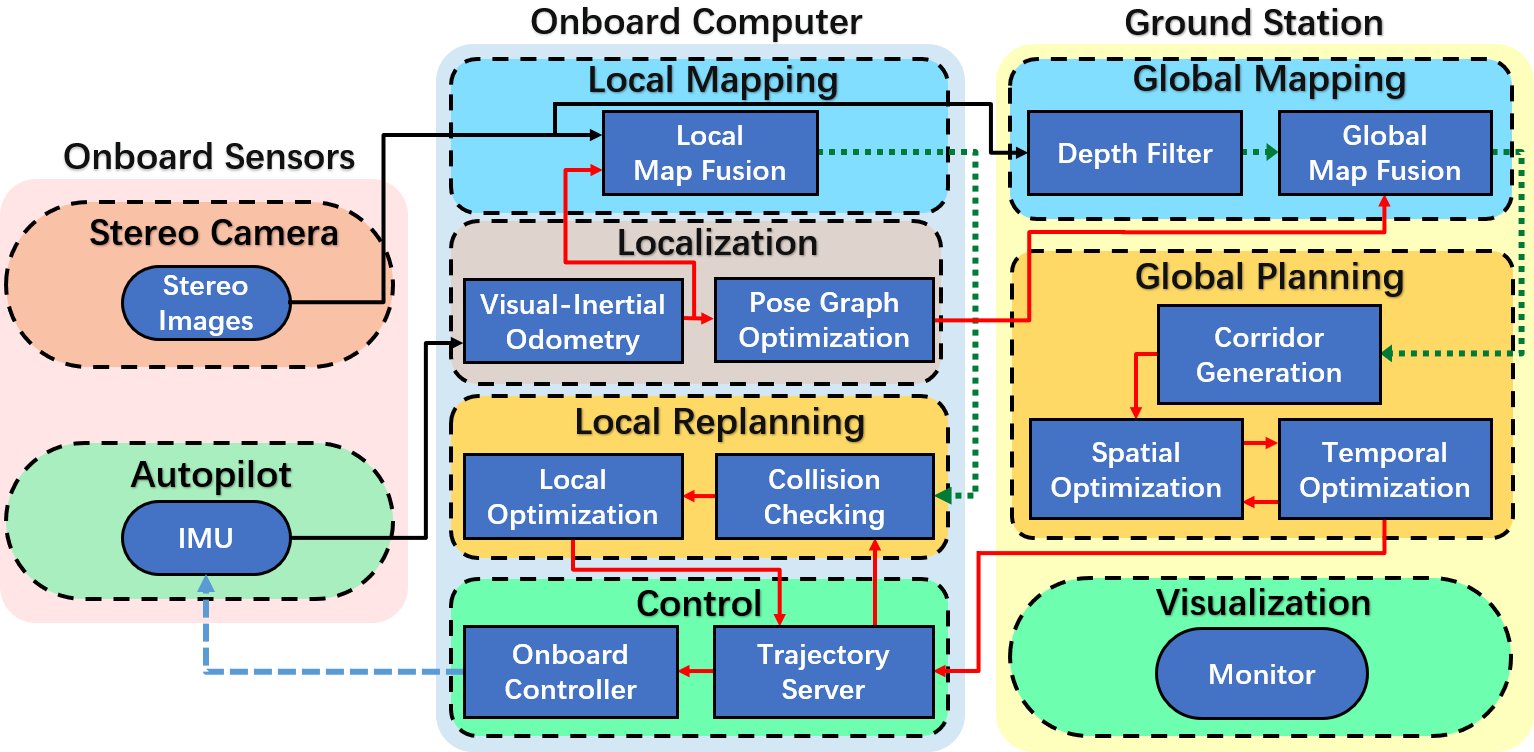}
\vspace{-0.25cm}
\caption{The software architecture of our quadrotor system. \label{fig:sys_architecture} 
Global mapping, planning, and visualization are running on a ground station, while state estimation, local sensing, and re-planning are running onboard. 
}
\vspace{-0.5cm}
\end{figure*}

\begin{figure}[t]
\centering
\includegraphics[width=0.9\columnwidth]{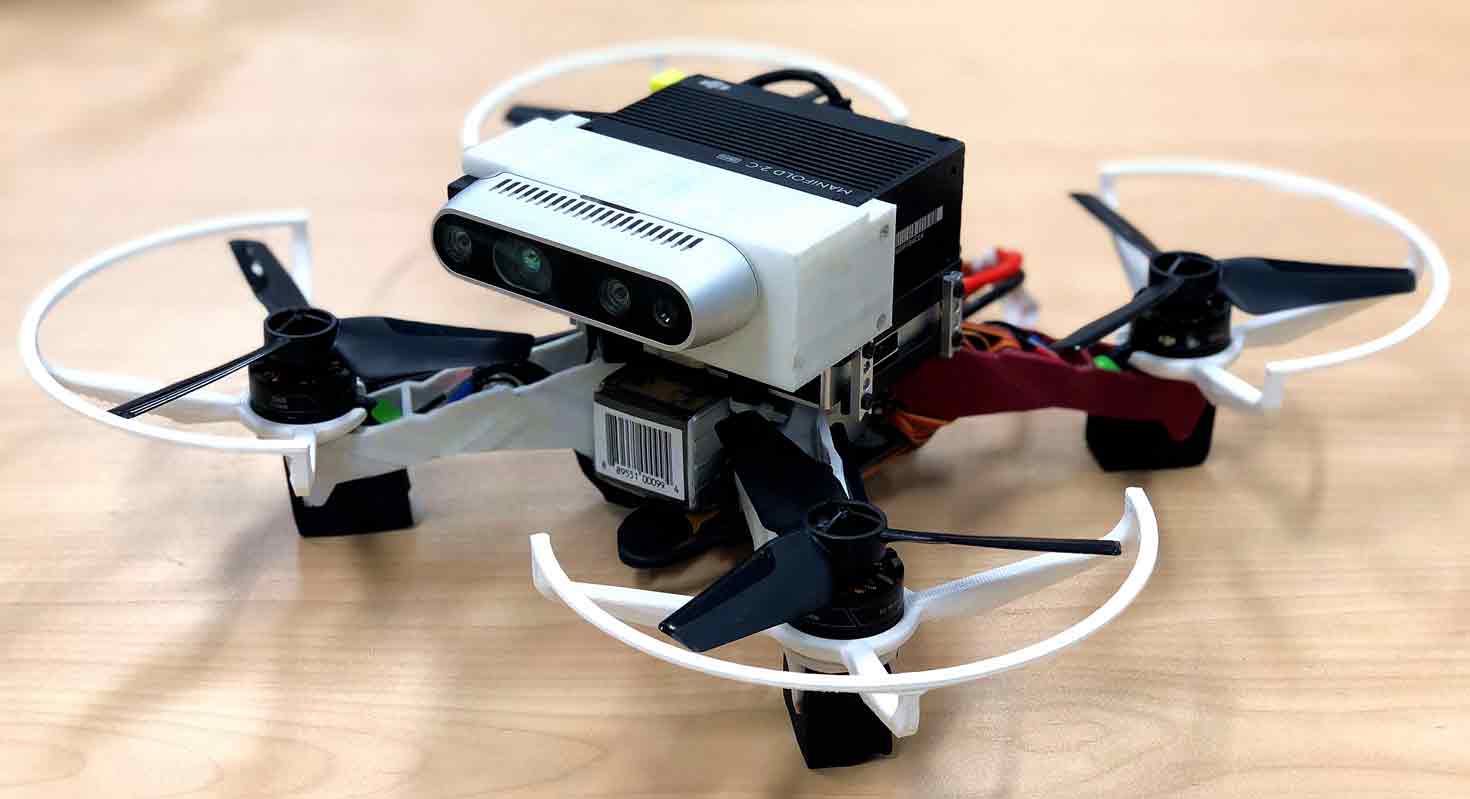}
\caption{The hardware setting of our autonomous drone system. 
\label{fig:sys_hardware} }
\vspace{-0.75cm}
\end{figure}
In what follows, we discuss related literature in Sect.~\ref{sec:related_work} and introduce our system in Sect.~\ref{sec:system_overview}. Our methods for finding a flight corridor consisting of large convex polyhedrons, and spatial-temporal trajectory optimization are detailed in Sect.~\ref{sec:corridor_generation} and Sect.~\ref{sec:trajectory_optimization}, respectively. The local re-planning is introduced in Sect.~\ref{sec:online_local_replanning}. Experimental and benchmarked results are given in Sect.~\ref{sec:results}. The paper is concluded in Sect.~\ref{sec:conclusion}.

\section{Related Works}
\label{sec:related_work}
\textbf{Robotics teach-and-repeat:}
Many robotics teach-and-repeat works, especially for mobile robots, have been published in recent years. 
Most of them focus on improving the accuracy or robustness in repeating/following the path by operators, which is fundamentally different from our motivation.
A lidar-based teach-and-repeat system is proposed in~\cite{sprunk2013lidar}, where laser scans are used to localize the ground vehicle against its taught path driven by the user.  
Furgale et al.~\cite{furgale2014there}~\cite{krusi2015lighting} also develop a lidar-based ground robot, which is specially designed for repeating long-term motions in highly dynamic environments. 
This system equips a local motion planner which samples local trajectory to avoid dynamic elements during route following. 
A map maintenance module is used to identify moving objects and estimate their velocities.
An iterative learning controller is proposed in~\cite{ostafew2013visual}, to reduce the tracking error during the repeating of the robot.
This controller can compensate disturbances such as unmodelled terrains and environmental changes by learning a feedforward control policy.
Vision-based teaching-and-repeat systems are also proposed in several works, such as the visual localization used by the rover in~\cite{furgale2010visual}. In this work, the authors build a manifold map during the teaching and then use it for localization in the repeating,
In~\cite{paton2016bridging}, a multi-experience localization algorithm is proposed to address the issue of environmental changes. 
The ground robot is localized robustly against several past experiences. 
In~\cite{paton2015s} and~\cite{berczi2016s}, to further improve the accuracy and robustness in localization, illumination and terrain appearances are considered in their proposed visual navigation system used for teach-and-repeat, 
Compared to ground teach-and-repeat works, research on aerial teach-and-repeat is few. 
In~\cite{fehr2018visual}, a vision-based drone is used to inspect infrastructure repetitively.
In the teaching phase, the desired trajectory is demonstrated by the operator, and some keyframes in the visual SLAM are recorded as checkpoints. 
While repeating, local trajectories are generated to connect those checkpoints by using the minimum-snap polynomials~\cite{MelKum1105}. 
To function properly, in this work, the teaching trajectory itself must be smooth, and the environments must have no changes during the whole repeating. 
In contrast, our proposed method can convert an arbitrarily poor path to a safe and efficient trajectory with expected flying aggressiveness. 
Also, our system is flexible. 
Since it records the teaching path by virtually controlling the drone in simulation, a manually piloted teaching process is not necessary.
Finally, our proposed system is robust to environmental changing and can even avoid moving obstacles.

\textbf{Quadrotor trajectory planning:}
Trajectory optimization is essential in generating safe and executable repeating trajectories from poor teaching ones. 
The minimum-snap trajectory optimization is proposed by Mellinger~\cite{MelKum1105}, where piecewise polynomials are used to represent the quadrotor trajectory and are optimized by quadratic programs (QP). 
A method for solving a closed-form solution of the minimum snap is proposed in~\cite{RicBryRoy1312}.
In this work, a safe geometric path is firstly found to guide the generation of the trajectory. 
By adding intermediate waypoints to the path iteratively, a safe trajectory is finally generated after solving the minimum-snap problem several times.
Our previous works~\cite{fei2018icra}~\cite{fei2016ssrr}~\cite{fei2018jfr} carve a flight corridor consisting of simple convex shapes (sphere, cube) in a complex environment. 
The flight corridor constructed by a series of axis-aligned cubes or spheres can be extracted very fast on occupancy map or Kd-tree. 
Then we use the flight corridor and physical limits to constrain a piecewise B\'ezier curve, to generate a guaranteed safe and kinodynamic feasible trajectory,
Other works are proposed to find general convex polyhedrons for constraining the trajectory. 
In~\cite{liu2017ral}, a piecewise linear path is used to guide and initialize the polyhedron generation. In~\cite{deits2015computing}, by assuming all obstacles are convex, SDP and QP are iteratively solved to find the maximum polyhedron seeded at a random coordinate in 3-D space.
Gradient information in maps is also valuable for local trajectory optimization.
In CHOMP~\cite{ratliff2009chomp}, the trajectory optimization problem is formulated as a nonlinear optimization over the penalty of safety and smoothness. 
In~\cite{oleynikova2016continuous},~\cite{lin2018autonomous} and~\cite{helen2019system}, gradient-based methods are combined with piecewise polynomials for local planning of quadrotors.
In this paper, we also utilize gradient-based optimization for local re-planning.

Time optimization or so-called time parametrization is used to optimize the time profile of a trajectory, given the physical limits of a robot. 
Methods can be divided as direct methods~\cite{choset2005principles} and indirect methods\cite{roberts2016generating}.
Direct methods generate an optimal spatial-temporal trajectory directly in the configuration space.
For indirect methods, a trajectory independent of time is firstly generated, the relationship between time and the trajectory is optimized by an additional optimization process. 
The method in~\cite{jamieson2016near} finds a mapping function between the time and the trajectory, which is done by recursively adding key points into the function, and squeeze out the infeasibility of the time profile. 
This method obtains an optimal local solution and is computationally expensive. 
\cite{roberts2016generating} also proposes a mapping function, which maps time to a virtual parametrization of the trajectory. 
The mapping function is then optimized under a complicated nonlinear formulation. 
However, the global optimality is not guaranteed, and a feasible initial solution is necessary to bootstrap the optimization. 
Convex optimization~\cite{verscheure2009time} and numerical integration~\cite{pham2014general} are two typical methods of robotics time optimal path parametrization (TOPP) problem. 
Although numerical integration~\cite{pham2014general}~\cite{pham2018new} has shown superior performance in computing efficiency, convex optimization~\cite{verscheure2009time} has the advantage of adding regularization terms other than total time into its objective function. 
This specialty suits well for our application where the user defines the expected aggressiveness of the drone, and sometimes the drone may not be expected to fly as fast as possible. 
As for the efficiency, since we do temporal optimization off-line before the repeating, computing time is not critical.

\section{System Overview}
\label{sec:system_overview}
\subsection{System Architecture}

The overall software and hardware architecture of our quadrotor system are shown in Fig.~\ref{fig:sys_architecture} and~\ref{fig:sys_hardware}. 
The global mapping, flight corridor generator, and global spatial-temporal planner are done on an off-board computer.  
Other online processings are running onboard on the drone during the flight. 
Before teaching, the global map is built by onboard sensors. 
During teaching, a flight corridor is generated by inflating the teaching trajectory. 
Then the spatial and temporal trajectories are optimized iteratively within the flight corridor under a coordinate descent scheme~\cite{wright2015coordinate}. 
The local planner using gradient-based optimization is running onboard to avoid unexpected obstacles observed in the repeating flights. 
For trajectory tracking, we use a geometric controller~\cite{lee2010}.
And the attitude is stabilized by the autopilot.

\subsection{Globally Consistent Localization and Mapping}
\label{subsec:localization_mapping}
We use VINS~\cite{qin2018vins}, a robust visual-inertial odometry (VIO) framework, to localize the drone. 
Moreover, the loop closure detection and global pose graph optimization are used in our system, to globally correct the pose estimation. 
The global mapping is done by fusing depth measurements from the stereo cameras with the pose estimation.
By using our previous research on deformable map~\cite{wang2019surfel}, our global mapping module maintains a series of sub-maps with each attached to a keyframe.
In this way, the map is attached to the pose graph and is therefore globally driftless. 
During the mapping, when a loop closure is detected, keyframes in the global pose graph are corrected, and all sub-maps are deformed accordingly. 
The global pose graph optimization is also activated during the repeating.
When loop closure is detected, the pose of the drone is corrected accordingly to eliminate the drift.

\subsection{Global Spatial-Temporal Planning}
\label{subsec:coordinate_descent}
For an extremely poor teaching trajectory, both the geometric shape and time profile of it is far from optimal and therefore useless, or even harmful for conducting optimization.
However, the topological information of the teaching trajectory is essential since it reflects the human's intention.
To preserve the topological information, we group the free space around the teaching trajectory to form a flight corridor (Sect.~\ref{sec:corridor_generation}). 
The corridor contains the teaching trajectory within it, shares the same topological structure, and provides large freedom for optimization.
It's hard to concurrently optimize a trajectory spatially and temporally in the flight corridor.
However, generating a safe spatial trajectory given a fixed time allocation (Sect.~\ref{subsec:space_optimization}) and optimizing the time profile of a fixed spatial trajectory (Sect.~\ref{subsec:time_optimization}) are both conquerable. 
Therefore, we iteratively optimize the trajectory in the space-time joint solution space by designing a coordinate descent~\cite{wright2015coordinate} framework. 
An objective with weighting energy and time duration is defined for optimization. 
We firstly generate a spatial trajectory whose energy is minimized, then we use the temporal optimization to obtain the optimal time profile of it. 
The optimal time profile is used to parametrize a trajectory again for spatial optimization. 
The spatial-temporal optimizations are done iteratively until the total cost cannot be reduced any more. 

\subsection{Local Collision Avoidance}
\label{subsec:local_collision_avoidance}
In practice, the accumulated drift of VIO is unavoidable, and the recall rate of loop closure is unstable. 
Although we have built a dense global map, when the drift is significant and not corrected by loop detection in time, the quadrotor may have collisions with obstacles. 
Moreover, the environment may change or contain moving obstacles.
Our previous work~\cite{fei2019ral} has to re-build the map when changes happen and can not deal with dynamic obstacles.
To resolve the above issues, we integrate our previous local map fusion module~\cite{han2019fiesta} into our system to detect collisions locally and serves the local trajectory optimization. 
Also, we propose a sliding-window local replanning method based on our previous research on quadrotor local planning~\cite{boyu2019ral}, to avoid collisions on the flight.

In the repeating phase, the drone controls its yaw angle to face its flying direction and build a local map by stereo cameras.
We consistently check the local trajectory within a replanning time horizon. 
If collisions along the local trajectory are reported, replanning is triggered to wrap the trajectory out of obstacles by gradient-based optimization~\cite{boyu2019ral}. 
\begin{figure}[t]
\centering
\includegraphics[width=0.8\columnwidth]{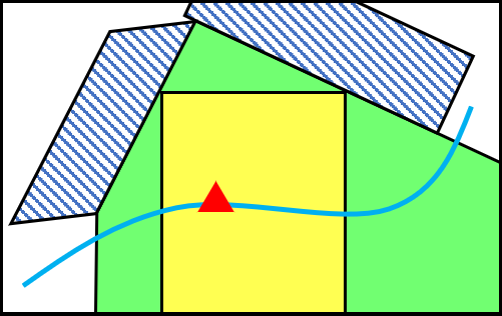}
\caption{An illustration of free space captured by an axis-aligned cube and a general polyhedron. Obstacles are shown in dashed lines. The blue curve is the teaching trajectory of humans. The red triangle is the seed for finding local free space. The axis-aligned cube and a corresponding general convex polyhedron are shown in yellow and green, respectively. \label{fig:cube_polygon_compare}}
\vspace{-0.5cm}
\end{figure}

\begin{figure}[t]
\begin{center}          
\subfigure[\label{fig:cube_rviz_1} Axis-aligned cube, side view]
{\includegraphics[height=0.38\columnwidth]{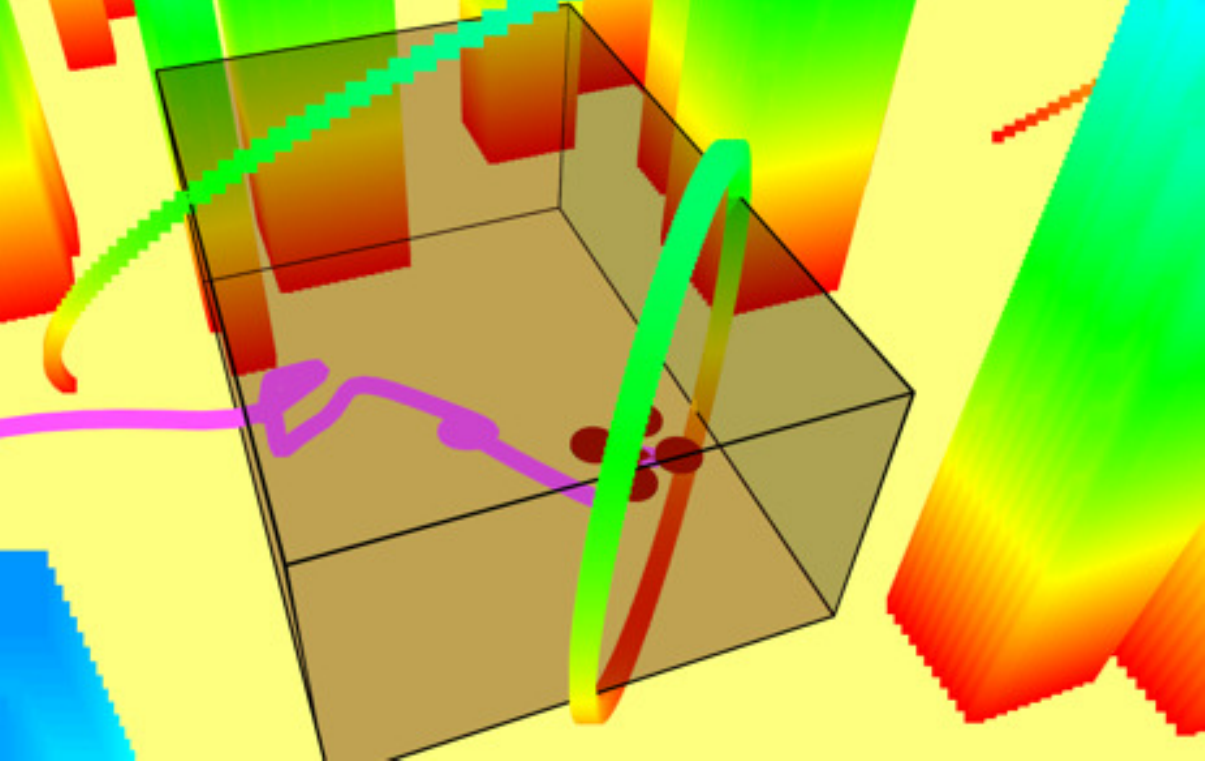}}       
\subfigure[\label{fig:cube_rviz_2} Front view]
{\includegraphics[height=0.38\columnwidth]{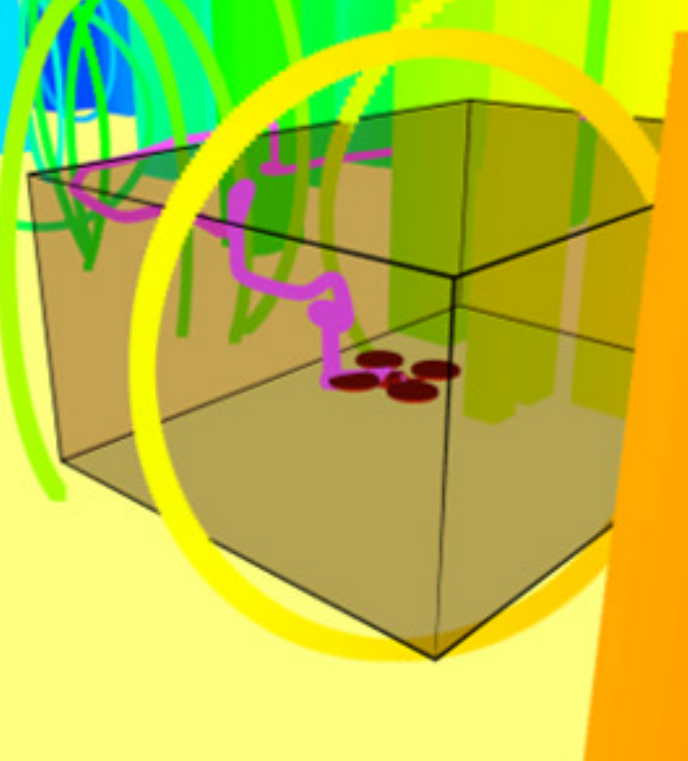}}     
\subfigure[\label{fig:poly_rviz_1} General convex polyhedron, side view]
{\includegraphics[height=0.38\columnwidth]{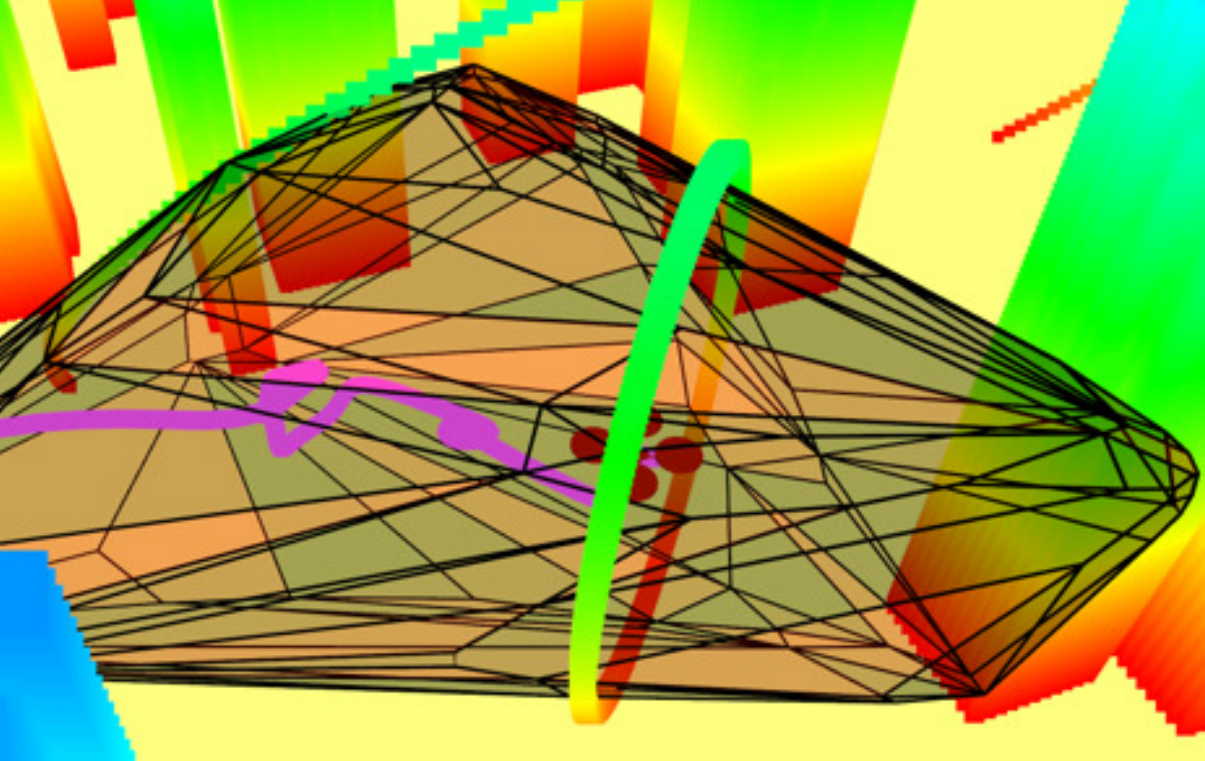}}       
\subfigure[\label{fig:poly_rviz_2} Front view]
{\includegraphics[height=0.38\columnwidth]{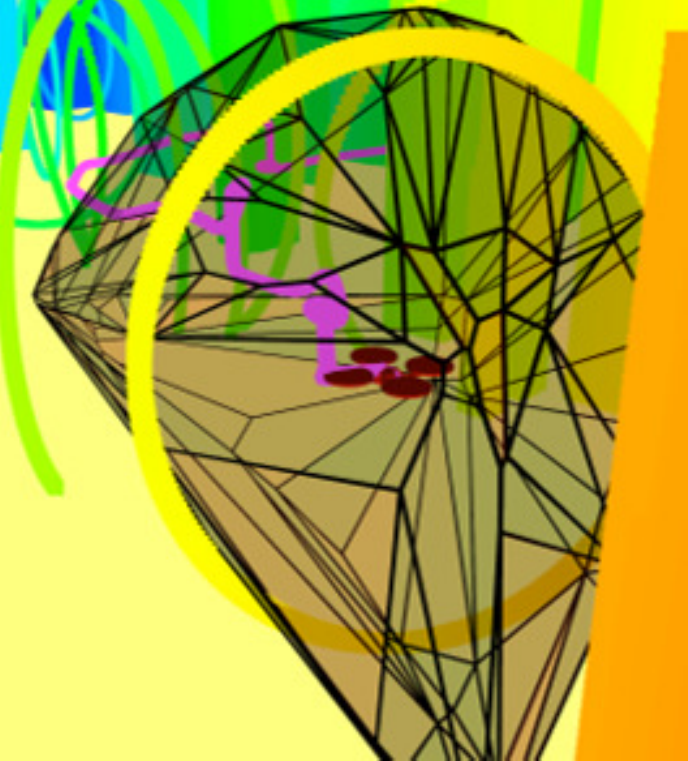}}     
\end{center}
\vspace{-0.25cm}
\caption{The comparison of an axis-aligned cube and a general convex polyhedron. The cube and the polyhedron are generated to the largest volume they may have, from the same seed coordinate. The way to inflate the cube is stated in our previous paper~\cite{fei2019ral}. The method to find the general free polyhedron will be detailed later in Sect.~\ref{subsec:convex_polyhedron_inflation}. 
\label{fig:poly_cube_rviz} 
}
\vspace{-1.5cm}
\end{figure}

\section{Flight Corridor Generation}
\label{sec:corridor_generation}
As stated in Sect.~\ref{subsec:coordinate_descent}, the first step of our global planning is to build a flight corridor around the teaching trajectory for spatial-temporal trajectory optimization.
In our previous work~\cite{fei2019ral}, the flight corridor is constructed by finding a series of axis-aligned cubes, which may sacrifice much space, especially in a highly nonconvex environment, as is shown in Fig~\ref{fig:cube_polygon_compare}. 
A more illustrative comparison is shown in Fig.~\ref{fig:poly_cube_rviz}, where the convex polyhedron captures much more free space than the simple cube.
Using simple axis-aligned cubes significantly limit the solution space of trajectory optimization, which may result in a poor solution.
What's more, in situations where the free space is very limited, such as flying through a very narrow circle, a cube-based corridor~\cite{fei2019ral} may even fail to cover all teaching trajectory and result in no solutions existing in the corridor. 
Therefore, to utilize the free space more sufficiently and adapt to even extremely cluttered maps, we propose a method to generate general, free, large convex polyhedrons. 

Since the human's teaching trajectory may be arbitrarily jerky, we cannot assume there is a piecewise linear path to initiate the polyhedron generation, as in~\cite{liu2017ral}. 
Also, we make no requirements on the convexity of obstacles in the map as in~\cite{deits2015computing}. 
Our method is based on convex set clustering, which is similar to~\cite{blochliger2018topomap}, but is different and advanced at:
\begin{enumerate}
\item We make no assumption on the growing directions of convex clusters and generate completely collision-free polyhedrons based on our dense occupancy map.
\item We introduce several careful engineering considerations which significantly speed-up the clustering.
\item We fully utilize the parallel structure of this algorithm and accelerate it over an order of magnitude in GPUs.
\item We introduce a complete pipeline from building the convex polyhedron clusters to establishing constraints in trajectory optimization.
\end{enumerate} 

\subsection{Convex Cluster Inflation}
\label{subsec:convex_polyhedron_inflation}

\begin{figure}[t]
\begin{center}          
\subfigure[\label{fig:cluster_ray_cast1} ]
{\includegraphics[width=0.49\columnwidth]{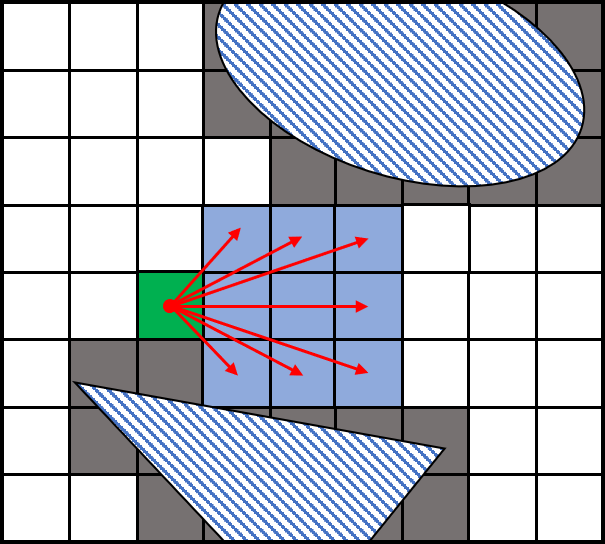}}       
\subfigure[\label{fig:cluster_ray_cast2} ]
{\includegraphics[width=0.49\columnwidth]{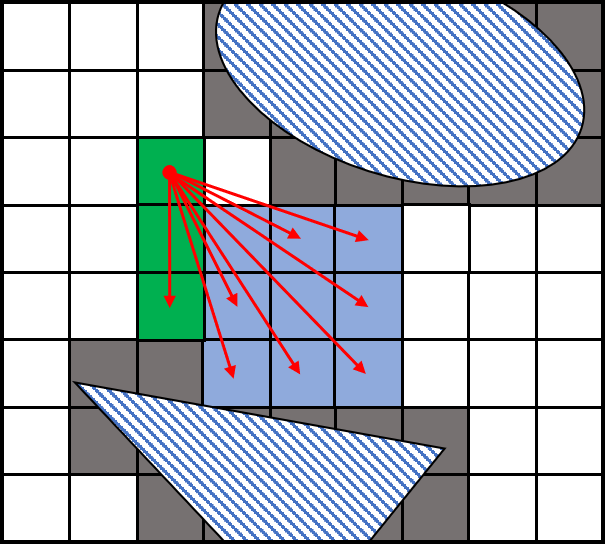}}     
\subfigure[\label{fig:cluster_ray_cast3} ]
{\includegraphics[width=0.49\columnwidth]{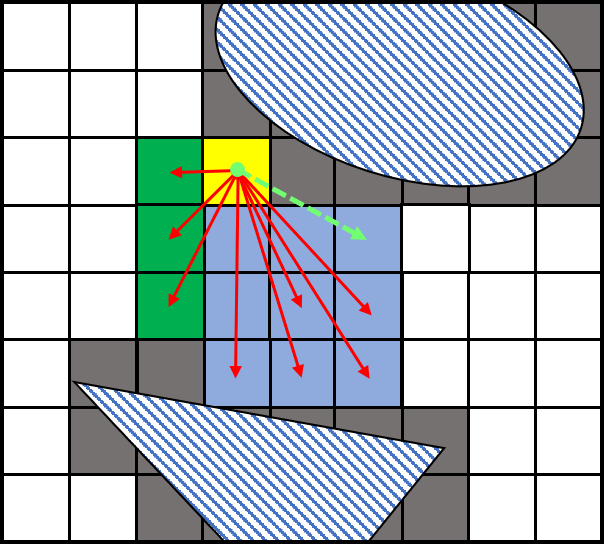}}       
\subfigure[\label{fig:cluster_ray_cast4} ]
{\includegraphics[width=0.49\columnwidth]{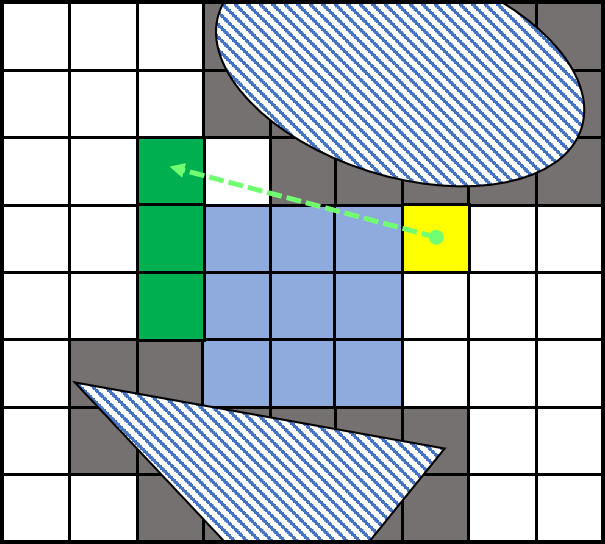}}     
\end{center}
\vspace{-0.5cm}
\caption{An illustration of the \textit{convex cluster inflation}. 
In (a) and (b), all qualified neighbor voxels are added to the \textit{convex cluster}. 
In (c) and (d), since an occupied voxel occludes a ray (the green arrow) to one of the clustered voxels, the testing voxel (in yellow) is excluded to the convex cluster.
\label{fig:cluster_ray_cast} 
}
\vspace{-1.5cm}
\end{figure}  

The core algorithm for the construction of the flight corridor is to find the largest convex free polyhedron at a given coordinate.
In this paper, we use an occupancy grid map to represent the environment. 
Each polyhedron in the flight corridor is the convex hull of a voxel set, which is convex and contains only free voxels.
The voxel set is found by clustering as many free voxels as possible around an arbitrary seed voxel.
In this paper, we name the voxel set as \textit{convex cluster}, and the process of finding such a set as \textit{convex cluster inflation}. Our method for finding such a \textit{convex cluster} is based on the definition of the convex set:

\textit{Definition}: A set $\mathcal{S}$ in a vector space over $\mathcal{R}$ is called a convex set if the line segment joining any pair of points of $\mathcal{S}$ lies entirely in $\mathcal{S}$.~\cite{lay2007convex}.

The pipeline for iteratively inflating such a cluster while preserving convexity is stated in Alg.~\ref{alg:polyhedron_inflation}. 
Our method operates on a 3D occupancy map $\mathcal{M}$ where voxels are labeled as \textit{obstacle} or \textit{free}. 
Three voxel sets are maintained in the algorithm. 
$\mathcal{C}$ stands for the targeting convex voxel cluster. 
$\mathcal{C}^+$ is the set of voxels that are tentative to be added to $\mathcal{C}$ in this iteration. 
And $\mathcal{C}^*$ contains newly added voxels which preserve the convexity. 
The cluster inflation starts by adding the seed voxel $p$ to $\mathcal{C}$, and adding all neighboring voxels of $p$ to $\mathcal{C}^+$. 
In an iteration, each voxel $p^+$ in $\mathcal{C}^+$ is checked whether it can preserve convexity using the function CHECK\_CONVEXITY($p^+, \mathcal{C}$, $\mathcal{M}$). 
This function, as is shown in Alg.~\ref{alg:convex_check}, casts rays from $p^+$ to each existing voxels in $\mathcal{C}$. According to the definition of the convex set, $\mathcal{M}$ with $p^+$ is convex if and only if all rays are collision-free. Based on this criteria, qualified voxels are considered as \textit{active} voxels and are added into $\mathcal{C}$ and $\mathcal{C}^*$. 
Neighboring voxels of all \textit{active} voxels $p^*$ are traversed and collected by the function GET\_NEIGHBORS($\mathcal{C}^*$) for next iteration. 
The inflation ends when $\mathcal{C}^+$ becomes empty, which implies no additional voxels can be added into $\mathcal{C}$. Fig.~\ref{fig:cluster_ray_cast} also illustrates the procedure of the \textit{convex cluster inflation}. 

\begin{algorithm}[t]
\caption{Convex Cluster Inflation}  
\label{alg:polyhedron_inflation}  
\begin{algorithmic}[1]
\State \textbf{Notation}: 
\State Seed Voxel: $p^s$, Grid Map: $\mathcal{M}$, Convex Cluster: $\mathcal{C}$, \\ Candidate Voxel Set: $\mathcal{C}^+$,  Active Voxel Set: $\mathcal{C}^*$
\State \textbf{Input}: $p^s$, $\mathcal{M}$
\Function{CONVEX\_INFLATION}{$p^s$, $\mathcal{M}$}
\State $\mathcal{C} \:\:\: \leftarrow \{p^s\}$
\State $\mathcal{C}^* \:   \leftarrow \varnothing$
\State $\mathcal{C}^+ \leftarrow$ GET\_NEIGHBORS($\mathcal{C}$)
\While{$\mathcal{C}^+ \neq \varnothing $}
\ForEach {$p^+ \in \mathcal{C}^+ $}
\If {CHECK\_CONVEXITY($p^+, \mathcal{C}$, $\mathcal{M}$)}
\State $\mathcal{C} \:\:\: \leftarrow \mathcal{C} \:\:\cup p^+ $
\State $\mathcal{C}^* \: \leftarrow \mathcal{C}^* \cup p^+$
\EndIf
\EndFor
\State $\mathcal{C}^+ \leftarrow \varnothing$
\State $\mathcal{C}^+ \leftarrow \mathcal{C}^+ \cup $ GET\_NEIGHBORS($\mathcal{C}^*$)
\State $\mathcal{C}^* \leftarrow \varnothing$
\EndWhile  
\State \textbf{Output}: $\mathcal{C}$
\EndFunction        
\end{algorithmic}  
\end{algorithm} 

\begin{algorithm}[t]
\caption{Convexity Checking}  
\label{alg:convex_check}  
\begin{algorithmic}[1]
\State \textbf{Notation}: Ray Cast: $l$
\Function{CHECK\_CONVEXITY}{$p^+, \mathcal{C}$, $\mathcal{M}$}
\ForEach {$p \in \mathcal{C} $}
\State $l \leftarrow$ CAST\_RAY($p^+$, $p$)
\If {PASS\_OBS($l, \mathcal{M}$)}
\State \Return False
\EndIf
\EndFor
\State \Return True  
\EndFunction
\end{algorithmic}  
\end{algorithm} 

Having a \textit{convex cluster} consists of a number of voxels, we convert it to the algebraic representation of a polyhedron for following spatial trajectory optimization. 
Quick hull algorithm~\cite{barber1996quickhull} is adopted here for quickly finding the convex hull of all clustered voxels. The convex hull is in vertex representation (V-representation) $\{V_0, V_1, ..., V_m \}$, and is then converted to its equivalent hyperplane representation (H-representation) by using double description method~\cite{fukuda1995double}. 
The H-representation of a 3-D polyhedron is a set of affine functions:
\begin{align}
\label{eq:h_representation}
&a_0^x \cdot \textbf{x} + a_0^y \cdot \textbf{y} + a_0^z \cdot \textbf{z} \leq k_0, \nonumber \\
&a_1^x \cdot \textbf{x} + a_1^y \cdot \textbf{y} + a_1^z \cdot \textbf{z} \leq k_1, \nonumber \\
&\:\:\:\:\:\:\:\:\:\:\:\:\:\:\:\:\:\:\:\: \vdots  \\
&a_n^x \cdot \textbf{x} + a_n^y \cdot \textbf{y} + a_n^z \cdot \textbf{z} \leq k_n, \nonumber
\end{align}
where $\{a_n^x, a_n^y, a_n^z \}$ is the normal vector of the 3-D hyperplane and $k_n$ is a constant.

\subsection{CPU Acceleration}
\label{subsec:cpu_acceleration}

As is shown in Alg.~\ref{alg:polyhedron_inflation}, determining whether a voxel preserves the convexity needs to conduct ray-casting to all existing voxels in the \textit{convex cluster}. 
Iterating with all voxels and rays makes this algorithm impossible to run in real-time, especially when the occupancy grid map has a fine resolution. 
To make the polyhedron generated in real-time, we take careful engineering considerations on the implementations and propose some critical techniques to significantly increase the overall efficiency. 

\subsubsection{Polyhedron Initialization}
We initialize each convex cluster as an axis-aligned cube using our previous method~\cite{fei2018icra}, which can be done very fast since only index query ($\mathcal{O}$(1)) operations are necessary. 
After inflating the cube to its maximum volume, as in Fig.~\ref{fig:cube_polygon_compare}, we switch to the convex clustering to further group convex free space around the cube. 

The proposed polyhedron initialization may result in a final polyhedron different from the one which is clustered from scratch.  This is because an axis-aligned cube only inflates in $x, y, z$ directions while a \textit{convex cluster} grows in all possible directions (26-connections in a 3D grid map). 
However, this initialization process is reasonable. 
Our purpose is not making each polyhedron optimal but capturing as much as possible free space than a simple cube cannot. 
In practice, the initialization provides a fast discovery of nearby space which is easy to group, and does not prevent the following \textit{convex cluster inflation} to refine the polyhedron and find sizeable free space.
In Sect.~\ref{subsubsec:compare_corridor}, we show that the initialization process significantly improves the computing efficiency with only a neglectable sacrifice on the volume of the final polyhedron. 

\subsubsection{Early Termination}
We label all voxels in the cluster as \textit{inner} voxels which inside the \textit{convex cluster}, and \textit{outer} voxels which on the boundary of the \textit{convex cluster}.
When traversing a ray from a candidate voxel to a voxel in the \textit{convex cluster}, we early terminate the ray casting when it arrives at a voxel labeled as \textit{inner}.

\textit{Theorem 1}: \label{Theo: early_termination} The early termination at \textit{inner} voxels is sufficient for checking convexity.

\textit{Proof}: According to the definition of convex set, a ray connecting an \textit{inner} voxel to any other voxel in the \textit{convex cluster} lies entirely in the \textit{convex cluster}. Hence, the extension line of an \textit{inner} voxel must lie inside the \textit{convex cluster} and therefore it must pass the convexity check.

\subsubsection{Voxel Selection}
To further reduce the number of voxels that need to be cast rays, given a candidate voxel, only \textit{outer} voxels are used to check its convexity.

\textit{Theorem 2}: \label{Theo: voxel_selection} Using \textit{outer} voxels of a \textit{convex cluster} is sufficient for checking convexity.

\textit{Proof}: Obviously, the \textit{convex cluster} is a closed set with \textit{outer} voxels at its boundary. 
The candidate voxel is outside this set.
Therefore, casting a ray from any \textit{inner} voxel to the candidate voxel must pass one of the \textit{outer} voxels. 
According to \textit{Theorem 1}, checking convexity of this ray can terminate after the ray passes an \textit{outer} voxels, which means for a candidate voxel, checking rays cast to \textit{outer} voxels is sufficient.
 
By introducing above techniques, the proposed \textit{convex cluster inflation} can work in real time for a mediate grid resolution ($0.2m$) on CPUs. The efficacy of these techniques is numerically validated in Sect.~\ref{subsubsec:compare_corridor}.

\subsection{GPU Acceleration}
\label{subsec:gpu_acceleration}
We propose a parallel computing scheme that significantly speeds up the inflation by one order of magnitude where a GPU is available. 
As is shown in Sec.~\ref{subsec:convex_polyhedron_inflation}, when the \textit{convex cluster} discovers a new neighboring voxel, sequentially traversing and checking all rays is naturally parallelizable. 
With the help of many core GPUs, we can cast rays and check collisions parallelly. 
Moreover, to fully utilize the massively parallel capability of a GPU, reduce the serialize operations, and minimize the data transferring between CPU and GPU, we examine all potential voxels of the cluster parallelly in one iteration. 
Instead of discovering a new voxel and checking its rays, we find all neighbors of the active set $\mathcal{C}^*$ and check their rays all in parallel. 
The detailed procedure is presented in Alg.~\ref{alg:parallel_polyhedron_inflation},
where GET\_NEIGHBORS($\mathcal{C}$) collects all neighbors of a set of voxels, and PARA\_CHECK\_CONVEXITY($\mathcal{C}^+, \mathcal{C}$, $\mathcal{M}$) checks the convexity of all candidate voxels parallelly in GPUs. 
Note that in the serialized version of the proposed method, the voxel discovered earlier may prevent later ones from being clustered, as is illustrated in Fig.~\ref{fig:cluster_ray_cast}.
However, in the parallel clustering, all voxels examined at the same time may add conflicting voxels to the cluster. 
Therefore, we introduce an additional variable, $r$ to record sequential information of voxels.
As shown in Alg.~\ref{alg:parallel_convex_check}, the kernel function is running on the GPU per block.
It checks the ray cast from every candidate voxel in $\mathcal{C}^+$ to a cluster voxel in $\mathcal{C}$ and to each other candidate voxel which has a prior index. 
After that, the function CHECK\_RESULTS($r$) selects all qualified voxels and adds them into $\mathcal{C}$. 
Firstly, candidate voxels that have collisions with $\mathcal{C}$ are directly excluded. 
Then, candidate voxels having collisions with other candidates that have already been added into $\mathcal{C}$ are excluded. In this way, we finally get the same results as in the serialized version of the clustering. The efficacy of the parallel computing is shown in Sect.~\ref{subsubsec:compare_corridor}.

\begin{algorithm}[t]
\caption{Parallel Convex Cluster Inflation}  
\label{alg:parallel_polyhedron_inflation}  
\begin{algorithmic}[1]
\State \textbf{Notation}: 
\State Parallel Raycasting Result: $r$
\Function{PARA\_CONVEX\_INFLATION}{$p^s$, $\mathcal{M}$}
\State $\mathcal{C} \:\:\: \leftarrow \{ p^s \}$
\State $\mathcal{C}^* \:   \leftarrow \varnothing$
\State $\mathcal{C}^+      \leftarrow$ GET\_NEIGHBORS($\mathcal{C}$)
\While{$\mathcal{C}^+ \neq \varnothing $}

\State /* \textit{GPU data uploads} */ 
\State $r \:     \leftarrow$ PARA\_CHECK\_CONVEXITY($\mathcal{C}^+, \mathcal{C}$, $\mathcal{M}$)
\State /* \textit{GPU data downloads} */ 
\State $\mathcal{C}^* \: \leftarrow$ CHECK\_RESULTS($r$)
\State $\mathcal{C}   \:\:\: \leftarrow \mathcal{C} \cup \mathcal{C}^*$
\State $\mathcal{C}^+        \leftarrow$ GET\_NEIGHBORS($\mathcal{C}^*$)
\EndWhile  
\State \textbf{Output}: $\mathcal{C}$
\EndFunction 
\end{algorithmic}  
\end{algorithm} 

\begin{algorithm}[t]
\caption{Parallel Convexity Checking}  
\label{alg:parallel_convex_check}  
\begin{algorithmic}[1]
\Function{PARA\_CHECK\_CONVEXITY}{$\mathcal{C}^+, \mathcal{C}$, $\mathcal{M}$}
\ForEach {$p^+_i \in \mathcal{C}^+ $}
\State $r[i].status \leftarrow True$
\ForEach {$p \in \mathcal{C} $}
\State /* \textit{Kernel function starts} */ 
\State $l \leftarrow$ CAST\_RAY($p^+_i$, $p$)
\If {PASS\_OBS($l, \mathcal{M}$)}
\State $r[i].status \leftarrow False$
\EndIf
\State /* \textit{Kernel function ends} */ 
\EndFor
\ForEach {$p^+_j \in \mathcal{C}^+$ AND $j < i$}
\State /* \textit{Kernel function starts} */ 
\State $l \leftarrow$ CAST\_RAY($p^+_i$, $p^+_j$)
\If {PASS\_OBS($l, \mathcal{M}$)}
\State $r[i].status \leftarrow Pending$
\State $r[i].list.push\_back(j)$
\EndIf
\State /* \textit{Kernel function ends} */ 
\EndFor
\EndFor
\State \Return $r$
\EndFunction

\State
\Function{CHECK\_RESULTS}{$r, \mathcal{C}^+$}
\ForEach {$p^+_i \in \mathcal{C}^+ $} \label{marker}
\If {$r[i].status == True$}
\State $\mathcal{C}^* \leftarrow \mathcal{C}^* \cup p^+$
\ElsIf {$r[i].status == Pending$}
\ForEach {$j \in r[i].list $}
\If {$p^+_j \in \mathcal{C}^* $}
\State \Goto{marker}
\EndIf
\EndFor
\State $\mathcal{C}^* \leftarrow \mathcal{C}^* \cup p^+$
\EndIf
\EndFor
\State \Return $\mathcal{C}^*$
\EndFunction
\end{algorithmic}  
\end{algorithm} 

\subsection{Corridor Generation and Loop Elimination}
\label{subsec:loop_elimination}
\begin{figure*}[t]
\centering
  \subfigure[\label{fig:corridor1} ]
{\includegraphics[width=0.39\columnwidth]{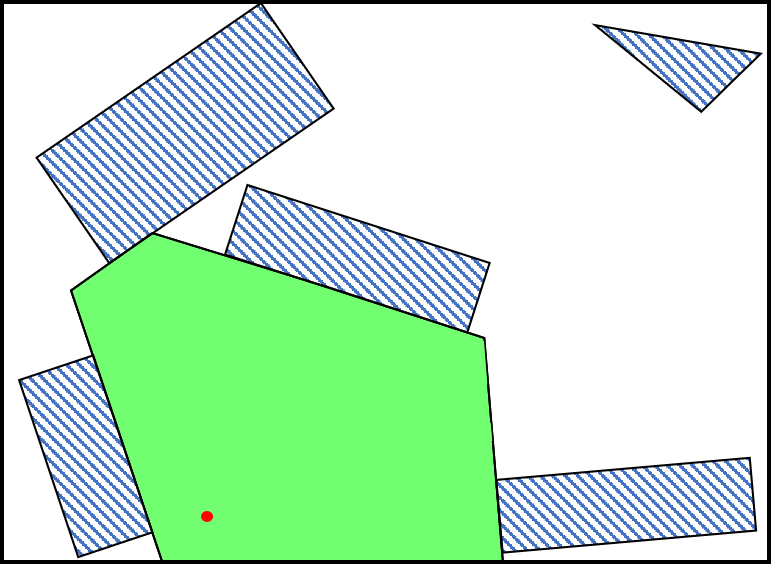}}
  \subfigure[\label{fig:corridor2} ]
{\includegraphics[width=0.39\columnwidth]{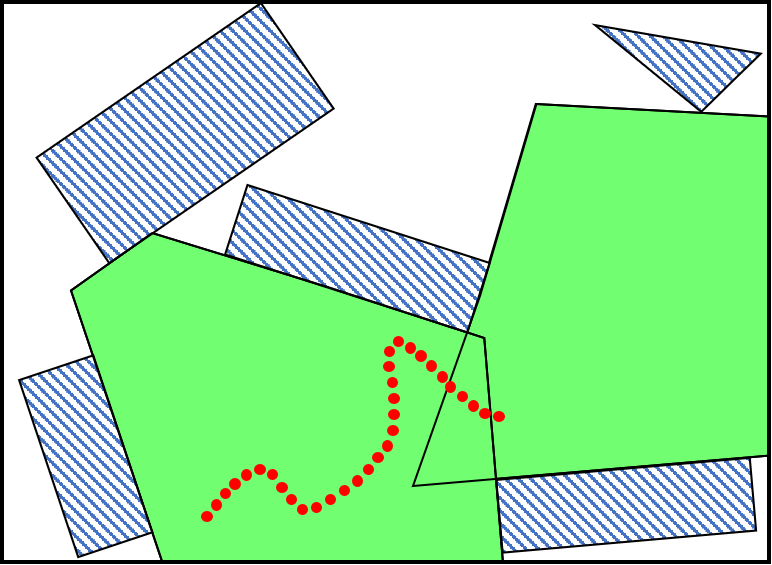}}
  \subfigure[\label{fig:corridor3} ]
{\includegraphics[width=0.39\columnwidth]{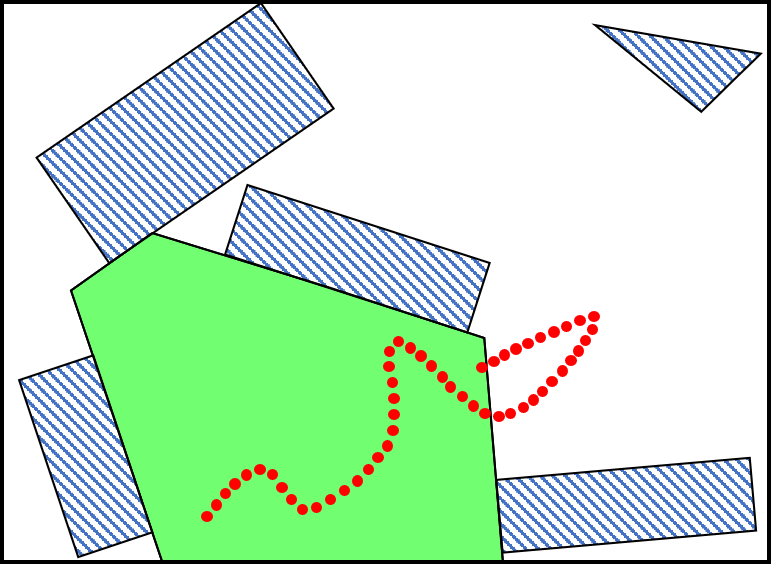}}
  \subfigure[\label{fig:corridor4} ]
{\includegraphics[width=0.39\columnwidth]{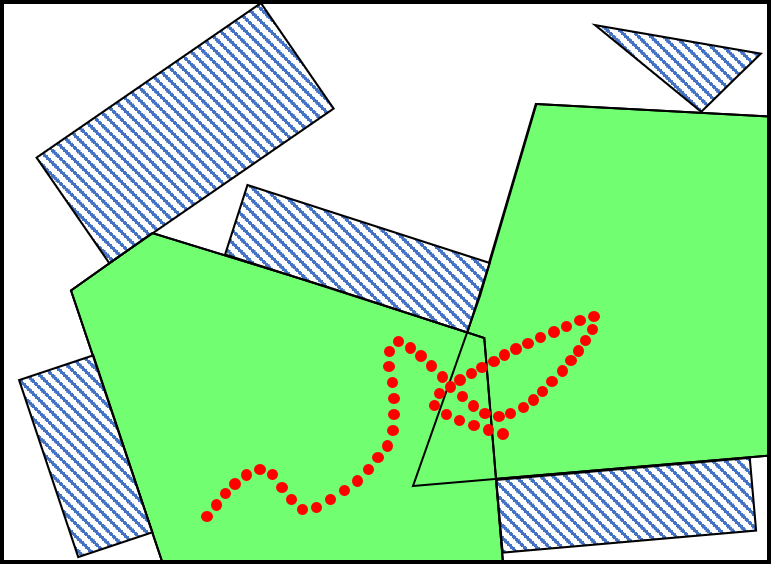}}
  \subfigure[\label{fig:corridor5} ]
{\includegraphics[width=0.39\columnwidth]{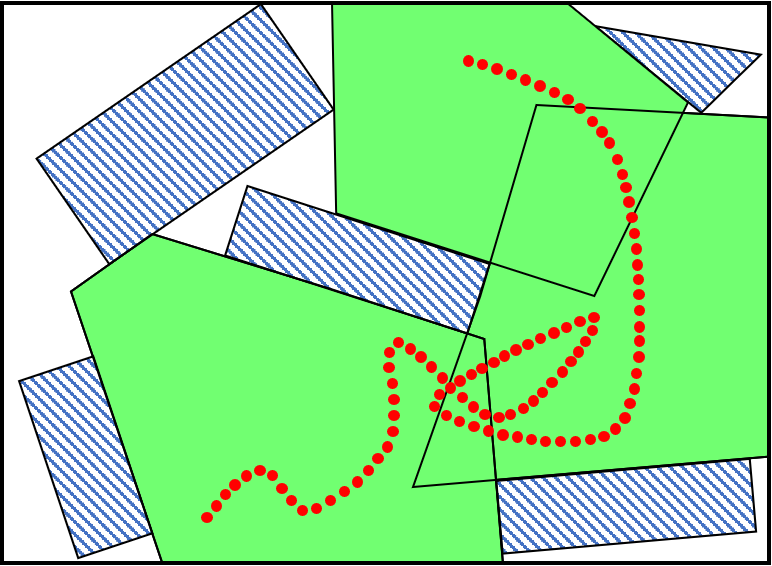}}
\caption{The flight corridor generation process. Red dots are coordinates along the teaching trajectory. (b), a new convex polyhedron is generated and added to the flight corridor when the drone leaves the corridor enters undiscovered space. (c), the drone leaves the last polyhedron and returns back to the second to last one, so the last polyhedron is deleted from the corridor. 
\label{fig:corridor_pipeline}}
\end{figure*}

Since the trajectory provided by a user may behave arbitrarily jerky and contain local loops, we introduce a specially designed mechanism to elliminate unnecessary loops, i.e., repeatable polyhedrons.
The exclusion of repeatable polyhedrons is essential since in following trajectory optimization (Sect.~\ref{sec:trajectory_optimization}), each polyhedron is assigned with one piece of the trajectory.
Repeatable polyhedrons would result in an optimized trajectory loops as the user does, which is obviously not efficient.
The pipeline of the corridor generation is shown in Alg.~\ref{alg:local_loop} and Fig.~\ref{fig:corridor_pipeline}. 
At the beginning of the teaching, the flight corridor is initialized by finding the maximum polyhedron around the position of the drone. 
Then as the human pilots the drone to move, we keep checking the drone's position. 
If it goes outside the last polyhedron ($\mathcal{G}$[-1]), we further check whether the drone discovers new free space or not. 
If the drone is contained within the second last polyhedron ($\mathcal{G}$[-2]), we can determine that the teaching trajectory
has a loop, as shown in Fig.~\ref{fig:corridor3}.
Then, the last polyhedron in the corridor is regarded as repeatable and is therefore popped out from the corridor.
Otherwise, as shown in Fig.~\ref{fig:corridor4}, the drone is piloted to discover new space. 
Then a new polyhedron $\mathcal{P}$ is inflated and added to the tail of the corridor. 
The corridor generation is terminated when the teaching finish.
The final flight corridor shares the same topological structure with the teaching trajectory since no obstacles are included in the corridor.
And it has no unnecessary loops. 

\begin{algorithm}[t]
\caption{Flight Corridor Generation}  
\label{alg:local_loop}  
\begin{algorithmic}[1]  
\State \textbf{Notation}: Flight Corridor $\mathcal{G}$, Drone Position $p$, Convex Polyhedron $\mathcal{P}$
\State Initialize :
\State $\mathcal{P} \leftarrow$ CONVEX\_INFLATION($p, \mathcal{M}$)
\State $\mathcal{G}$.push\_back($\mathcal{P}$)
\While{ Teaching }
\State $p \leftarrow$ UPDATE\_POSE()
\If { OUTSIDE($p$, $\mathcal{G}$[-1]) }
\If { INSIDE($p$, $\mathcal{G}$[-2]) }
\State $\mathcal{G}$.pop\_back()
\Else
\State $\mathcal{P}$ = CONVEX\_INFLATION($p, \mathcal{M}$)
\State $\mathcal{G}$.push\_back($\mathcal{P}$)
\EndIf
\EndIf
\EndWhile  
\State \Return $\mathcal{G}$
\end{algorithmic}  
\end{algorithm} 

\section{Spatial-Temporal Global Trajecotry Optimization}
\label{sec:trajectory_optimization}
\subsection{Spatial Trajecotry Optimization}
\label{subsec:space_optimization}
For the spatial optimization, we use the Bernstein basis to represent the trajectory as a piecewise B\'ezier curve, since it can be easily constrained in the flight corridor by enforcing constraints on control points. 
An $i^{th}$-order Bernstein basis is:
\begin{equation}
b_{n}^{i}(t) = \binom{n}{i} \cdot t^i \cdot (1-t)^{n-i},
\end{equation}
where $n$ is the degree of the basis, $\binom{n}{i}$ is the binomial coefficient and $t$ is the variable parameterizing the trajectory. 
An $N$-piece piecewise B\'ezier curve is written as:
\begin{equation} 
\label{eq:spatial_curve_n_piece}
 \textit{f}_{\mu}\textit{(t)} =  
 \begin{cases} 
	\sum_{i=0}^n c_{\mu, 1}^ib_{n}^{i}(t / T_1), & t\in[0, T_1], \\
    \sum_{i=0}^n c_{\mu, 2}^ib_{n}^{i}(t / T_2), & t\in[0, T_2], \\
      \:\:\: \:\:\:\:\:\:\:\:\:\vdots      &\:\:\:\:\:\:\:\:\:\vdots \\
	\sum_{i=0}^n c_{\mu, N}^ib_{n}^{i}(t / T_N), &t\in[0, T_N].
   \end{cases}
\end{equation}
For the $m^{th}$ piece of the curve, $c_{\mu. m}^i$ is the $i^{th}$ control point, and $T_m$ is the time duration. 
The spatial trajectory is generated in $x, y, z$ dimensions, and $\mu \in {x, y, z}$.
$\mu$ is omitted in following derivation for brevity.
In this equation, $t$ is scaled by $T_m$ since a standard B\'ezier curve is defined on $[0,1]$.

Follow the formulation in minimum-snap~\cite{MelKum1105}, the squared jerk is minimized in this paper. 
Since the $3^{rd}$ order derivative of a curve corresponds to the angular velocity, the minimization of jerks alleviates the rotation and therefore facilitates visual tracking. 
The objective of the piecewise curve is:
\begin{equation}
J = \sum_{\mu}^{x,y,z} \sum_{m=1}^N \int_{0}^{T_m} \left(\frac{d^3f_{\mu, m}(t)}{dt^3}\right)^2\, dt.
\end{equation}
which is in a quadratic form denoted as $\mathbf{c}^T \mathbf{Q} \mathbf{c}$. 
Here $\mathbf{c}$ is composited by all control points in \textit{x, y, z} dimensions. 
$\mathbf{Q}$ is a semi-definite Hessian matrix. 

For a B\'ezier curve, its higher order derivatives can be represented by linear combinations of corresponding lower-order control points. For the $1^{st}$ and $2^{nd}$ order derivatives of the $m^{th}$ piece of the curve in Eq.~\ref{eq:spatial_curve_n_piece}, we have:
\begin{align} 
\label{eq:bezier_v_m}
& f^\prime_m(t) 		 = \sum_{i=0}^{n-1} n (c_m^{i+1} - c_m^{i}) b_{n-1}^{i}(\frac{t}{T_m}),  \\
& f^{\prime \prime}_m(t) = \sum_{i=0}^{n-2} n (n - 1) (c_m^{i+2} - 2 c_m^{i+1} + c_m^{i}) b_{n-2}^{i}(\frac{t}{T_m}). \nonumber
\end{align}

\subsubsection{Boundary Constraints}
The trajectory has the boundary constraints on the initial state ($p^0, v^0, a^0$) and the final state ($p^f, v^f, a^f$) of the quadrotor. 
Since a B\'ezier curve always passes the first and last control points, we enforce the boundary constraints by directly setting equality constraints on corresponding control points in each dimension:
\begin{align}
\label{eq:boundary_constraints}
&c_0^{0} = p^0, \nonumber \\
&c_N^{n}  = p^f, \nonumber \\
&n (c_0^{1} - c_0^{0}) = v^0,  \\
&n (c_N^{n} - c_N^{n-1}) = v^f, \nonumber \\
&n (n - 1) (c_0^{2} - 2 c_0^{1} + c_0^{0}) = a^0, \nonumber \\
&n (n - 1) (c_N^{n} - 2 c_N^{n-1} + c_N^{n-2}) = a^f. \nonumber 
\end{align}

\subsubsection{Continuity Constraints}
For ensuring smoothness, the minimum-jerk trajectory must be continuous for derivatives up to $2^{nd}$-order at all connecting points on the piecewise trajectory. The continuity constraints are enforced by setting equality constraints between corresponding control points of two consecutive curves. 
For the $j^{th}$ and $(j+1)^{th}$ pieces of the curve, we can write the equation in each dimension as:
\begin{align}
\label{eq:continunous_constraints}
&  c_j^{n} = c_{j+1}^{0}, \nonumber \\
& (c_j^{n} - c_j^{n-1}) / T_j = (c_{j+1}^{1} - c_{j+1}^{0}) / T_{j+1},  \\
& (c_j^{n} - 2 c_j^{n-1} + c_j^{n-2}) / T^{2}_j = (c_{j+1}^{2} - 2 c_{j+1}^{1} + c_{j+1}^{0}) / T^{2}_{j+1}, \nonumber 
\end{align}

\subsubsection{Safety Constraints}
The safety of the trajectory is guaranteed by enforcing each piece of the curve to be inside the corresponding polyhedron. Thanks to the convex hull property, an entire B\'ezier curve is confined within the convex hull formed by all its control points. Therefore we constrain control points using hyperplane functions obtained in Eq.~\ref{eq:h_representation}. For the $i^{th}$ control point $c_{j, x}^{i}, c_{j, y}^{i}, c_{j, z}^{i}$ of the $j^{th}$ piece of the trajectory in $x, y, z$ dimensions, constraints are:
\begin{align}
\label{eq:safety_constraints}
&a_0^x \cdot c_{j, x}^{i} + a_0^y \cdot c_{j, y}^{i} + a_0^z \cdot c_{j, z}^{i} \leq k_0, \nonumber \\
&a_1^x \cdot c_{j, x}^{i} + a_1^y \cdot c_{j, y}^{i} + a_1^z \cdot c_{j, z}^{i} \leq k_1, \nonumber \\
&\:\:\:\:\:\:\:\:\:\:\:\:\:\:\:\:\:\:\:\: \vdots  \\
&a_n^x \cdot c_{j, x}^{i} + a_n^y \cdot c_{j, y}^{i} + a_n^z \cdot c_{j, z}^{i} \leq k_n, \nonumber
\end{align}

Constraints in Equs.~\ref{eq:boundary_constraints} and~\ref{eq:continunous_constraints} are affine equality constraints (\textit{$\mathbf{A}_{eq}\mathbf{c} = \mathbf{b}_{eq}$}) and Eq.~\ref{eq:safety_constraints} is in affine in-equality formulation (\textit{$\mathbf{A}_{ie}\mathbf{c} \leq \mathbf{b}_{ie}$}). Finally, the spatial trajectory optimization problem is formulated as a QP as follows:
\begin{align}
\label{eq:spatial_qp_program}
 \text{min}  \:\:\:\:\:\: &\mathbf{c}^T \mathbf{Q} \mathbf{c} \nonumber \\
 \text{s.t.} \:\:\:\:\:\: &\mathbf{A}_{eq} \mathbf{c}   = \mathbf{b}_{eq}, \\
&\mathbf{A}_{ie}\mathbf{c} \leq \mathbf{b}_{ie}.  \nonumber
\end{align} 
Unlike our previous works on corridor constrained trajectory~\cite{fei2018icra,fei2018jfr}, here the kinodynamic feasibility (velocity and acceleration) is not guaranteed by adding higher-order constraints into this program, but by temporal optimization (Sect.~\ref{subsec:time_optimization}). 
For a rest-to-rest trajectory, the program in Eq.~\ref{eq:spatial_qp_program} is always mathematically feasible. 

\subsection{Temporal Trajectory Optimization}
\label{subsec:time_optimization}
\begin{figure}[t]
\centering
\includegraphics[width=0.99\columnwidth]{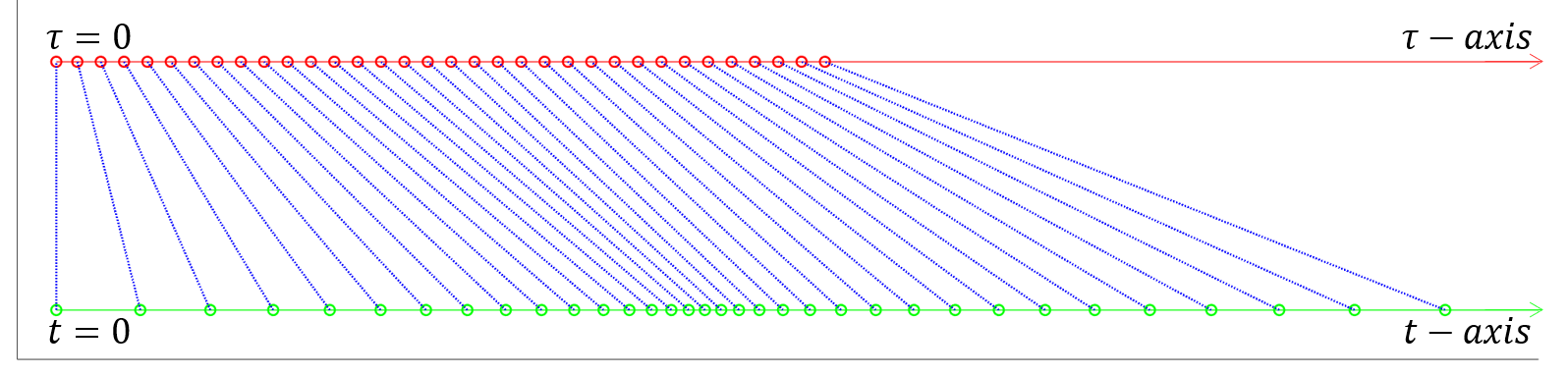}
\caption{The effect of the temporal optimization. $t$ and $\tau$ are the time profile of the spatial trajectory before and after optimization. \label{fig:t_tau_timeline}}
\vspace{-0.75cm}
\end{figure}

In spatial optimization, a corridor-constrained spatial trajectory is generated given a fixed time allocation. 
To optimize the trajectory temporally, we design a re-timing function $\{t(\tau): \tau \rightarrow t \}$ to map the  original time variable $t$ to a variable $\tau$.
The relation between $\tau$ and $t$ is shown in Fig.~\ref{fig:t_tau_timeline}.
In this paper, the re-timing function $t(\tau)$ is named as the temporal trajectory, and finding the optimal $t(\tau)$ is called the temporal optimization.
For the N-piece spatial curve defined in Equ.~\ref{eq:spatial_curve_n_piece}, we write $t(\tau)$ as a corresponding N-piece formulation:
\begin{equation} 
\label{eq:piecewise_tau_function}
t(\tau) =  
 \begin{cases} 
      t_1(\tau), &t_1(0) = 0, t_1(\mathcal{T}^*_1) = T_1, t_1 \in [0, T_1] \\
      t_2(\tau), &t_2(0) = 0, t_2(\mathcal{T}^*_2) = T_2, t_2 \in [0, T_2]   \\
      \:\:\: \vdots     & \:\:\:\:\:\:\:\:\:\:\:\:\:\:\:\:\:\:\:\:\:\:\:\:\:\:\:\vdots \\
      t_N(\tau), &t_N(0) = 0, t_N(\mathcal{T}^*_N) = T_N, t_N \in [0, T_N]
   \end{cases}
\end{equation}
where $T_1, T_2, ... T_N$ are original time durations of the spatial curve $f_{\mu}(t)$, and $\mathcal{T}^*_1, \mathcal{T}^*_2, ... \mathcal{T}^*_N$ are time durations after temporal optimization. 
Since physically time only increases, $t(\tau)$ is a monotonically increasing function. 
Therefore we have $\dot{t}(\tau) \geq 0$. 
For clarity, in what follows, we use $c^\prime = dc/dt$ to denote taking derivatives with respect to $t$, and $\dot{c} = dc/d\tau$ for taking derivatives with respect to $\tau$.
By substituting $t$ with $t(\tau)$ in $f_{\mu}(t)$ and taking derivatives with chain rule, we can write the velocity as:
\begin{equation}
\label{eq:velocity}
\dot{f}(t(\tau))  = f^\prime(t) \cdot \dot{t},\nonumber \\
\end{equation}
\label{eq:kinodynamic}
and acceleration as:
\begin{equation}
\ddot{f}(t(\tau)) = f^\prime(t) \cdot \ddot{t} + f^{\prime\prime}(t) \cdot \dot{t}^2.
\end{equation}
The velocity and acceleration are also piecewise functions. 

\subsection{Minimum-Time Formulation}
\label{subsec:minimum_time_form}
\subsubsection{Objective}
\label{subsubsec:objective}
The total time $\mathcal{T}$ of the temporal trajectory can be written as:
\begin{equation} 
\label{eq:objective_raw}
\mathcal{T} =  \int_{0}^{\mathcal{T}} 1 d\tau = \sum_{m = 1}^{N} \int_{0}^{T_m} \frac{1}{\dot{t_m}} dt,
\end{equation}
considering $\dot{t} = dt/d\tau$.
We can introduce a regularization term that penalizes the changing rate of $t$, to trade-off the minimization of time and control extremeness, or so-called motion aggressiveness, in our final temporal trajectory.
The objective function is then written as:
\begin{equation} 
\label{eq:objective}
\mathcal{J} =  \sum_{m = 1}^{N} \int_{1}^{T_m} \Big( \frac{1}{\dot{t_m}} + \rho \cdot \ddot{t_m}^2 \Big) dt,
\end{equation}
where $\rho$ is a weight of the aggressiveness. 
By setting a larger $\rho$ we can obtain more gentle motions in the temporal trajectory.
If $\rho = 0$, the temporal optimization is solved for generating motions as fast as possible. 
The motions generated with a large $\rho$ can be viewed in our previous work~\cite{fei2019ral}.

Following the direct transcription method in~\cite{verscheure2009time}, $\alpha(t)$ and $\beta(t)$ are introduced as two additional piecewise functions:
\begin{align}
\label{eq:a_b_def}
\alpha_m(t) = \ddot{t}_m, \:\:\: \beta_m(t) = \dot{t}^2_m. \:\:\:\:\: m = 1,2,...,N.
\end{align}
According to the relationship between $\ddot{t}_m$ and $\dot{t}$, we can have:
\begin{align}
\beta_m(t) \geq 0, \:\:\: \beta_m^\prime(t) = 2\cdot \alpha_m(t). \label{eq:basic_con_2}
\end{align}
Then the objective function in Equ.~\ref{eq:objective} is re-formulated as:
\begin{equation} 
\label{eq:objective_ab}
\mathcal{J} =  \sum_{m = 1}^{N} \int_{0}^{T_m} \Big( \frac{1}{\sqrt{\beta_m(t)}} + \rho \cdot \alpha_m(t)^2 \Big) dt,
\end{equation}

\subsubsection{Constraints}
\label{subsubsec:temporal_constraints}
The continuities of $t(\tau)$ are enforced by setting constraints between every two consecutive pieces of it. 
In each dimension $\mu$ $\in {x, y, z}$, we have:
\begin{align}
&f_{\mu, m}^\prime(T_m) \cdot \sqrt{\beta_m(T_m)} = f_{\mu, m+1}^\prime(0) \cdot \sqrt{\beta_{m+1}(0)},\\
&f_{\mu, m}^\prime(T_m) \cdot \alpha_m(T_m) + f_{\mu, m}^{\prime\prime}(T_m) \cdot \beta_m(T_m) \nonumber \\
= &f_{\mu, m+1}^\prime(0) \cdot \alpha_{m+1}(0) + f_{\mu, m+1}^{\prime\prime}(0) \cdot \beta_{m+1}(0)
\end{align}
Then, to satisfy the initial and the final velocity and acceleration $a_{0}, v_0, a_f, v_f$, we set boundary constraints:
\begin{align}
& f_{\mu, 1}^\prime(0) \cdot \sqrt{\beta_1(0)} &= v_0, \\
& f_{\mu, N}^\prime(T_N) \cdot \sqrt{\beta_N(T_N)} &= v_f, \\
& f_{\mu, 1}^\prime(0) \cdot \alpha_1(0) + f_{\mu, 1}^{\prime\prime}(0) \cdot \beta_1(0) &= a_0, \\
& f_{\mu, N}^\prime(T_N) \cdot \alpha_N(T_N) + f_{\mu, N}^{\prime\prime}(T_N) \cdot \beta_N(T_N) &= a_f,
\end{align}
Finally, kinodynamic feasibility constraints are set as: 
\begin{align}
& - v_{max} \leq f_{\mu, m}^\prime(t) \cdot \sqrt{\beta_m(t)} \leq v_{max}, \\
& - a_{max} \leq f_{\mu, m}^\prime(t) \cdot \alpha_m(t) + f_{\mu, m}^{\prime\prime}(t) \cdot \beta_m(t) \leq a_{max},
\end{align}
where $v_{max}$ and $a_{max}$ are the physical limits of the drone.

\subsubsection{SOCP Re-formulation}
\label{subsubsec:socp_reform}
The above optimization problem has convex objective and constraints and is, therefore, a convex program.
To make it easily solvable,  for each piece of the trajectory, $t_m \in [0, T_m]$ is discretized to $t_m^0, t_m^1, ... t_m^{K_m}$ according to a given resolution $\delta t$. 
$K_m = \lceil T_m / \delta t\rceil + 1$. 
Then, $\alpha_m(t)$ becomes piecewise constant at each discretization point.
According to Equ.~\ref{eq:basic_con_2}, $\beta_m(t)$ is piecewise linear. 
In this way, $\alpha_m(t)$ and $\beta_m(t)$ are modeled by a series of discrete variables $\alpha_m^k$ and $\beta_m^k$, where $\beta_m^k$ is evaluated at $t_m^k$ and $\alpha_m^k$ is evaluated at $(t_m^k + t_m^{k+1}) / 2$.

By applying the above discretization, the objective in Eq.~\ref{eq:objective_ab} is derived as: 
\begin{align}
\label{eq:objective_dis}
\mathcal{J} = \sum_{m = 1}^{N} \sum_{k = 0}^{K_i-1} \bigg( \frac{2}
  { \sqrt{\beta_{m}^{k+1}} + \sqrt{ \beta_m^{k} } } + \rho \cdot (\alpha_m^k)^2 \bigg) \cdot \delta t,  
\end{align}
which is mathematically equivalent to the affine formulation:
\begin{equation}
\label{eq:objective_equ}
\sum_{m = 1}^{N} \sum_{k = 0}^{K_i-1} \Big( 2  \cdot \gamma_m^k + \rho \cdot (\alpha_i^k)^2 \Big) \cdot \delta t,
\end{equation}
by introducing $\gamma_m^k$ and
\begin{equation}
\label{eq:slack_equ_1}
\frac{1}{\sqrt{\beta_m^{k+1}} + \sqrt{ \beta_m^{k}}} \leq \gamma_m^k, \:\: k = 0,...K_i-1; m = 1, ... N    
\end{equation}
as slack variables and additional constraints.

Eq.~\ref{eq:slack_equ_1} is further derived to a quadratic form:
\begin{align}
\frac{1}{\zeta_m^{k+1} + \zeta_m^k} &\leq \gamma_m^k, \:\:\:\:\: k = 0,...K_i-1; m = 1, ... N. \label{eq:slack_equ_2}  \\
\zeta_m^k &\leq \sqrt{\beta_m^k}, \:\: k = 0,...K_i;  \:\:\:\:\:\: m = 1, ... N. \label{eq:slack_equ_3}
\end{align}
by introducing $\zeta_m^k$ as slack variables .

Eq.~\ref{eq:slack_equ_2} can be formulated as a standard rotated quadratic cone:
\begin{equation}
\label{eq:objective_r_cone}
2 \cdot \gamma_m^k \cdot \Big( \zeta_m^{k+1} + \zeta_m^k \Big) \geq \sqrt{2}^2,
\end{equation}
which is denoted as
\begin{equation}
\label{eq:objective_r_cone_q}
(\gamma_m^k, \zeta_m^{k+1} + \zeta_m^k, \sqrt{2}) \in Q_r^3.
\end{equation}
Also, Equ.~\ref{eq:slack_equ_3} can be written as a standard (non-rotated) quadratic cone:
\begin{equation}
\label{eq:objective_cones}
\big( \beta_m^k + 1 \big) ^2 \geq \big( \beta_m^k - 1 \big) ^2 + \big( 2 \cdot \zeta_m^k \big)^2, 
\end{equation}
and is denoted as 
\begin{equation}
\label{eq:objective_cones_q}
(\beta_m^k + 1, \beta_m^k - 1, 2\zeta_m^k) \in Q^3.
\end{equation}
Finally, a slack variable $s$ is introduced to transform the objective in Equ.~\ref{eq:objective_equ} to an affine function:
\begin{equation}
\label{eq:objective_equ_2}
 \sum_{m = 1}^{N} \sum_{k = 0}^{K_i-1} (2 \cdot \gamma_m^k + \rho \cdot s) \cdot \delta t,
\end{equation}
with a rotated quadratic cone:
\begin{equation}
\label{eq:objective_cone}
2 \cdot s \cdot 1 \geq \sum_{m=1}^{N}\sum_{k=0}^{K_i-1}(\alpha_m^k)^2,
\end{equation}
i.e. 
\begin{equation}
\label{eq:objective_cone_q_r}
(s, 1, \boldsymbol{\alpha}) \in Q_r^{2+\sum_{m=1}^{N}(K_i)},
\end{equation}
where $\boldsymbol{\alpha}$ contains $\alpha_m^k$ in all pieces of the trajectory.

Also, the discretization is applied to $\alpha_m(t)$ and $\beta_m(t)$ in constraints listed in Sect.~\ref{subsubsec:temporal_constraints}. 
Details are omitted for brevity. 
After that, we re-formulate these constraints as affine equality and in-equality functions. 
Besides, although we assume $\alpha_k$ is piecewise constant, we bound the changing rate of $\alpha_k$ considering the response time of the actuators of our quadrotor. 
We also write this changing rate constraint in an affine form:
\begin{equation}
\label{eq:a_d_constraints}
-\delta \alpha \leq (\alpha_m^k - \alpha_m^{k-1}) / \delta t \leq -\delta \alpha,
\end{equation}
where $\delta \alpha$ (not jerk) is a pre-defined bound of the changing rate of acceleration. 
Since the difference of $\tau$ between $t_m^k$ and $t_m^{k-1}$ cannot be determined during the optimization, we only bound the changing rate of $\alpha_k$ in $t$ domain. 

The temporal optimization problem in Sect.~\ref{subsec:minimum_time_form} is formulated as a standard Second Order Cone Program (SOCP):
\begin{align}
\label{eq:socp_final_form}
\text{min}  \:\:\:\:\:\: & \mathbf{h}^T \boldsymbol{\gamma} + \rho \cdot s, \nonumber \\
\text{s.t.} \:\:\:\:\:\: & \mathbf{A}_{eq} \cdot \mathbf{x} = \mathbf{b}_{eq}, \nonumber \\
& \mathbf{A}_{ie} \cdot \mathbf{x} \leq \mathbf{b}_{ie}, \nonumber \\
&(s, 1, \boldsymbol{\alpha}) \in Q_r^{2+\sum_{m=1}^{N}(K_i)},  m = 1, ... N.  \\
&(\gamma_m^k, \zeta_m^{k+1} + \zeta_m^k, \sqrt{2}) \in Q_r^3, k = 0,..., K_i-1,\nonumber \\
&(\beta_m^k + 1, \beta_m^k - 1, 2\zeta_m^k) \in Q^3, k = 0,..., K_i. \nonumber
\end{align}

Here $\boldsymbol{\gamma}$ and $\mathbf{x}$ consist of all $\gamma^k$ and $\alpha^k, \beta^k, \zeta^k, \gamma^k$. 
$\delta t$ is the resolution of discretization of the problem. 
The effect of different $\rho$ and $\delta t$ to the temporal trajectory and a more detailed derivation of the SOCP can be viewed in~\cite{fei2018iros}.

In our \textit{teach-repeat-replan} system, since the global repeating trajectory always has static initial and final states, Equ.~\ref{eq:socp_final_form} is always mathematically feasible regardless of the solution of spatial optimization.
Because a feasible solution of the optimization program can always be found by infinitely enlarging the time. 
Combined with the fact that the spatial optimization also always has a solution (Sect.~\ref{subsec:space_optimization}), once a flight corridor is given, a spatial-temporal trajectory must exist.

\section{Online Local Re-planning}
\label{sec:online_local_replanning}
\begin{figure}[t]
\centering
\includegraphics[width=0.9\columnwidth]{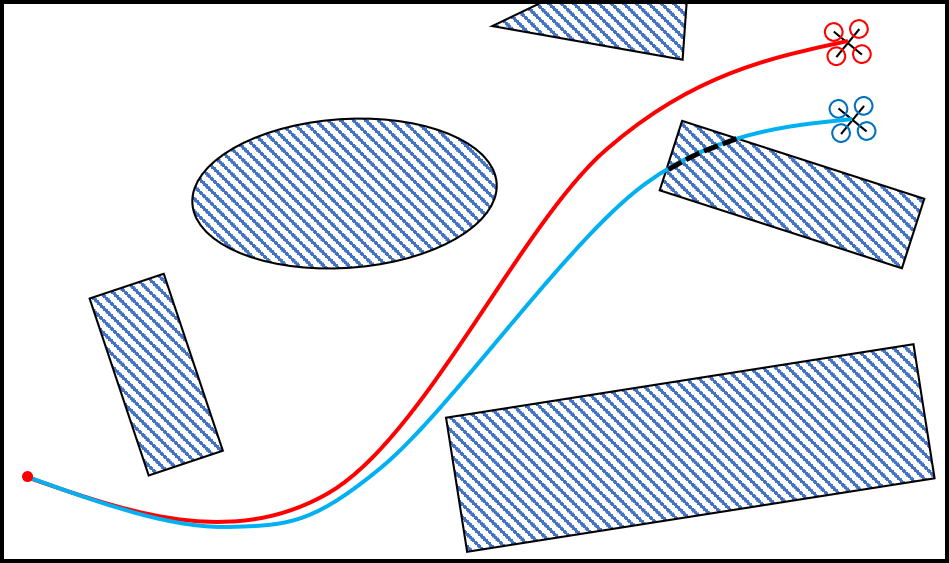}
\vspace{-0.25cm}
\caption{An illustration of colliding with obstacles when there are significant pose drifts but no timely loop closure corrections. 
Obstacles are depicted in the global frame. The flight path of the drone in the VIO frame is shown in the red curve. But the actual trajectory in the global frame is the blue curve, which collides with obstacles on the global map. 
\label{fig:drift_crash}
}
\vspace{-0.25cm}
\end{figure}

In our previous work~\cite{fei2019ral}, once the global planning finished, the drone would execute the trajectory without other considerations. 
This strategy is based on assumptions that 1) the map of the environment is perfectly built and remains intact; 2) globally consistent pose estimation is provided. 
We use a VIO system with loop closure to correct local pose drifts, and our dense map is globally deformed according to the global pose graph. 
However, the first assumption does not always hold, especially when new obstacles suddenly appear or the environment changes. 
As for the second assumption, our global pose estimation relies on the loop closure detection, which also does not guarantee an extremely high recall rate. 
In situations where there are significant pose drifts but no timely loop closure corrections, the drone may have collisions with obstacles, as in Fig.~\ref{fig:drift_crash}.

\subsection{Local Re-planning Framework}
To address above issues fundamentally, we propose a local re-planning framework which reactively wraps the global trajectory to avoid unmodeled obstacles. 
A sliding local map is maintained onboard, where obstacles are fused, and an ESDF (Euclidean Signed Distance Field)~\cite{felzenszwalb2012distance} is updated accordingly. 
Note that the dense global map is attached to the global pose graph but the local map introduced here is associated with the local VIO frame and sliding with the drone. 
\subsubsection{ESDF Mapping}
\begin{figure}[t]
\centering
\includegraphics[width=0.9\columnwidth]{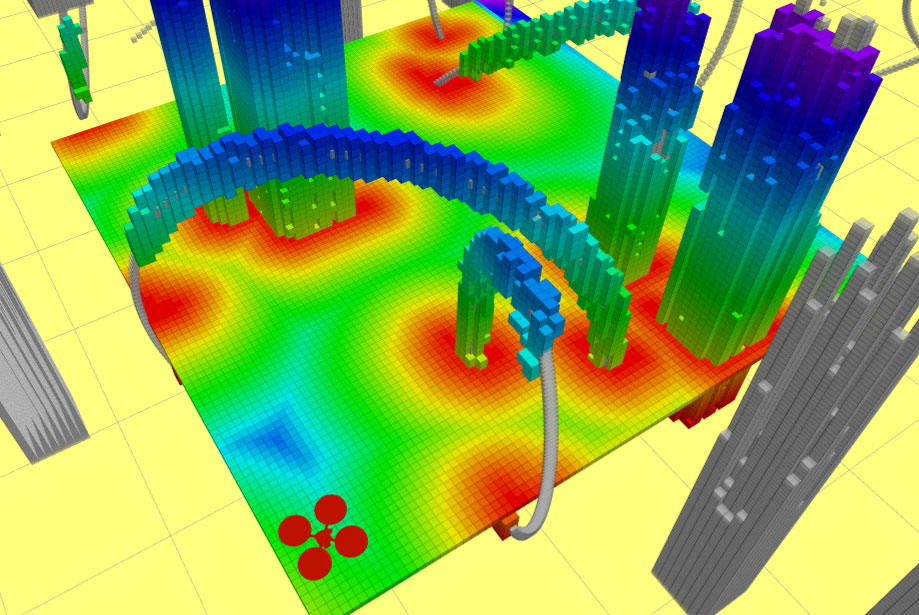}
\vspace{-0.25cm}
\caption{\label{fig:esdf_example}
The local occupancy map its corresponding ESDF map visualized at a given height of $0.6m$.
}
\vspace{-1.25cm}
\end{figure}

We adopt our previous work FIESTA~\cite{han2019fiesta}, which is an advanced incremental ESDF~\cite{schouten2010incremental} mapping framework, to build the local map for online re-planning. 
FIESTA fuses the depth information into a voxel-hashed occupancy map~\cite{klingensmith2015chisel} and updates the distance value of voxels as few as possible using a breadth-first search (BFS) framework. 
It is lightweight, efficient, and produces near-optimal results. Details can be checked in~\cite{han2019fiesta}. 
The ESDF is necessary for the following gradient-based trajectory wrapping. 
An example of a local occupancy map and its corresponding ESDF map are shown in Fig.~\ref{fig:esdf_example}. 
Note, in our system the range of the local map is decided by the range of current depth observation. 

\subsubsection{Sliding Window Re-planning}
\begin{figure*}[t]
\centering
  \subfigure[\label{fig:replan_traj1} ]
{\includegraphics[width=0.99\columnwidth]{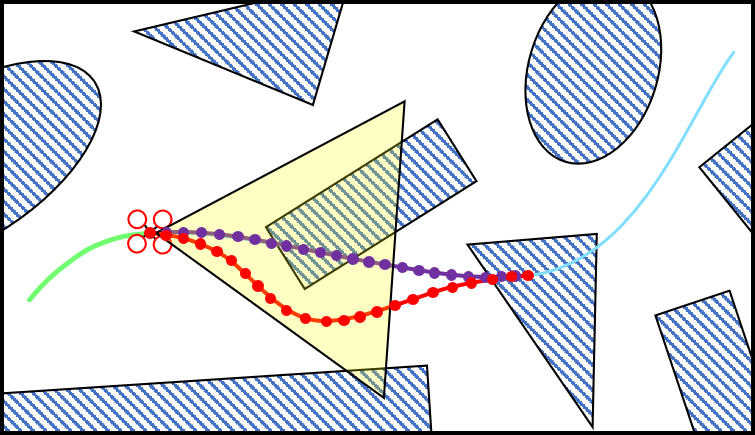}}
  \subfigure[\label{fig:replan_traj2} ]
{\includegraphics[width=0.99\columnwidth]{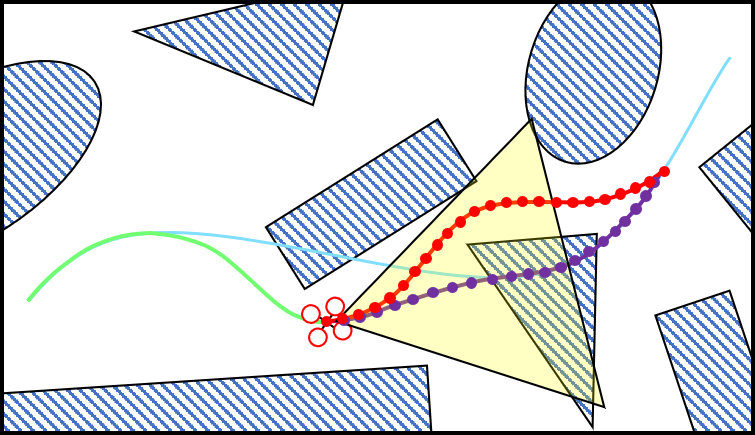}}
\vspace{-0.25cm}
\caption{An illustration of the online re-planning mechanism. 
The blue and green curves are the global trajectory and the actual flight path of the drone, respectively. 
The purple curve and dots are the global trajectory in the sliding window and its corresponding control points. 
The red curve and dots are the re-planned local trajectory and its control points. 
The yellow frustum shows the sensing horizon of the drone. 
\label{fig:replan_traj}}
\vspace{-0.5cm}
\end{figure*}

Due to the limited onboard sensing range and computing resources, it is impossible and unnecessary to conduct global re-planning. 
In this article, we maintain a temporal sliding window over the global trajectory and conduct local re-planning within it.
As is shown in Fig.~\ref{fig:replan_traj}, when obstacles are observed to block the trajectory in the sliding window, a re-planed trajectory is generated to avoid obstacles, and rejoin the global trajectory afterward. 

\subsection{Gradient-Based B-spline Optimization}
\subsubsection{B-spline Trajectory Formulation}
A B-spline is a piecewise polynomial function defined by a series of control points $ \{ \mathbf{Q}_{0},\mathbf{Q}_{1}, \cdots, \mathbf{Q}_{N}  \} $ and knot vector $ [ t_{0}, t_{1}, \cdots, t_{m} ] $. For a $p$-degree B-spline, we have $ m = N+p+1 $. 
Following the matrix representation of the De Boor–Cox formula~\cite{de1971subroutine}, the value of a B-spline can be evaluated as:
\begin{equation}
\label{equ:matrix}
	P(u) = 
	\left[
		1, u, \dots, u^{p} 
	\right] 
	\cdot
	\mathbf{M}_{p+1}
	\cdot
	\left[
		\mathbf{Q}_{i-p}, \mathbf{Q}_{i-p+1}, \dots, \mathbf{Q}_{i}
	\right]^T
\end{equation}
here $ \mathbf{M}_{p+1} $ is a constant matrix depends only on $ p $. And $ u = (t - t_{i})/ (t_{i+1} - t_{i})$, for $ t \in [t_{i}, t_{i+1}) $.

\subsubsection{B-spline Initialization}
We initialize the local trajectory optimization by re-parameterizing the trajectory in the re-planning horizon as a uniform B-spline. The reason we use uniform B-spline is that it has a simple mathematical formula that is easy to evaluate in the following optimization.
For a uniform B-splines, each knot span $ \Delta t_{i} = t_{i+1} - t_{i} $ has identical value $ \Delta t $.  
The local trajectory is first discretized to a set of points according to a given $\Delta t$. 
Then these points are fitted to a uniform B-spline by solving a min-least square problem. 

Note that, a $ p $ degree uniform B-spline is naturally $ p-1 $ order continuous between consecutive spans. 
Therefore, there is no need to enforce continuity constraints in the following optimization explicitly. 
Besides, for a $ p $ degree B-spline trajectory defined by $ N+1 $ control points, the first and last $ p $ control points are fixed due to the continuous requirement of the starting and ending states of the local trajectory. 

\subsubsection{Elastic Band Optimization}
The basic requirements of the re-planed B-spline are three-folds: smoothness, safety, and dynamical feasibility. 
We define the smoothness cost $ J_{s} $ using a jerk-penalized elastic band cost function\cite{quinlan1993elastic,zhu2015convex}:
\begin{align}\label{equ:elastic_cost}
& J_{s} = \nonumber \\ 
	&\sum\limits_{i=1}^{N-1} \Vert \underbrace{(\mathbf{Q}_{i+2}-2\mathbf{Q}_{i+1}+\mathbf{Q}_{i})}_{\mathbf{F}_{i+1,i}} \ - \ \underbrace{(\mathbf{Q}_{i+1}-2\mathbf{Q}_{i}+\mathbf{Q}_{i-1})}_{\mathbf{F}_{i-1,i}} \Vert^{2}  \nonumber \\
	&= \sum\limits_{i=1}^{N-1} \Vert \mathbf{Q}_{i+2}-3\mathbf{Q}_{i+1}+3\mathbf{Q}_{i}-\mathbf{Q}_{i-1} \Vert^{2},
\end{align}
which can be viewed as a sum of the squared jerk of control points on the B-spline. 
Note here we use this formulation which is independent of the time parametrization of the trajectory instead of the traditional time integrated cost function~\cite{MelKum1105}. 
Because the time duration in each span of the B-spline may be adjusted after the optimization (Sec.~\ref{subsubsec:time_adjust}), Eq.~\ref{equ:elastic_cost} captures the geometric shape of the B-spline regardless of the time parametrization.
Besides, Eq.~\ref{equ:elastic_cost} bypasses the costly evaluation of the integration and is, therefore, more numerically robust and computationally efficient in optimization.

The safety and dynamical feasibility requirements of the B-spline are enforced as soft constraints and added to the cost function. 
Also, the collision cost $J_c$, dynamical feasibility cost $J_v$, and $J_a$ are evaluated at only control points. 
The collision cost $J_c$ is formulated as the accumulated L2-penalized closest distance to obstacles along the trajectory, which is written as
\begin{equation}\label{equ:colli}
	J_{c} = \sum\limits_{i=p}^{N-p} F_{c}(d(\mathbf{Q}_{i})),
\end{equation}
where $d(\mathbf{Q}_{i})$ is the distance between $ \mathbf{Q}_{i} $ to its closet obstacle and is recorded in the ESDF. $ F_{c} $ is defined as
\begin{equation}\label{equ:potential}
	F_{c}(d) = \left\{ 
	\begin{array}{ccl}
	(d-d_{0})^{2} & & d \le d_{0} \\
	0 & & d > d_{0}
	\end{array}
	 \right.
\end{equation}
where $ d_{0} $ is the expected path clearance. $J_v$ and $J_a$ are applied to velocities and accelerations, which exceed the physical limits. The formulations of $J_v$ and $J_a$ are similar to Eq.~\ref{equ:colli} and are omitted here.
The overall cost function is:
\begin{equation}\label{equ:cost}
    J_{total} = \lambda_{1} J_{s} + \lambda_{2} J_{c} + \lambda_{3} (J_{v} + J_{a}),
\end{equation}
where $\lambda_{1}, \lambda_{2}, \lambda_{3}$ are weighting coefficients. $J_{total}$ can be minimized for a local optimal solution by general optimization methods such as Gauss-Newton or Levenberg-Marquardt.

\subsubsection{Iterative Refinement}
\label{subsubsec:time_adjust}
In the above-unconstrained optimization problem, although collisions and dynamical infeasibilities are penalized, there is no hard guarantee on generating a strictly feasible solution. 
To improve the success rate in practice, we add a post-process to refine the trajectory iteratively. 
In each iteration, we check collisions and feasibilities of all optimized control points. 
If collisions are detected, we increase the collision term $J_c$ by increasing $\lambda_2$ and solve the optimization problem (Eq.~\ref{equ:cost}) again. 

Since we wrap the local trajectory to go around obstacles, the trajectory is always lengthened after the optimization. 
Consequently, using the original time parametrization will unavoidably result in a higher aggressiveness, which means the quadrotor tends to fly faster. 
Then its velocity and acceleration would easily exceed the predefined limits. 
Therefore, we adjust the time parameterization of the local trajectory to squeeze out dynamical infeasibilities.
We slightly enlarge infeasible knots spans of the B-spline by the following heuristic.
\begin{equation}\label{equ:adj2}
\Delta t_{i}^{'} = \min \{ \alpha, \max \{\frac{v_{m}}{v_{max}}, (\frac{a_{m}}{a_{max}})^{\frac{1}{2}}\}\} \cdot \Delta t_{i},
\end{equation}
where $ \alpha $ is a constant slightly larger than $ 1 $. $ v_{m}, a_{m} $ are infeasible velocity and acceleration and $ v_{max}, a_{max}$ are maximum allowed acceleration and velocity of the drone.
The time duration is iteratively enlarged until obtaining a feasible solution or exceeding the maximum iteration limit. 
If no feasible solution exists after the time adjustment, $\lambda_3$ is increased, and the trajectory is optimized again.

\section{Results}
\label{sec:results}

\subsection{Implementation Details}
\label{subsec:implementation_details}
The global planning method proposed in this paper is implemented with a QP solver OOQP\footnote{\url{http://pages.cs.wisc.edu/~swright/ooqp/}} and a SOCP solver Mosek\footnote{\url{https://www.mosek.com}}. 
The local re-planning depends on a nonlinear optimization solver NLopt\footnote{\url{https://nlopt.readthedocs.io}}.
The source code of all modules in our quadrotor system, including local/global localization, mapping, and planning, are released as ros-packages\footnote{\url{https://github.com/HKUST-Aerial-Robotics/Teach-Repeat-Replan}} for the reference of community.
Readers of this paper can easily replicate all the presented results.
The state estimation, pose graph optimization, local mapping, local re-planning, and the controller is running onboard on a Manifold-2C\footnote{\url{https://store.dji.com/product/manifold-2?vid=80932}} mini-computer.
Other modules are running on an off-board laptop which has a GTX 1080\footnote{\url{https://www.nvidia.com/en-us/geforce/20-series/}} graphics card.

Our global map is built to attach to a global pose graph. 
Both the map and the pose graph are saved for repeating.
Before the repeating, the drone is handheld to close the loop of the current VIO frame with the saved global pose graph.
The relative transformation of these two frames is used to project the control commands to the VIO frame.    
During the repeating, pose graph optimization is also activated to calculate the pose drift and compensate for the control command.

\begin{figure}[t]
\begin{center}          
\subfigure[\label{fig:simu1} The flight corridor catpures local free space.]
{\includegraphics[width=0.8\columnwidth]{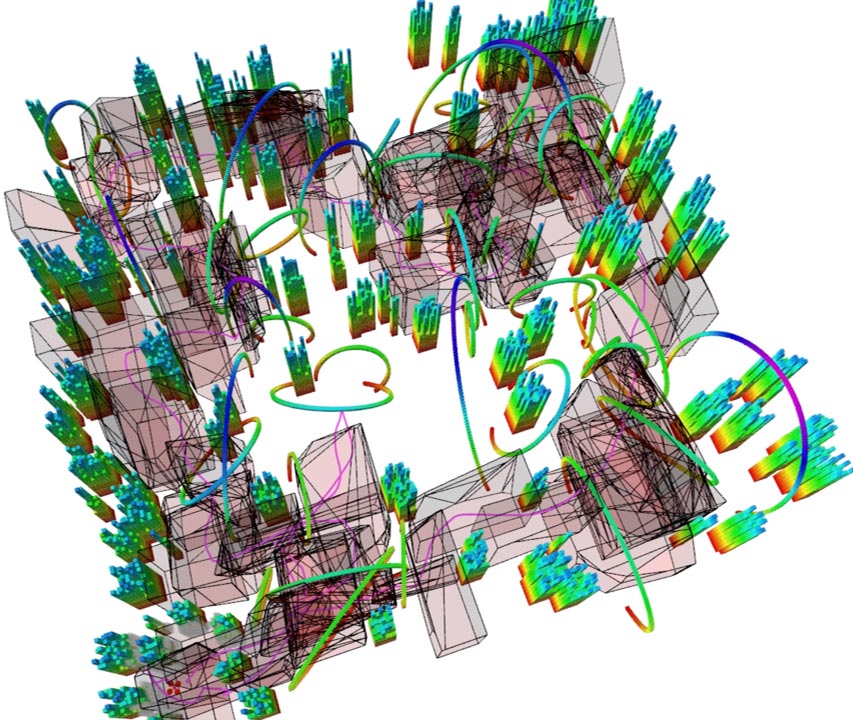}}     
\subfigure[\label{fig:simu6} The local re-planning trajectory and current depth image.]
{\includegraphics[width=0.8\columnwidth]{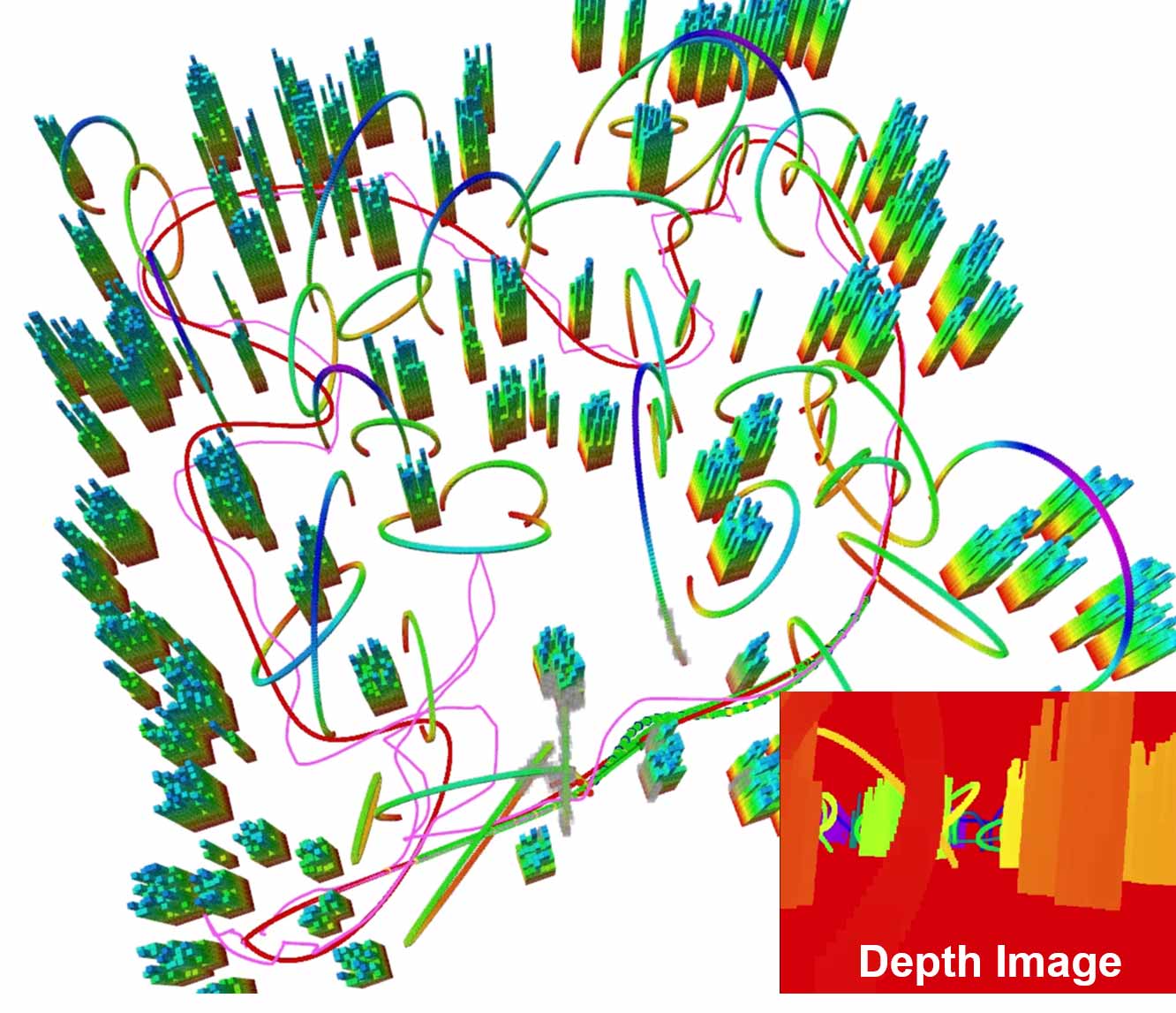}}     
\end{center}
\vspace{-0.25cm}
\caption{\label{fig:simulation_rviz} 
The trajectory generated in a complex simulated environment. The flight corridor consists of large free convex polyhedrons are shown in (a), and the optimized space-time trajectory is shown in (b).
}
\vspace{-0.75cm}
\end{figure} 

\subsection{Simulated Flight Test}
\label{subsec:simulation}
We first test our global and local planning methods in simulations. 
The simulated environments are randomly deployed with various types of obstacles and circles for drone racing, as shown in Fig.~\ref{fig:simulation_rviz}. 
The simulating tool we use is a light-weight simulator MockaFly\footnote{\url{https://github.com/HKUST-Aerial-Robotics/mockasimulator}}, which contains quadrotor dynamics model, controller, and map generator. 
And the simulator's is also released as an open-source package with this paper. 
In the simulation, a drone is controlled by a joystick to demonstrate the teaching trajectory. 
The simulated drone is equipped with a depth camera whose depth measurements are real-time rendered in GPU by back-projecting the drone's surrounding obstacles. 
We randomly add noise on the depth measurements to mimic a real sensor. The re-planning module is activated in the simulation and is triggered by the noise added on the depth. 
The teaching trajectory and the flight corridor is shown in Fig.~\ref{fig:simu1}.
The global trajectory, local re-planned trajectory, and depth measurement are shown in Fig.~\ref{fig:simu6}.
More details about the simulation are presented in the attached video.

\subsection{Benchmark Comparisons}
\label{subsec:benchmark}

\subsubsection{Corridor Generation}
\label{subsubsec:compare_corridor}
We test the performance of the flight corridor generation methods (Sect.~\ref{sec:corridor_generation}), to show the efficacy of the proposed techniques for CPU (Sect.~\ref{subsec:cpu_acceleration}) and GPU (Sect.~\ref{subsec:gpu_acceleration}) accelerations. 
For convenience, we denote the basic process for doing \textit{convex cluster inflation} as CPU\_raw; CPU\_raw added cube initialization as CPU+; the one with cube initialization, vertex selection and early termination as CPU++; and the parallel version of the \textit{convex cluster inflation} as GPU. 
We first compare the time consumed for finding the largest flight corridor with these methods, to validate the improvements of efficiency by using our proposed CPU and GPU acceleration techniques.
Then, we compare the ratio of space capturing by methods with and without the polyhedron initialization, and by our previous method~\cite{fei2019ral}.
The motivation of the latter comparison is two-fold:
\begin{enumerate}
\item It serves to show superior performance by replacing cubes with polyhedrons.
\item As discussed in Sect.~\ref{subsec:cpu_acceleration}, the initialization process would result in different final clustering results compared to the pure \textit{convex cluster inflation}. 
This comparison also validates that the initialization process only makes neglectable harm to free space capturing.
\end{enumerate}
We generate 10 random maps, with 10 $\sim$ 20 random teaching trajectories given in each map. 
The average length of teaching trajectories is $20 m$. 
Results are given in Tabs.~\ref{tab:benchmark_compare_corridor_time} and~\ref{tab:benchmark_compare_corridor_space}.

\begin{figure}[t]
\begin{center}          
\subfigure[\label{fig:compare_corridor_1} ]
{\includegraphics[width=0.49\columnwidth]{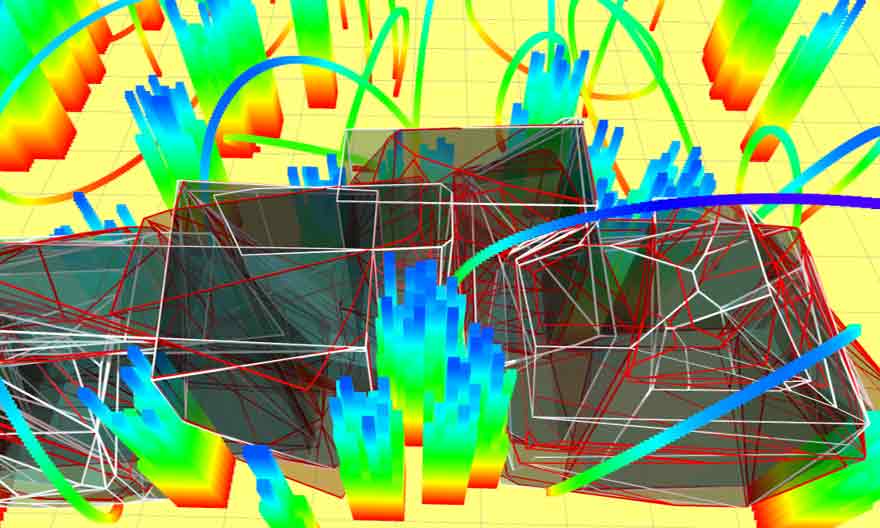}}     
\subfigure[\label{fig:compare_corridor_2} ]
{\includegraphics[width=0.49\columnwidth]{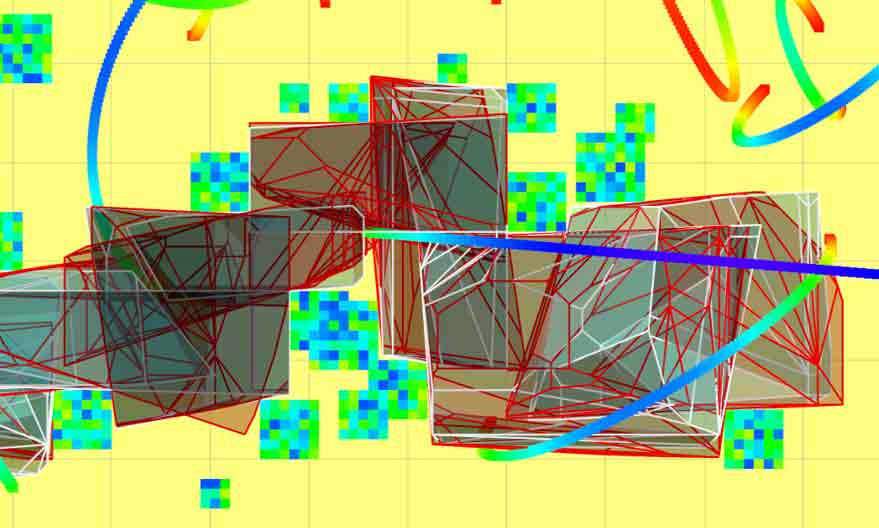}}     
\end{center}
\vspace{-0.25cm}
\caption{\label{fig:compare_corridor_rviz} 
The flight corridor generated with and without the initialization process. Polyhedrons with bounding edges in white and red are found by methods with and without the initialization, respectively.
}
\vspace{-0.25cm}
\end{figure} 

\begin{table}[t]
\begin{center}
\caption{Comparison of Computing Time of Corridor Generation}
\label{tab:benchmark_compare_corridor_time}
\begin{tabular}{|c|c|c|c|c|}
\hline
\multirow{2}{*}{Computing Time (s)} & \multirow{2}{*}{GPU} & \multirow{2}{*}{CPU++} & \multirow{2}{*}{CPU+} & \multirow{2}{*}{CPU\_raw} 
\\ &&&& \\
\hline
Res. = $0.25m$\:\: & 0.031& 0.111& 0.162& 0.359\\
\cline{1-5}
Res. = $0.20m$\:\: & 0.055& 0.310& 0.503& 1.309\\
\cline {1-5}
Res. = $0.15m$\:\: & 0.169& 1.423& 2.803& 9.583\\
\cline{1-5}
Res. = $0.10m$\:\: & 0.942 & 13.940 & 30.747 & 141.659 \\
\cline{1-5}
Res. = $0.075m$ &3.660  & 71.862 &157.181 &927.131 \\
\cline{1-5}
\end{tabular}
\end{center}
\vspace{-1.25cm}
\end{table}

As shown in Tab.~\ref{tab:benchmark_compare_corridor_time}, as the resolution of the map being finer, the computing time of the simple \textit{convex cluster inflation} quickly becomes unacceptable huge. 
In CPU, with the help of \textit{polyhedron initialization}, the computational efficiency is improved several times. 
Moreover, according to Tab.~\ref{tab:benchmark_compare_corridor_time}, introducing the \textit{voxel selection} and \textit{early termination} can increase the speed more than one order of magnitude in a fine resolution.
The efficacy of the GPU acceleration is even more significant.
As shown in Tab.~\ref{tab:benchmark_compare_corridor_time}, the GPU version improves the computing speed 30 times at a fine resolution ($0.075m$), and 10 times at a coarse resolution ($0.25m$). 
For a finer resolution, more candidate voxels are discovered in one iteration of Alg.~\ref{alg:parallel_polyhedron_inflation}, thus more computations are conducted parallelly to save time.

\begin{table}[t]
	\centering
	\caption{Comparison of Space Captured of Corridor Generation}
\label{tab:benchmark_compare_corridor_space}
	\begin{tabular}{|c|c|c|c|} 
	\hline
	\multirow{2}{*}{Space Ratio (\%)} & \multirow{2}{*}{w/ Initialization} & \multirow{2}{*}{ w/o Initialization} & \multirow{2}{*}{Previous~\cite{fei2019ral}} \\  &&&  \\
	\cline{2-4}
	\hline
	Res. = $0.25m$\:\:      & 99.22 & 100.00 & 82.28\\
	\hline
	Res. = $0.20m$\:\:      & 99.56 & 100.00 & 82.92\\
	\hline
	Res. = $0.15m$\:\:      & 98.93 & 100.00 & 81.82 \\
	\hline
	Res. = $0.10m$\:\:      & 97.06 & 100.00 & 82.78  \\
	\hline
	Res. = $0.075m$         & 97.14 & 100.00 & 83.03 \\ 
	\hline	
	\end{tabular}
	\vspace{-0.3cm}
\end{table}

For the second comparison, we count the number of free voxels included in the flight corridor found by each method. 
At each resolution, we take the result of the method without initialization as 100\% and compare others against it. 
Tab.~\ref{tab:benchmark_compare_corridor_space} indicates two conclusions:
\begin{enumerate}
\item Using polyhedrons instead of axis-aligned cubes can significantly increase the volume of the flight corridor.
\item Using initialization only slightly sacrifices the volume of the flight corridor. 
And the sacrifice is neglectable in a medium or coarse resolution ($0.15 \sim 0.25 m$).
\end{enumerate}  
The first conclusion holds because a simple cube only discovers free space in $x, y, z$ directions and sacrifices much space in a highly nonconvex environment, as in Fig.~\ref{fig:cube_polygon_compare}.
The second conclusion comes from the fact that in a highly nonconvex environment, a regular shaped polyhedron (a cube) does not prevent the following voxel clustering in its nearby space.
It shows that the initialization plus the clustering refinement does not harm the volume of the final polyhedron, and is acceptable in practice, especially for a resolution not very fine.

\subsubsection{Global Planning}
\label{subsubsec:compare_planning}
\begin{figure}[t]
\centering
\includegraphics[width=0.99\columnwidth]{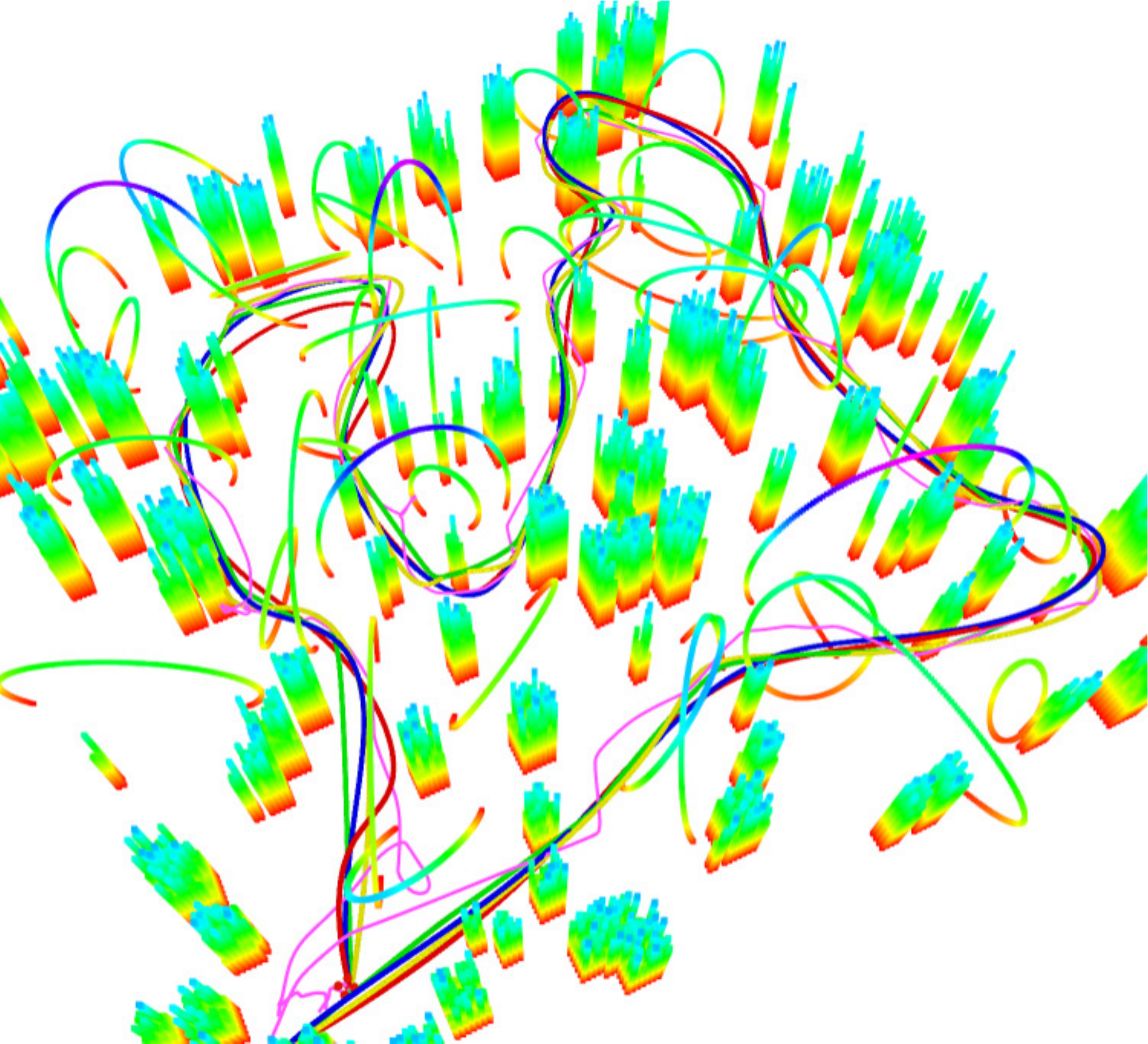}
\caption{
\label{fig:benchmark_compare_rviz}
The comparison of trajectories optimized by different methods. The manual flight trajectory is shown as the purple curve. Blue, red, green, and yellow trajectories are generated by our proposed method, our previous method~\cite{fei2019ral}, gradient-based method~\cite{fei2017iros} and waypoints-based method~\cite{RicBryRoy1312}.
}
\vspace{-1.25cm}
\end{figure}

\begin{table}[t]
\begin{center}
\caption{Comparison of Trajectory Optimization}
\label{tab:benchmark_compare}
\begin{tabular}{|c|c|c|c|}
\hline
\multirow{2}{*}{Method} & \multirow{2}{*}{Length (m)} & \multirow{2}{*}{Time (s)} & \multirow{2}{*}{Energy $((m/s^3)^2)$} 
\\ &&& \\
\hline
\textbf{Proposed Method} 		    &\textbf{84.607} &\textbf{55.154} &\textbf{83.350} \\
\cline{1-4}
Previous Method 		    	    &86.723 &57.736 &89.883 \\
\cline{1-4}
Gradient-based~\cite{fei2017iros}   &89.622 &111.398 &109.575 \\
\cline{1-4}
Waypoint-based~\cite{RicBryRoy1312} &97.045 &94.895 &204.267\\
\cline{1-4}
\end{tabular}
\end{center}
\vspace{-1.75cm}
\end{table}

We compare the proposed global planning method against our previous work~\cite{fei2019ral} and other representative optimization-based trajectory generation methods, such as the waypoint-based method~\cite{RicBryRoy1312} and the gradient-based method~\cite{fei2017iros}. 
For the latter two benchmarked methods, there is no explicit way to capture the topological structure of the teaching trajectory. 
Therefore, we convert the teaching trajectory to a piecewise path by recursively finding a collision-free straight line path along with it. 
Then we use this path to initialize the waypoint-based~\cite{RicBryRoy1312} method and the gradient-based method~\cite{fei2017iros}. 
Benchmarked methods are also integrated into the coordinate descent framework with temporal optimization. 
Some parameters dominate the performance of these benchmarked methods, especially for the gradient-based method~\cite{fei2017iros} where the trade-off between collision and smoothness is essential. 
For a fair comparison, parameters are tuned to achieve the best performances before the test.
We randomly generate 10 simulated environments with dense obstacles, as in Sect.~\ref{subsec:simulation}, and conduct 10 teach-and-repeat trials in each map. 
A sample result of generated trajectories is shown in Fig.~\ref{fig:benchmark_compare_rviz}. 

As shown in Tab.~\ref{tab:benchmark_compare}, our proposed method outperforms in all length, time, and energy aspects. 
The waypoint-based~\cite{RicBryRoy1312} method and the gradient-based~\cite{fei2017iros} method both require a piecewise linear path as initialization.
The waypoint-based~\cite{RicBryRoy1312} method can only add intermediate waypoints on the initial path. 
Therefore, it is mostly dominated by its initialization and tends to output a solution with low quality.
The gradient-based~\cite{fei2017iros} method has no such restriction and can adjust the path automatically by utilizing gradient information.
However, its optimization formulation is underlying non-convex, since the collision cost is defined on a non-convex ESDF. 
Therefore, the gradient-based~\cite{fei2017iros} method always finds a locally optimal solution around its initial guess.
Compared to these two methods, our method is initialization-free. 
Both the spatial and temporal optimization of our proposed method enjoys the convexity in its formulation.
They are guaranteed to find the global energy-optimal and time-optimal solutions in the flight corridor. 
Naturally, a smoother trajectory also tends to generate a faster time profile.
So finally, under the same coordinate descent framework, our method always outperforms~\cite{fei2017iros} and ~\cite{RicBryRoy1312}.
Compared to~\cite{fei2019ral}, the advanced corridor generation proposed in this paper (Sect.~\ref{sec:corridor_generation}) can always capture more free space than using our previous axis-aligned corridor. Naturally, it provides much more freedom for global planning and results in much better solutions. 

\begin{figure}[t]
\begin{center}          
\subfigure[\label{fig:indoor_map} The indoor testing environment.]
{\includegraphics[width=0.99\columnwidth]{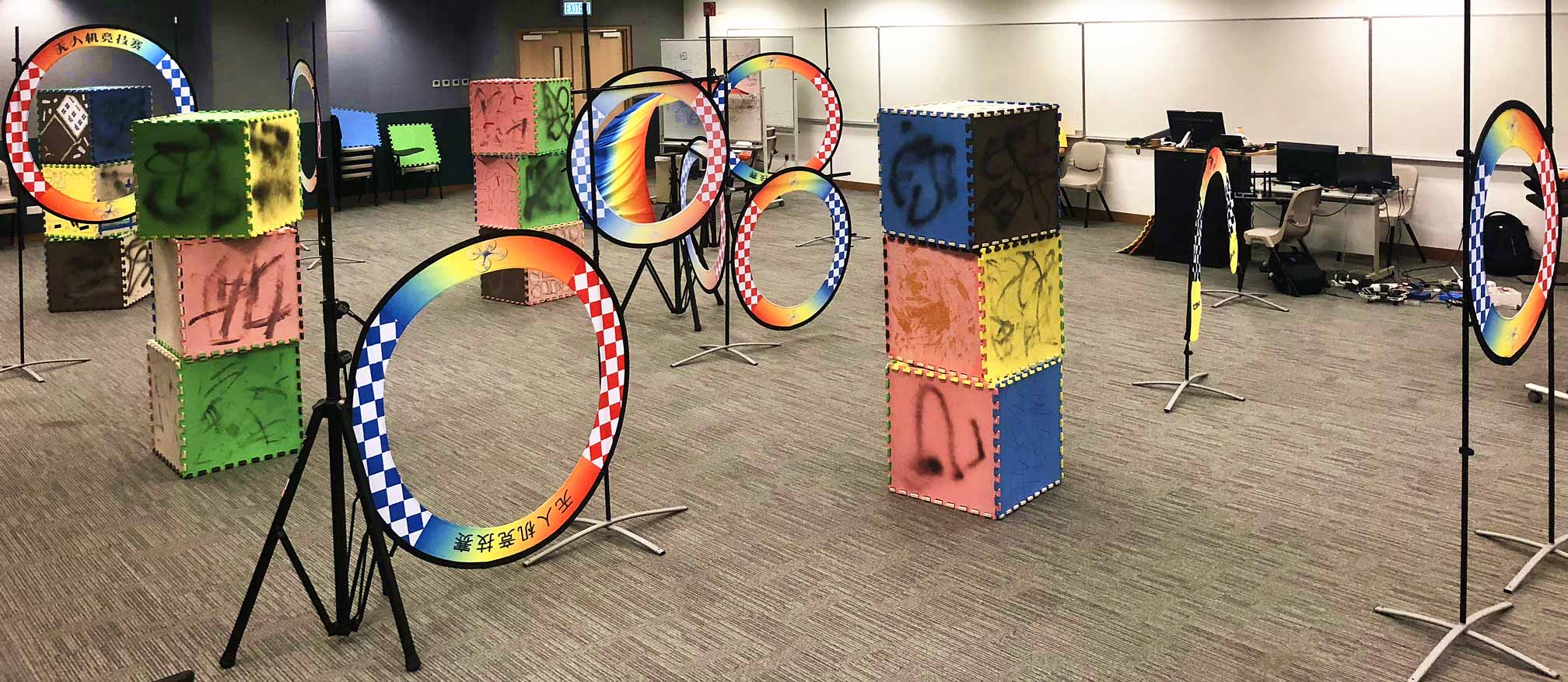}}
\subfigure[\label{fig:indoor_rviz_map} The global consistent dense map.]
{\includegraphics[width=0.99\columnwidth]{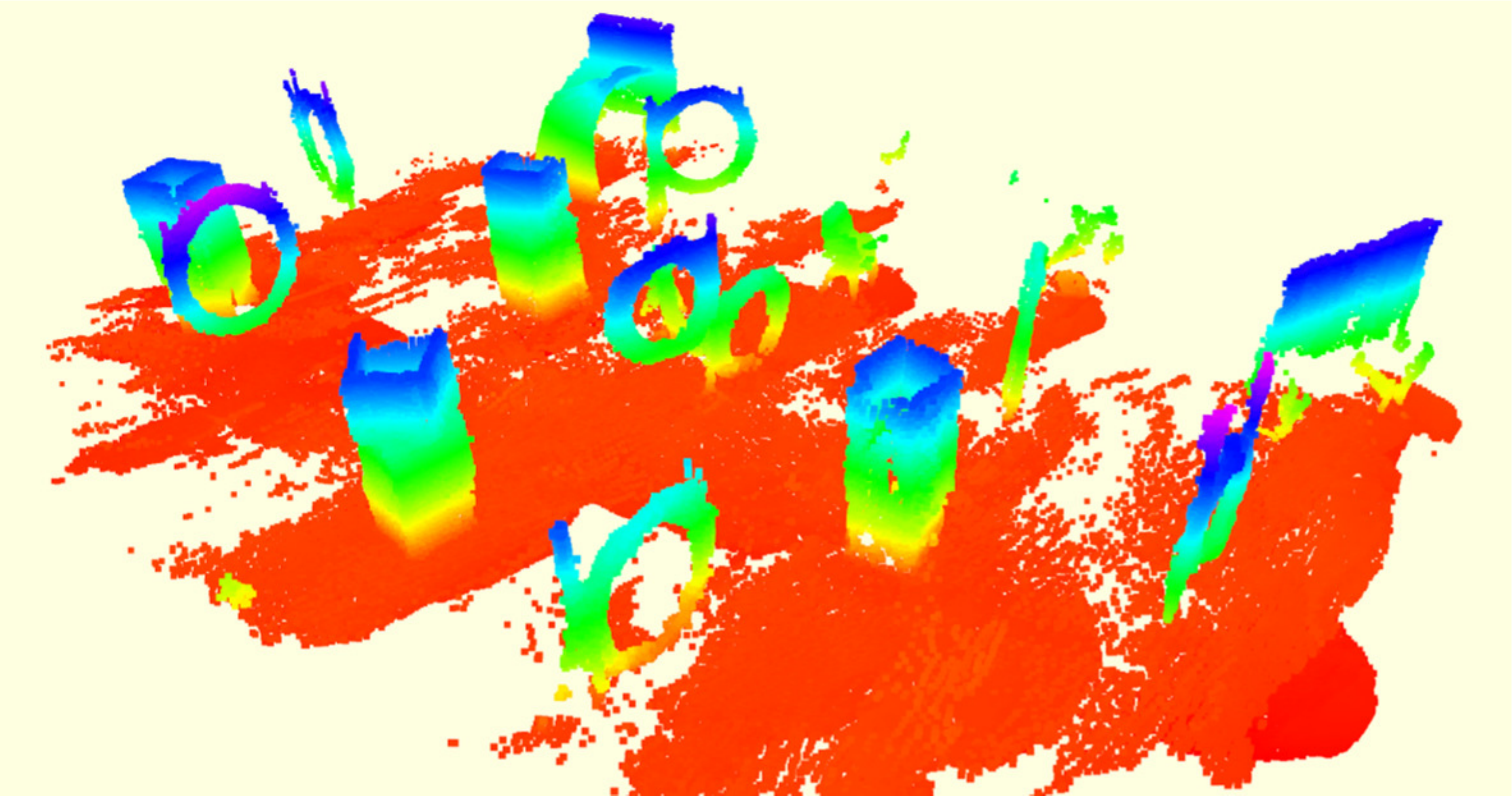}}
\end{center}
\caption{\label{fig:indoor_rviz} 
The experimental set-up of the fast indoor drone racing flights. (a), the obstacles deployment. (b), the pre-built globally consistent map. 
}
\vspace{-0.5cm}
\end{figure} 

\begin{figure}[t]
\centering
  \subfigure[\label{fig:indoor_exp_1}]
{\includegraphics[width=0.49\columnwidth]{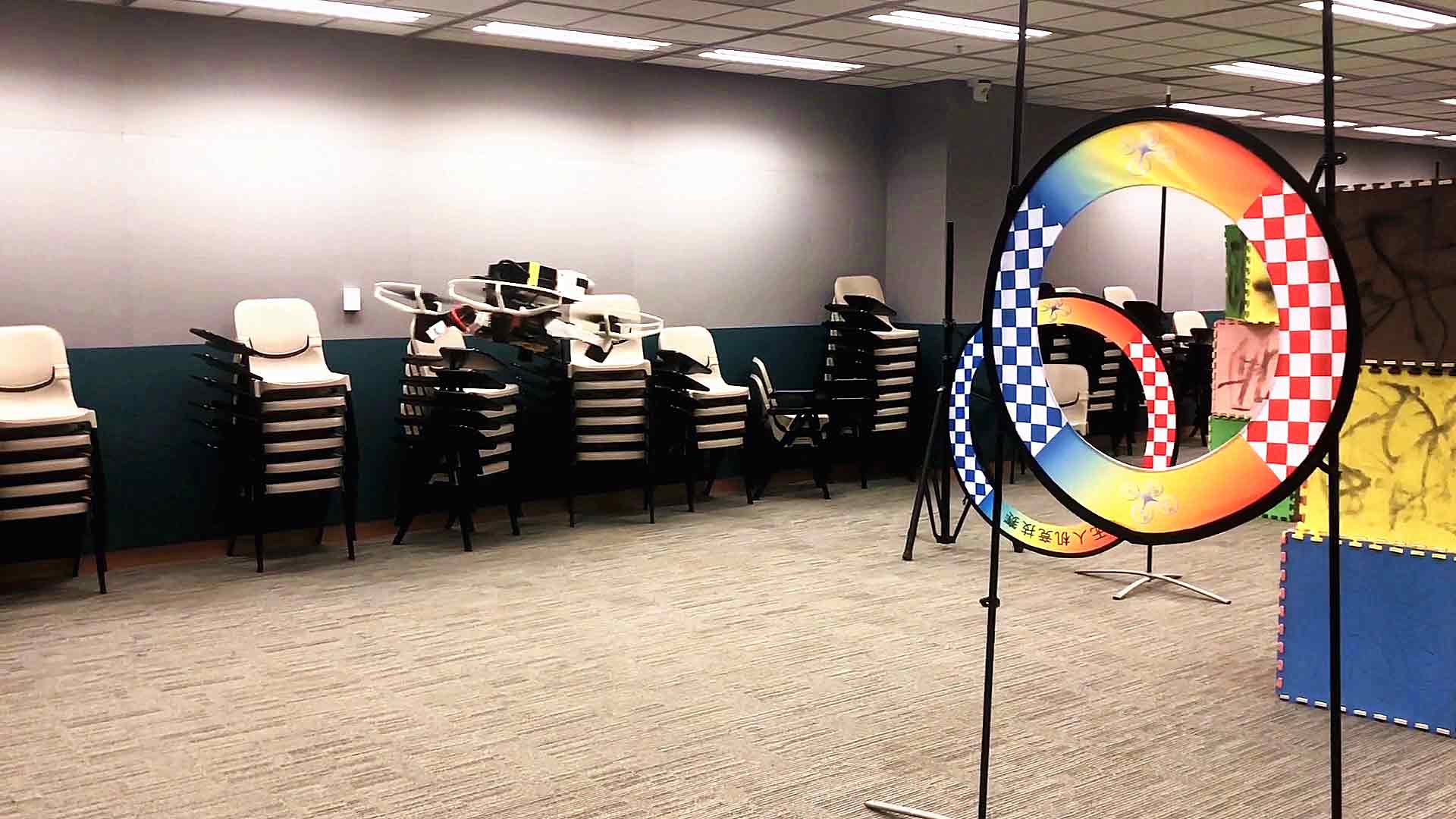}}
  \subfigure[\label{fig:indoor_exp_2}]
{\includegraphics[width=0.49\columnwidth]{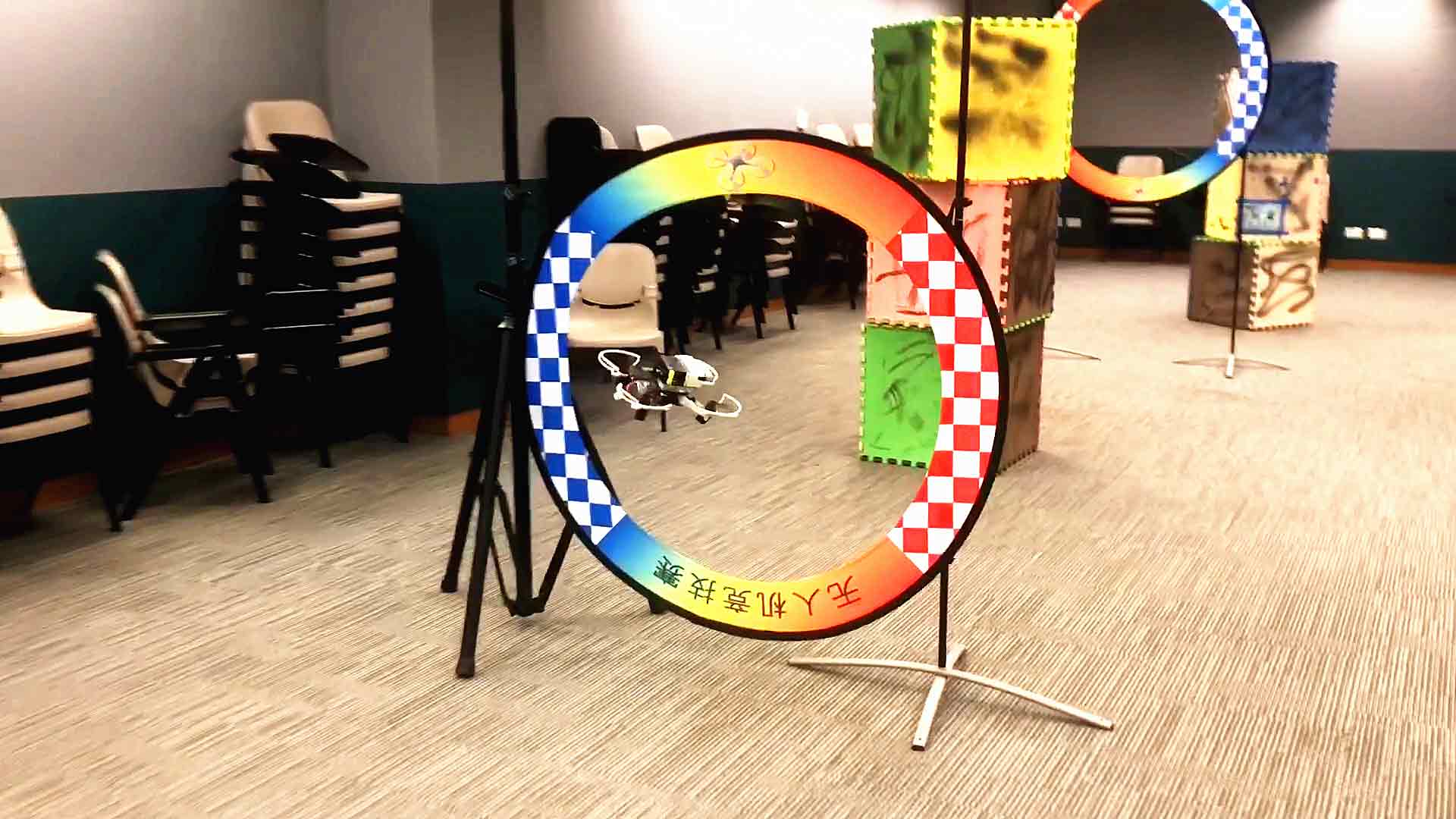}}
  \subfigure[\label{fig:indoor_exp_3}]
{\includegraphics[width=0.49\columnwidth]{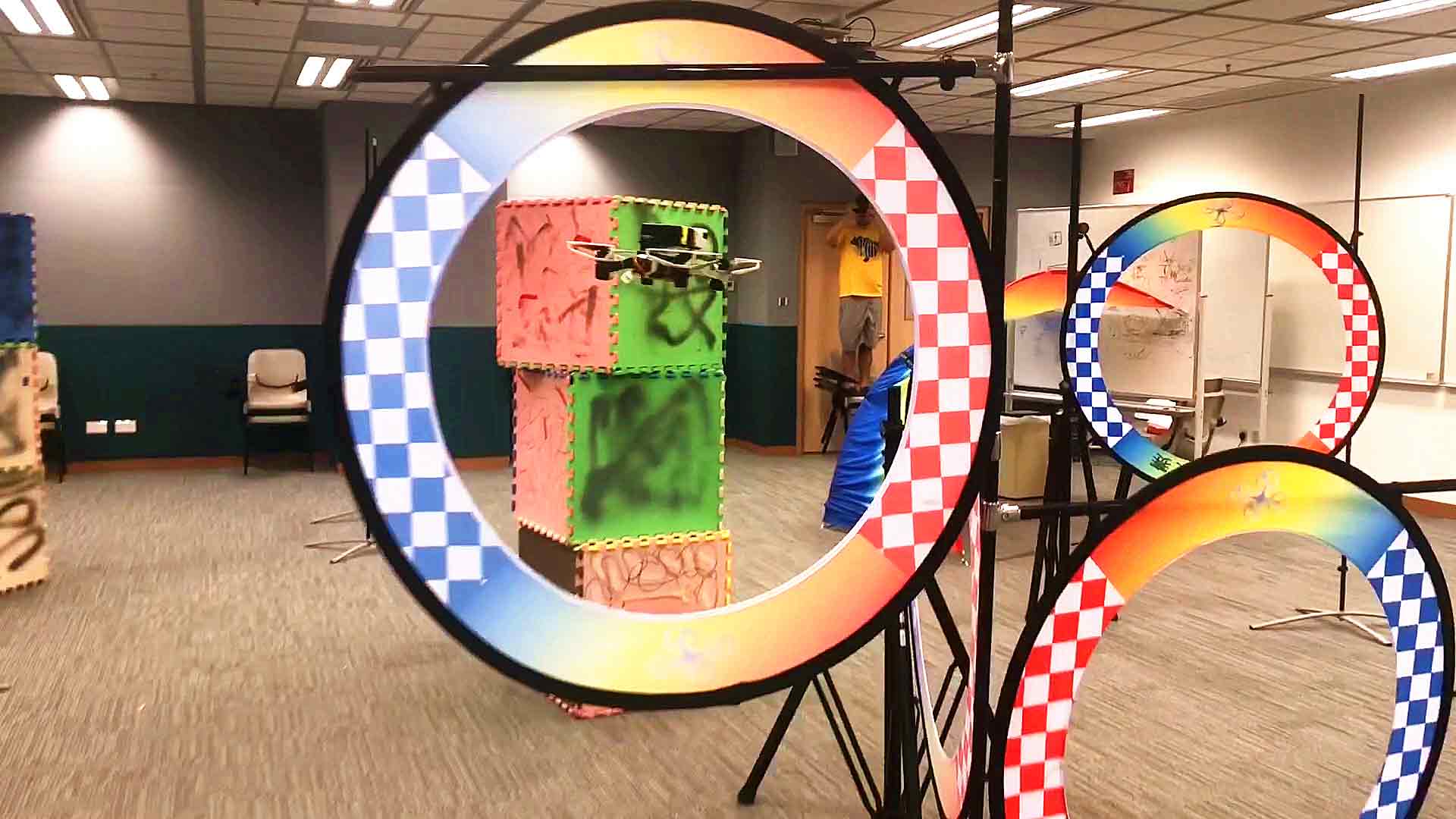}}
  \subfigure[\label{fig:indoor_exp_4}]
{\includegraphics[width=0.49\columnwidth]{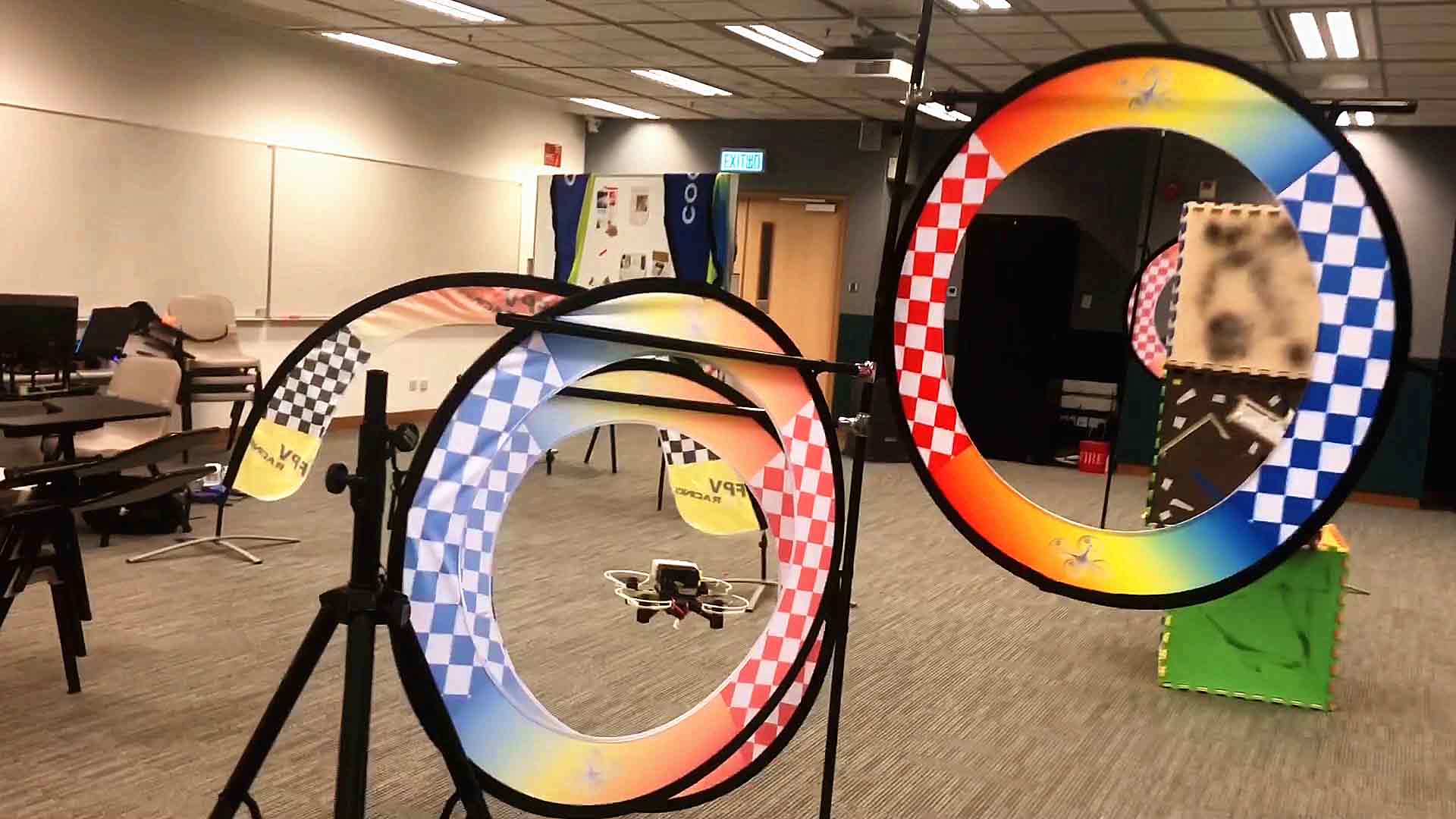}}
\caption{Snapshots of the fast autonomous flight in a static indoor environment. The maximum velocity and acceleration for the quadrotor are set as $3m/s$ and $3m/s^2$. 
\label{fig:indoor_exp}}
\end{figure}

\subsection{Indoor Flight Test}
\label{subsec:indoor_exp}
\subsubsection{Fast Flight in a Static Environment}
\begin{figure}[t]
\centering
\subfigure[\label{fig:indoor_teach_rviz_2} The teaching trajectory and the flight corridor.]
{\includegraphics[width=0.99\columnwidth]{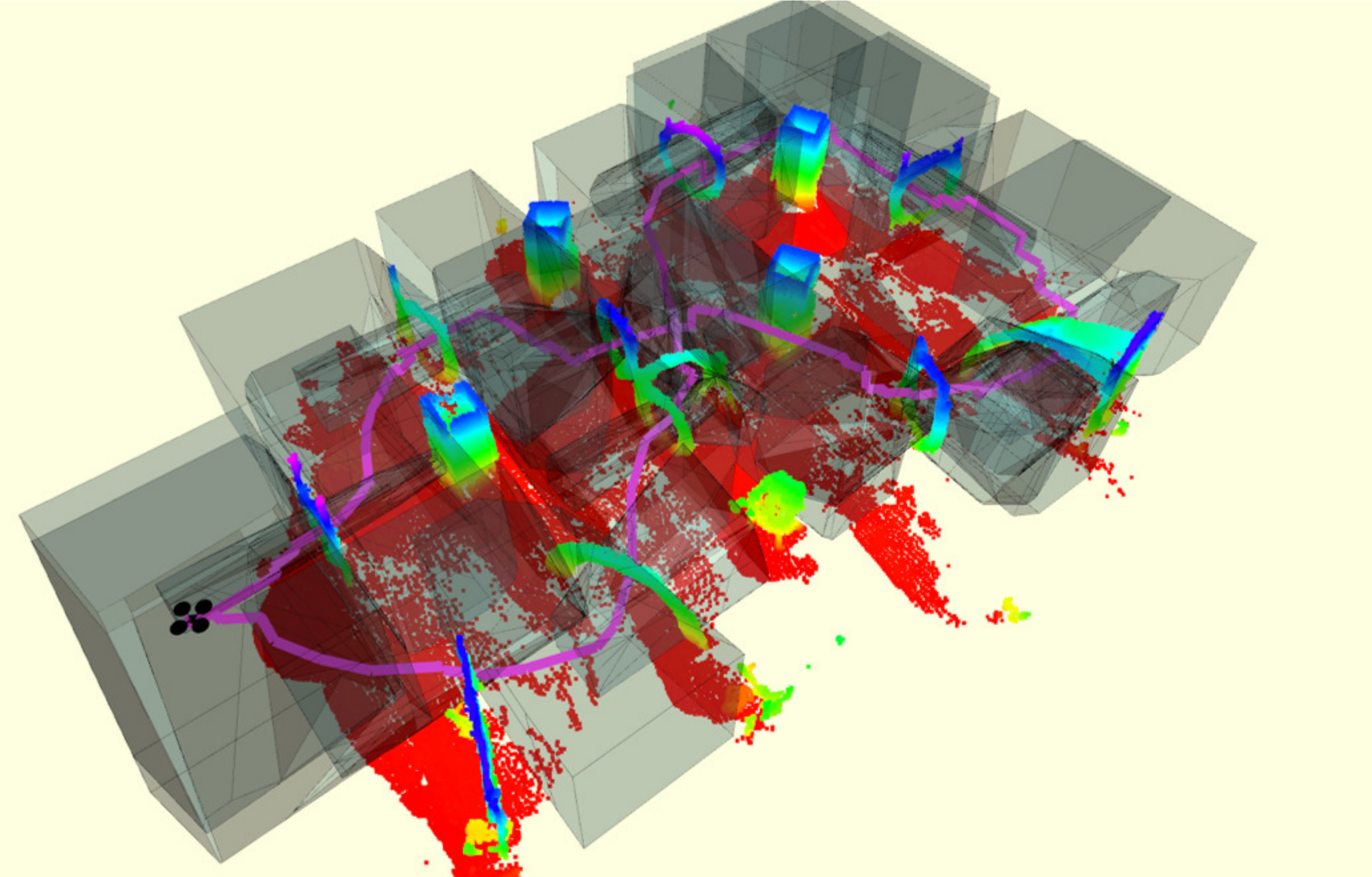}} 
\subfigure[\label{fig:indoor_repeat_rviz_2} The spatial-temporal optimal repeating trajectory.]
{\includegraphics[width=0.99\columnwidth]{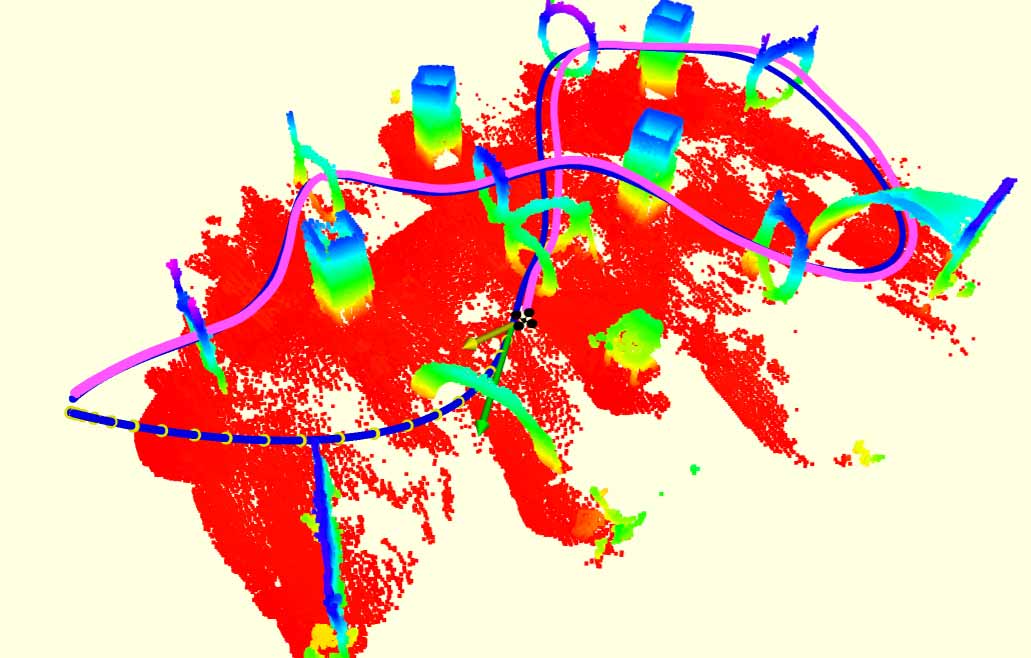}}   
\caption{
An overview of the teaching trajectory, flight corridor, and repeating trajectory in a pre-built dense map.
The colored code indicates the height of obstacles. The flight corridor is represented by transparent blue polyhedrons in (a). The global trajectory, local trajectory, quadrotor flight path are shown in blue, green, and purple curves, respectively.
\label{fig:indoor_repeat_rviz}}
\vspace{-0.25cm}
\end{figure}

Firstly, we conduct experiments in a cluttered drone racing scenario.
This experiment validates the robustness of our proposed system, and also pushes the boundary of aggressive flight of quadrotors. 
Several different types of obstacles, including circles, arches, and tunnels, are deployed randomly to composite a complex environment, as shown in Fig.~\ref{fig:indoor_map}. 
The smallest circle only has a diameter of $0.6 m$, which is very narrow compared to the $0.3m \times 0.3m$ tip-to-tip size of our drone.
The maximum velocity and acceleration of the drone are set as $3m/s$ and $3m/s^2$, respectively. 
And the parameter $\rho$ in Equ.~\ref{eq:objective} is set as 0, which means the quadrotor is expected to fly as fast as possible as long as it respects the kinodynamic limits. 
A dense global consistent map is pre-built using the method stated in Sect.~\ref{subsec:localization_mapping}. 
During the teaching phase, the quadrotor is virtually piloted by a human to move amid obstacles. 
Then the quadrotor autonomously converts this teaching trajectory to a global repeating trajectory and starts to track it. 
Snapshots of the drone in the flight are shown in Fig.~\ref{fig:indoor_exp}.
The teaching trajectory and the convex safe flight corridor are visualized in Fig~\ref{fig:indoor_teach_rviz_2}. 
And the global repeating trajectory is in Fig.~\ref{fig:indoor_repeat_rviz_2}. 

\subsubsection{Local Re-planning Against Unknown Obstacles }
\label{subsec:indoor_dynamic_exp}
\begin{figure}[t]
\centering
  \subfigure[\label{fig:indoor_dyn_exp_1} Side-view.]
{\includegraphics[height=0.33\columnwidth]{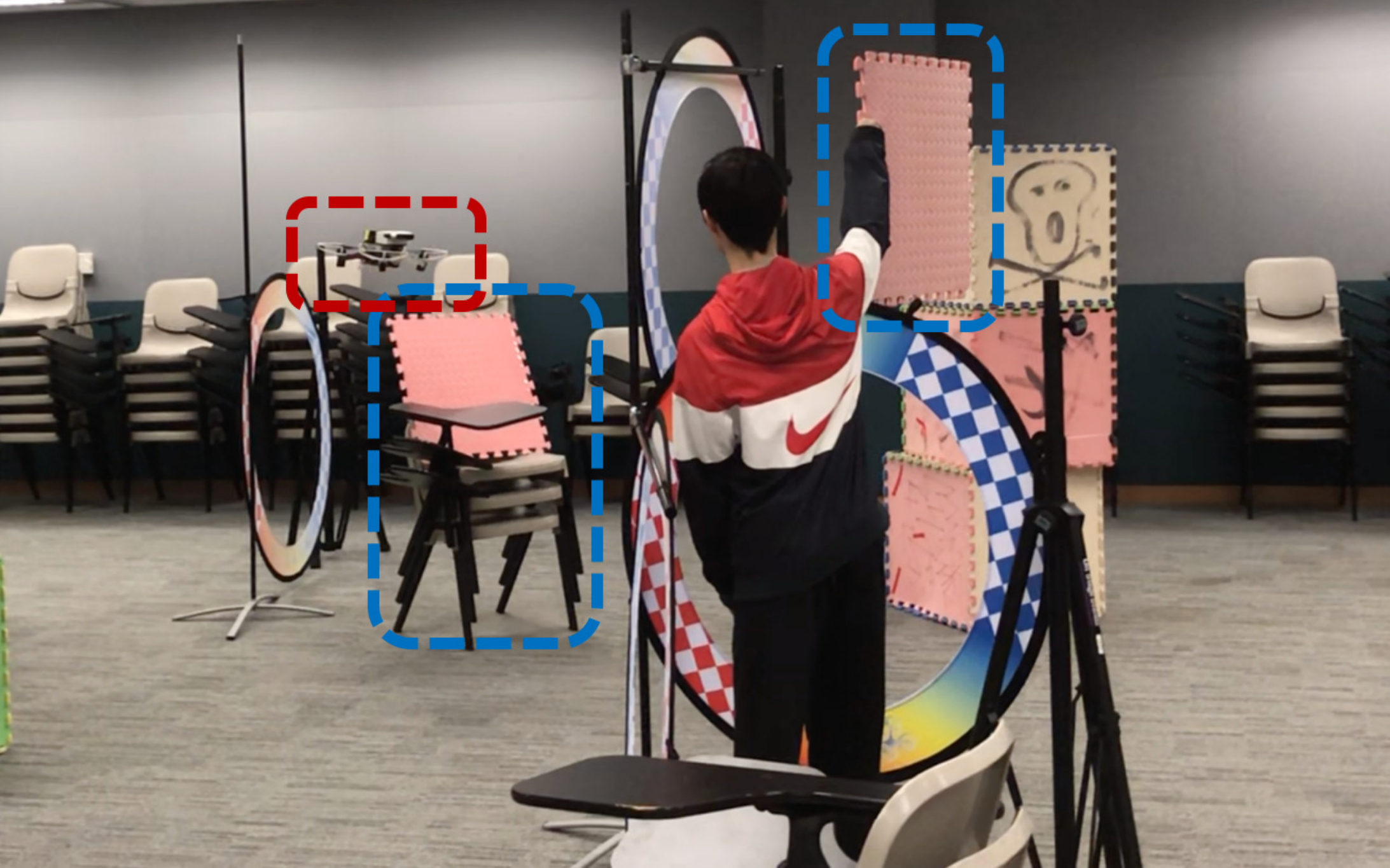}}
  \subfigure[\label{fig:indoor_dyn_exp_2} Onboard first-person view. ]
{\includegraphics[height=0.33\columnwidth]{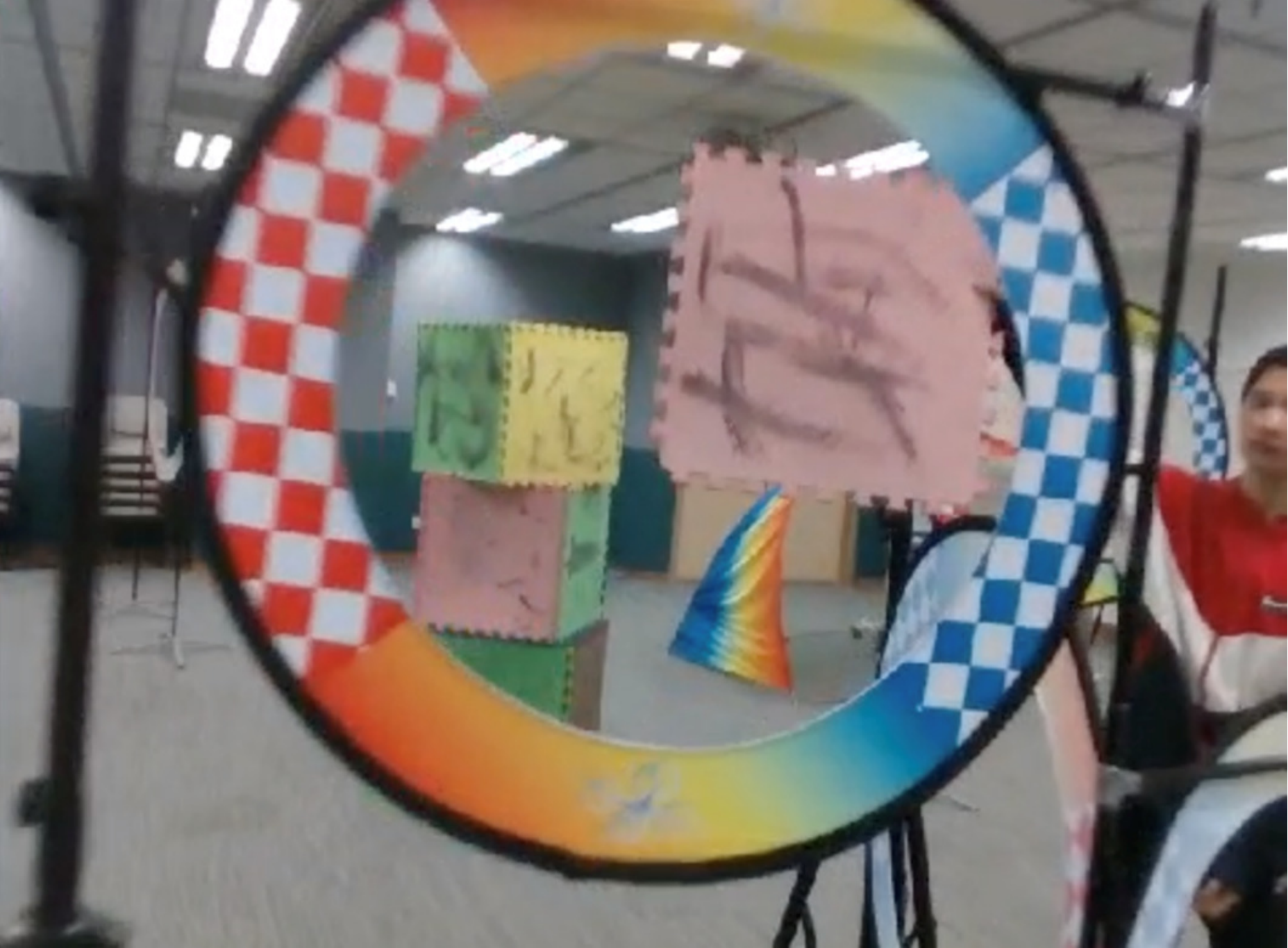}}
\caption{
The local re-planning experiment against unmapped and moving obstacles. The drone and unmapped obstacles are labeled by the red and blue dashed rectangles, respectively, for clear visualization.
\label{fig:indoor_dyn_exp}}
\vspace{-1.75cm}
\end{figure}

\begin{figure}[t]
\begin{center}           
\subfigure[\label{fig:indoor_dyn_repeat_rviz_1} Re-planning, unmapped obstacle.]
{\includegraphics[width=0.49\columnwidth]{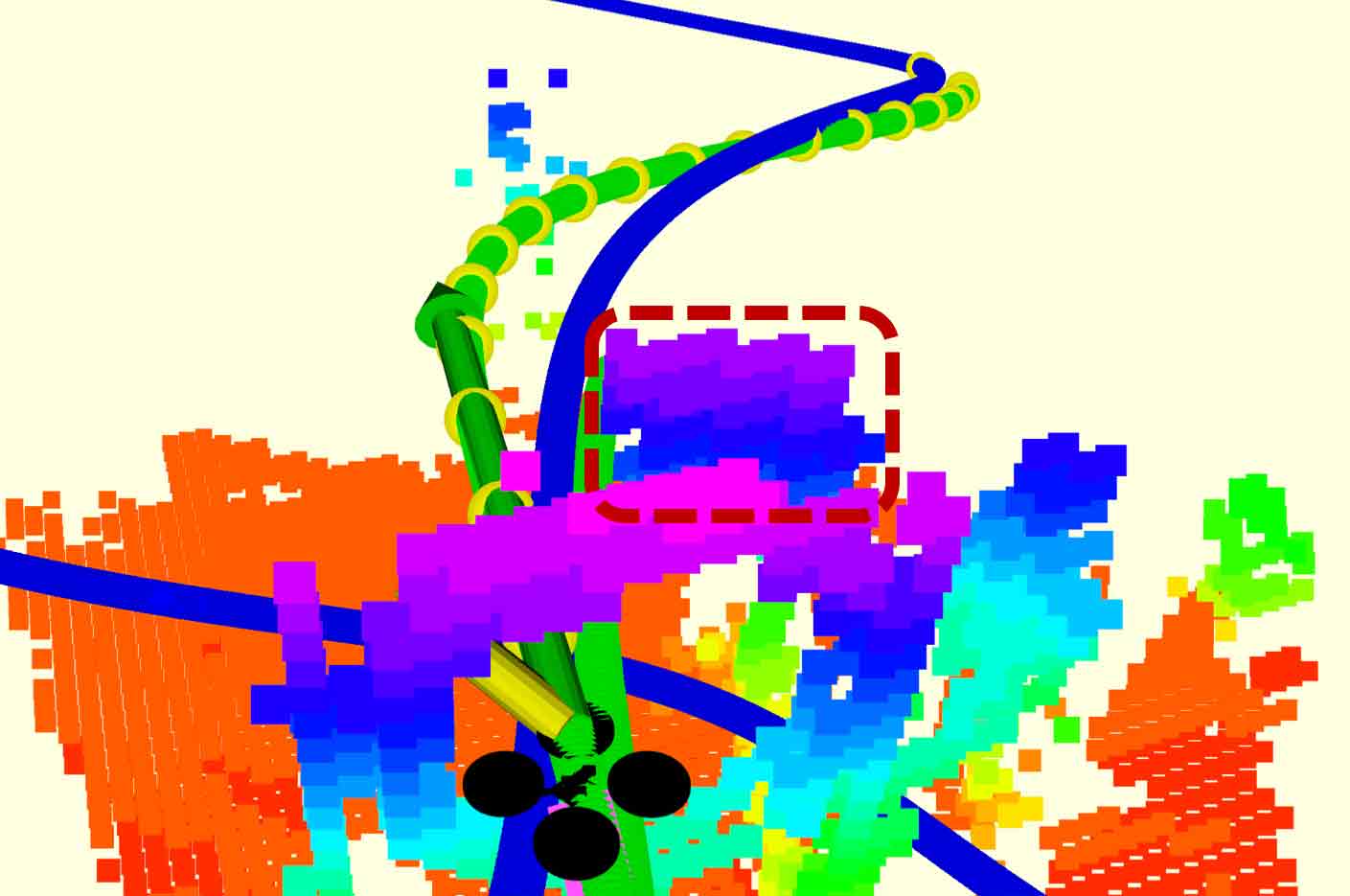}}     
\subfigure[\label{fig:indoor_dyn_repeat_rviz_2} Re-planning, moving obstacle.]
{\includegraphics[width=0.49\columnwidth]{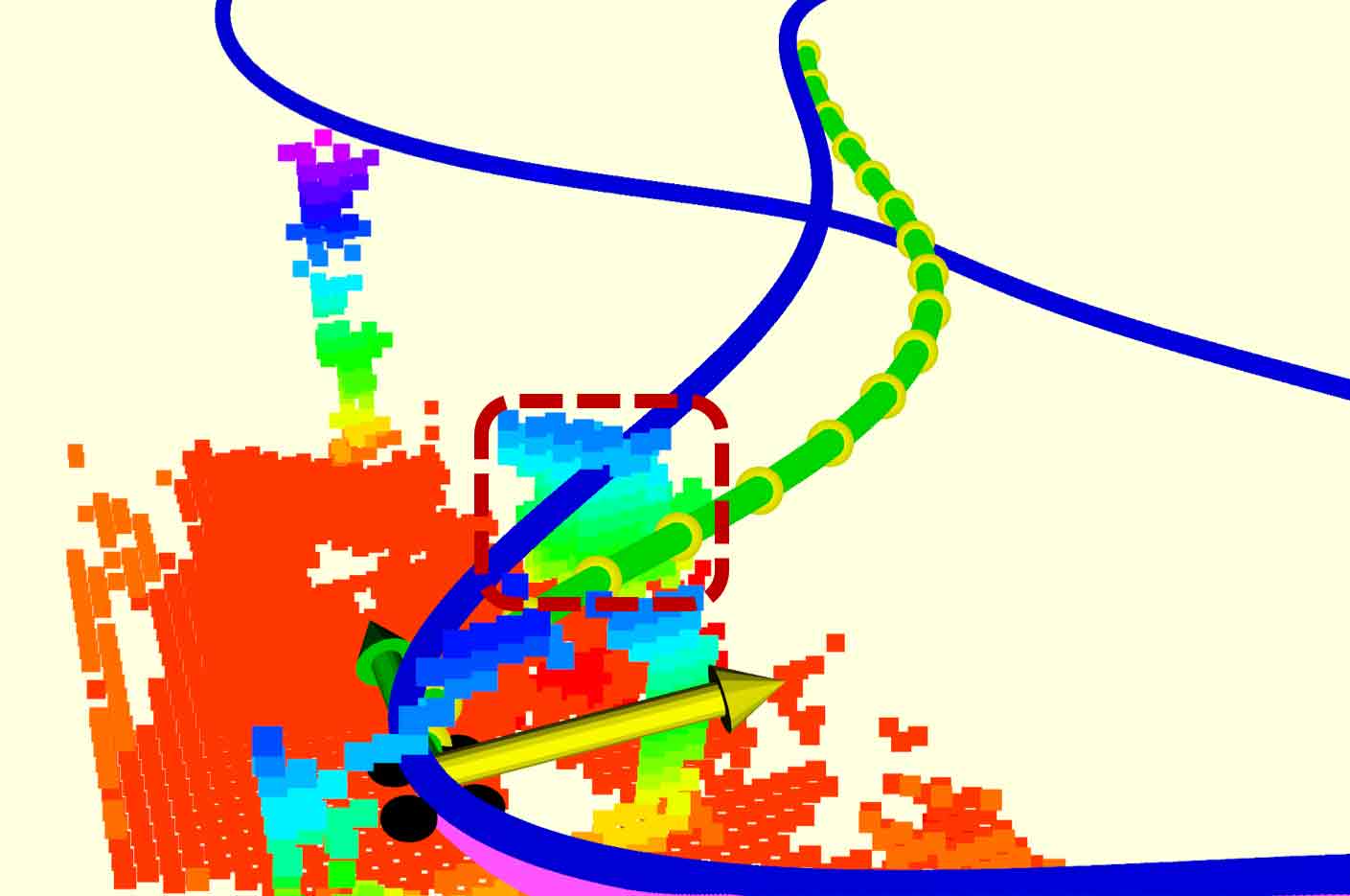}}     
\subfigure[\label{fig:indoor_dyn_repeat_rviz_all} Overview of all re-planning trajectories.]
{\includegraphics[width=0.99\columnwidth]{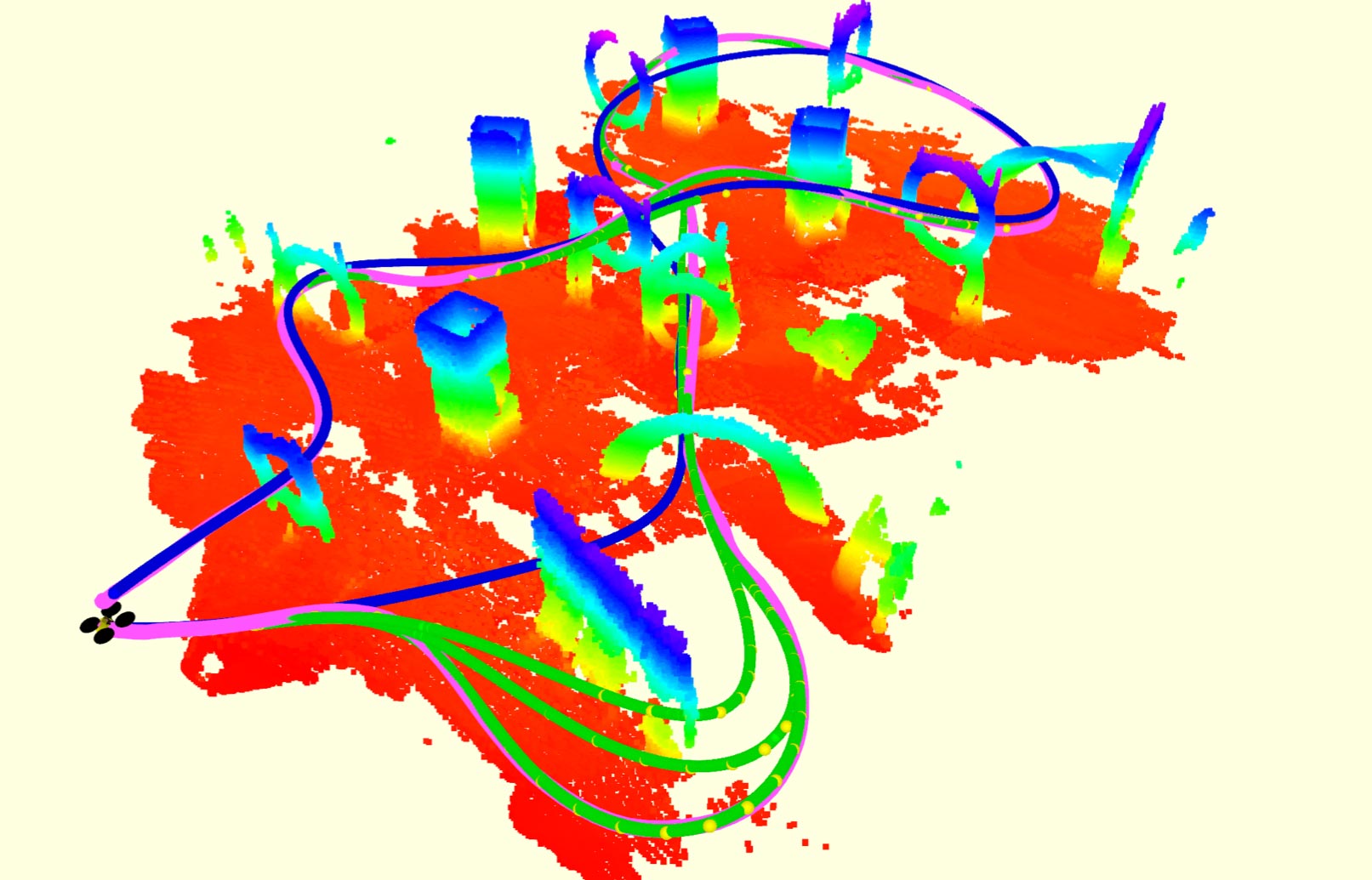}}     
\end{center}
\caption{\label{fig:indoor_dyn_repeat_rviz}
Indoor flight in a dynamic environment. 
In (a) and (b), the unmapped new obstacle and moving obstacles are labeled by red dashed rectangles, and colored voxels represent local obstacle maps. 
In (c), colored voxels show the global map.
Other marks are interpreted as the same as in previous figures.
}
\vspace{-0.5cm}
\end{figure} 

Our system can deal with changing environments and moving obstacles. 
In this experiment, we test our system also in the drone racing site to validate our local re-planning module. 
Several obstacles are moved or added to change the drone racing environment significantly, and some others are dynamically added during the repeating flight, as shown in Fig.~\ref{fig:indoor_dyn_exp}
In this experiment, the maximum velocity and acceleration for the quadrotor are set as $2m/s$ and $2m/s^2$. 
The local ESDF map is sliding with the drone using a ring-buffered updating mechanism~\cite{usenko2017real}. 
The resolution of the local perception is $0.075 m$.
The size of the map is decided by points observed spreading in the current frame.
The horizon and frequency of the local re-planning are $3.5 s$ and $15 Hz$, respectively. 
Re-planning is triggered 8 times during the flight in this experiment, and local safe and dynamical feasible splines are generated on time accordingly.
Local trajectories, local maps, and the overview of this experiment are shown in Fig.~\ref{fig:indoor_dyn_repeat_rviz}. 
We refer readers to the attached video for more details.

\begin{figure}[t]
\centering
  \subfigure[\label{fig:outdoor_exp_2} Outdoor experimet, trial 1. ]
{\includegraphics[width=0.49\columnwidth]{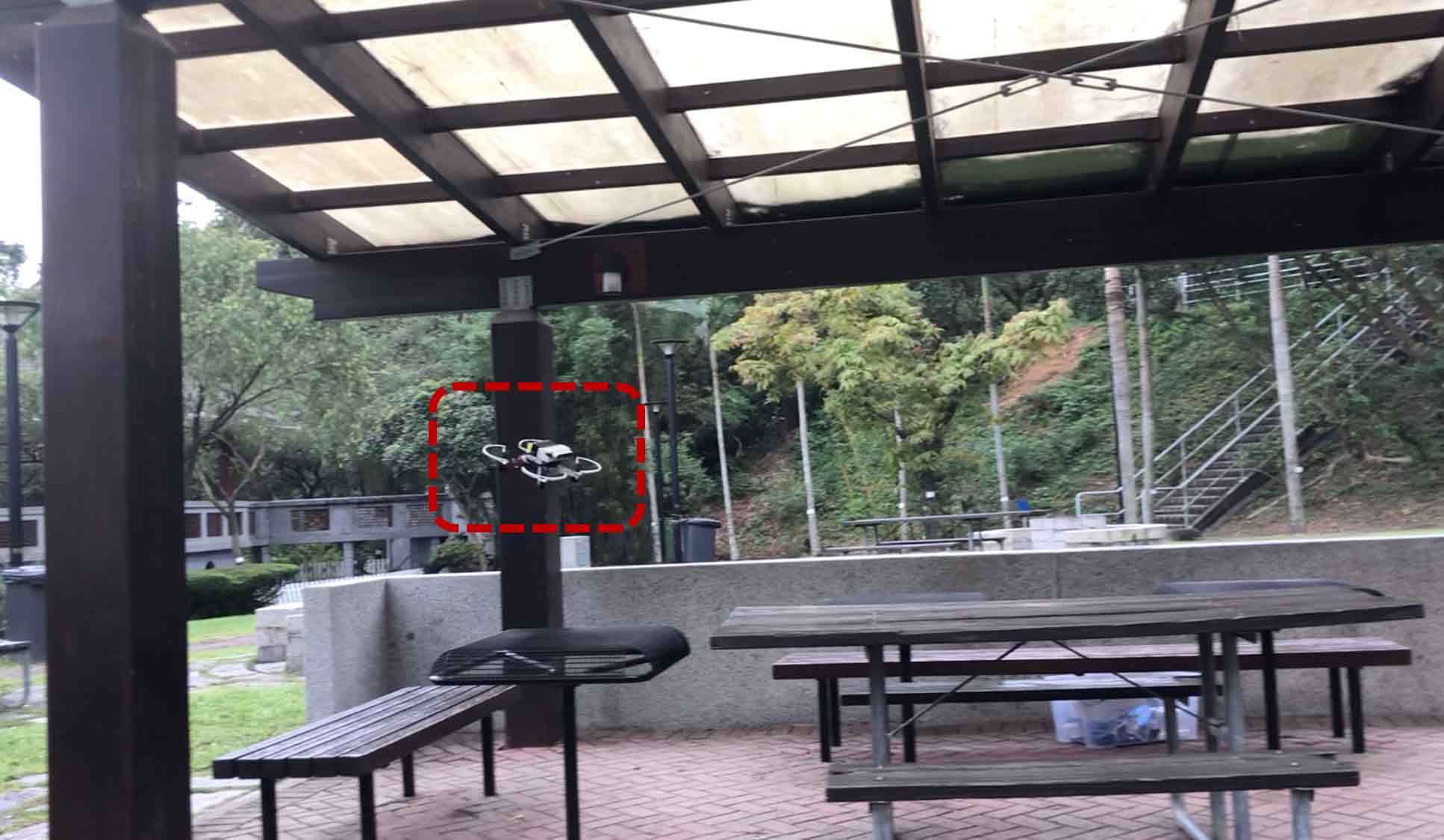}}
  \subfigure[\label{fig:outdoor_exp_3} Outdoor experimet, trial 2. ]
{\includegraphics[width=0.49\columnwidth]{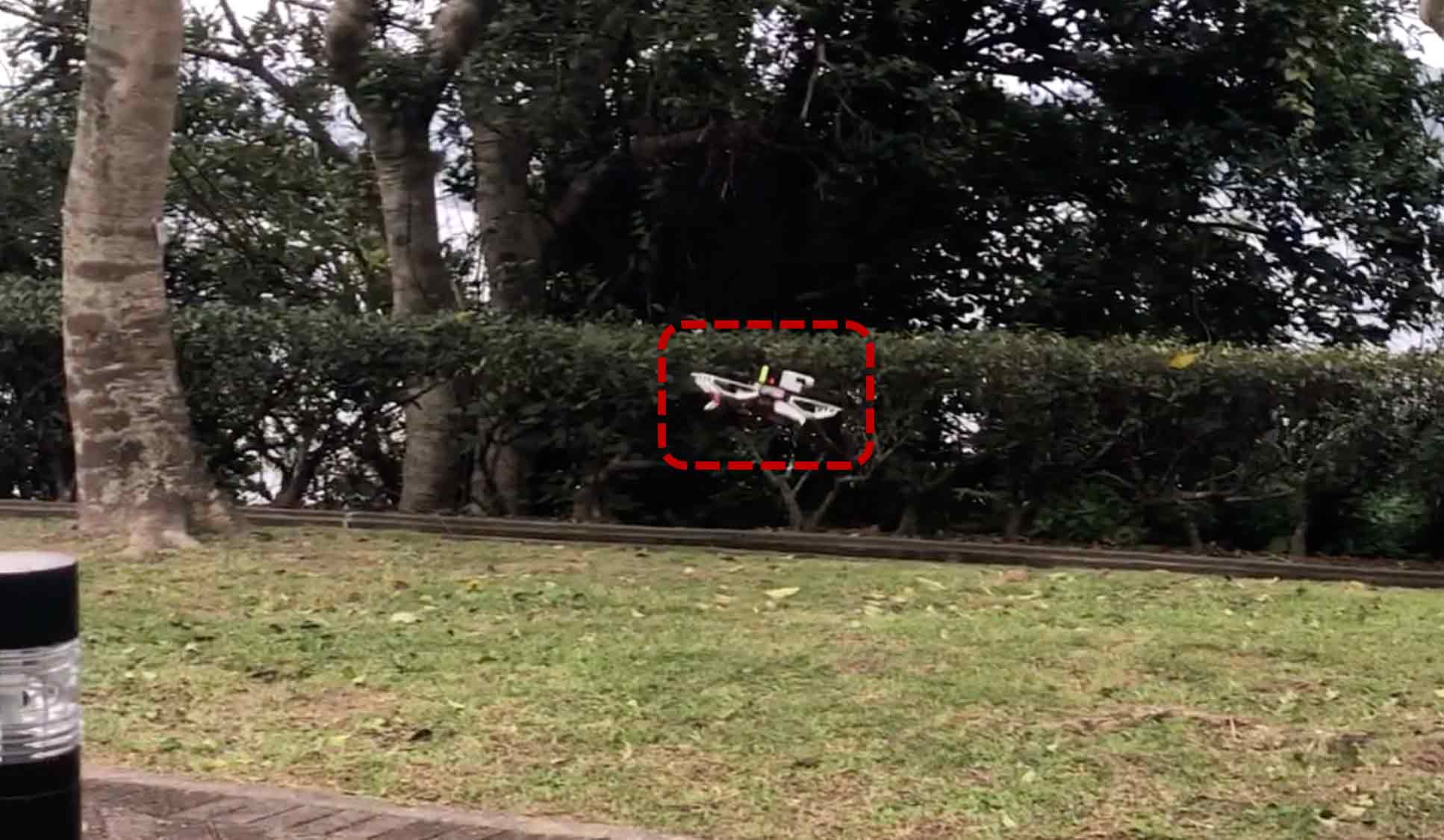}}
\caption{
Snapshots of the experiments in outdoor environments.
\label{fig:outdoor_exp}}
\vspace{-1.75cm}
\end{figure}

\begin{figure}[t]
\centering
\includegraphics[width=0.99\columnwidth]{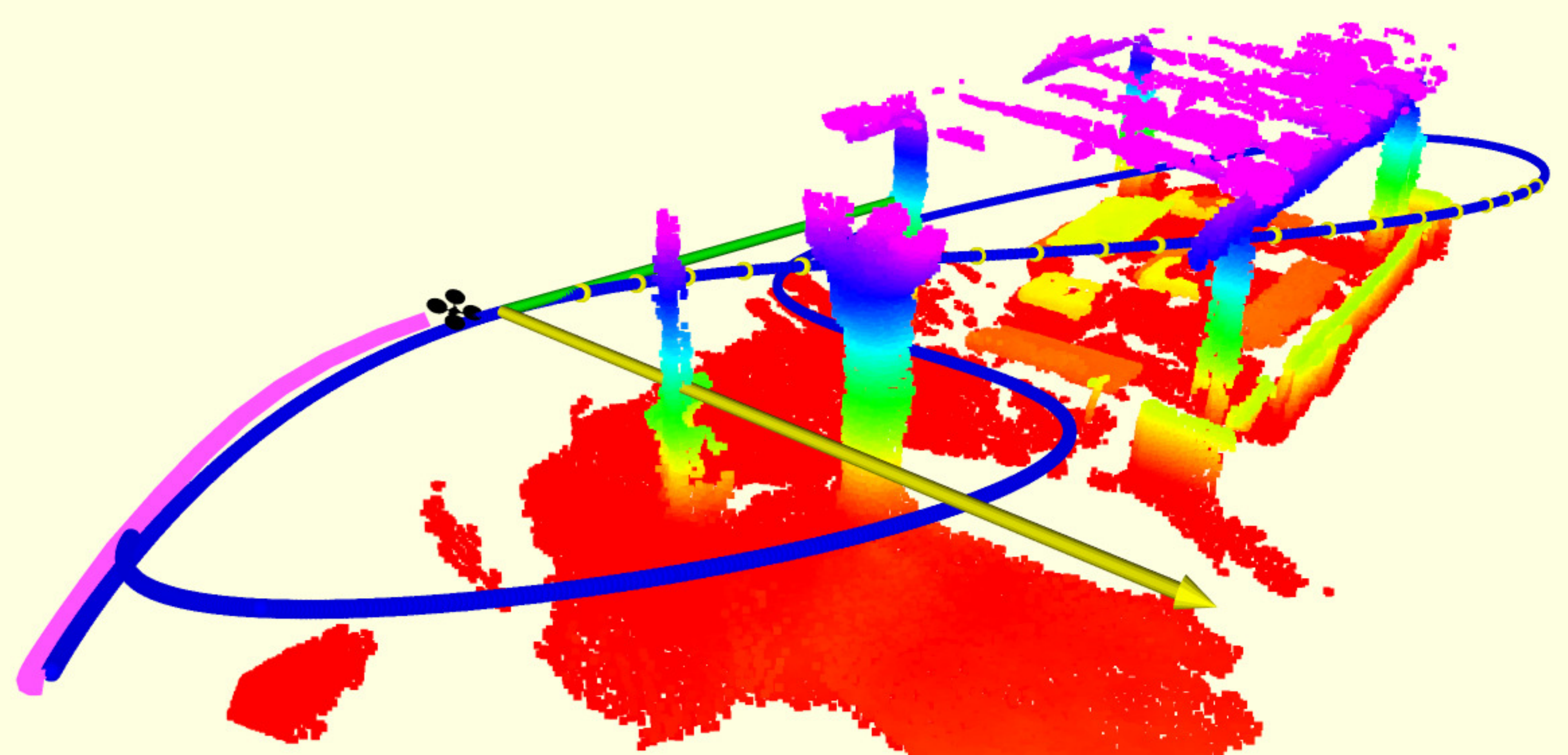}
\caption{
The repeating trajectory in outdoor experiments, trial 1. 
Marks are interpreted as the same as in previous figures.
\label{fig:rviz_outdoor_1}}
\vspace{-0.25cm}
\end{figure}

\begin{figure}[t]
\begin{center}           
\subfigure[\label{fig:rviz_outdoor_2_2} A close-up of outdoor experiments, trial 2.]
{\includegraphics[width=0.99\columnwidth]{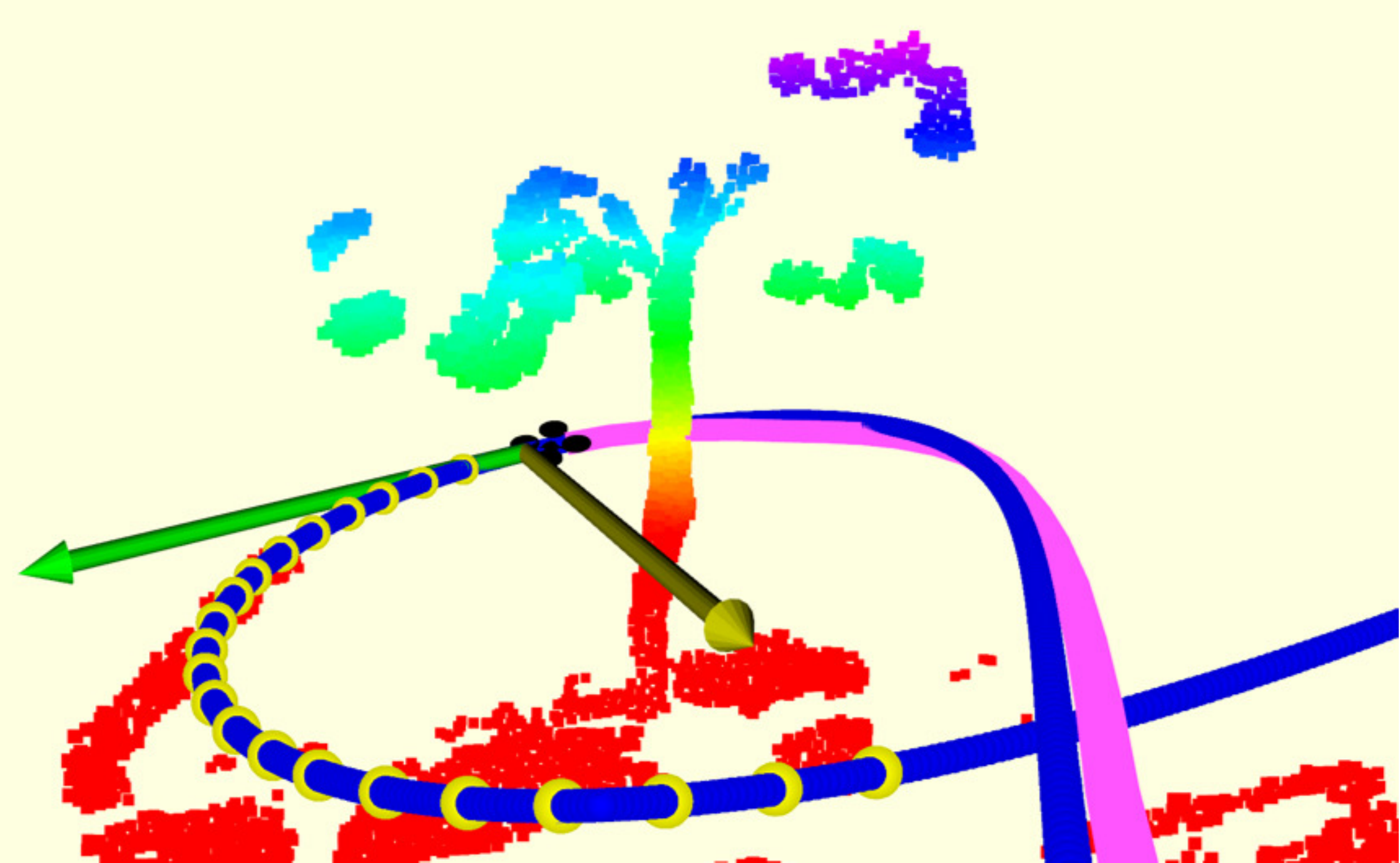}}     
\subfigure[\label{fig:rviz_outdoor_2_1} An overview of outdoor experiments, trial 2.]
{\includegraphics[width=0.99\columnwidth]{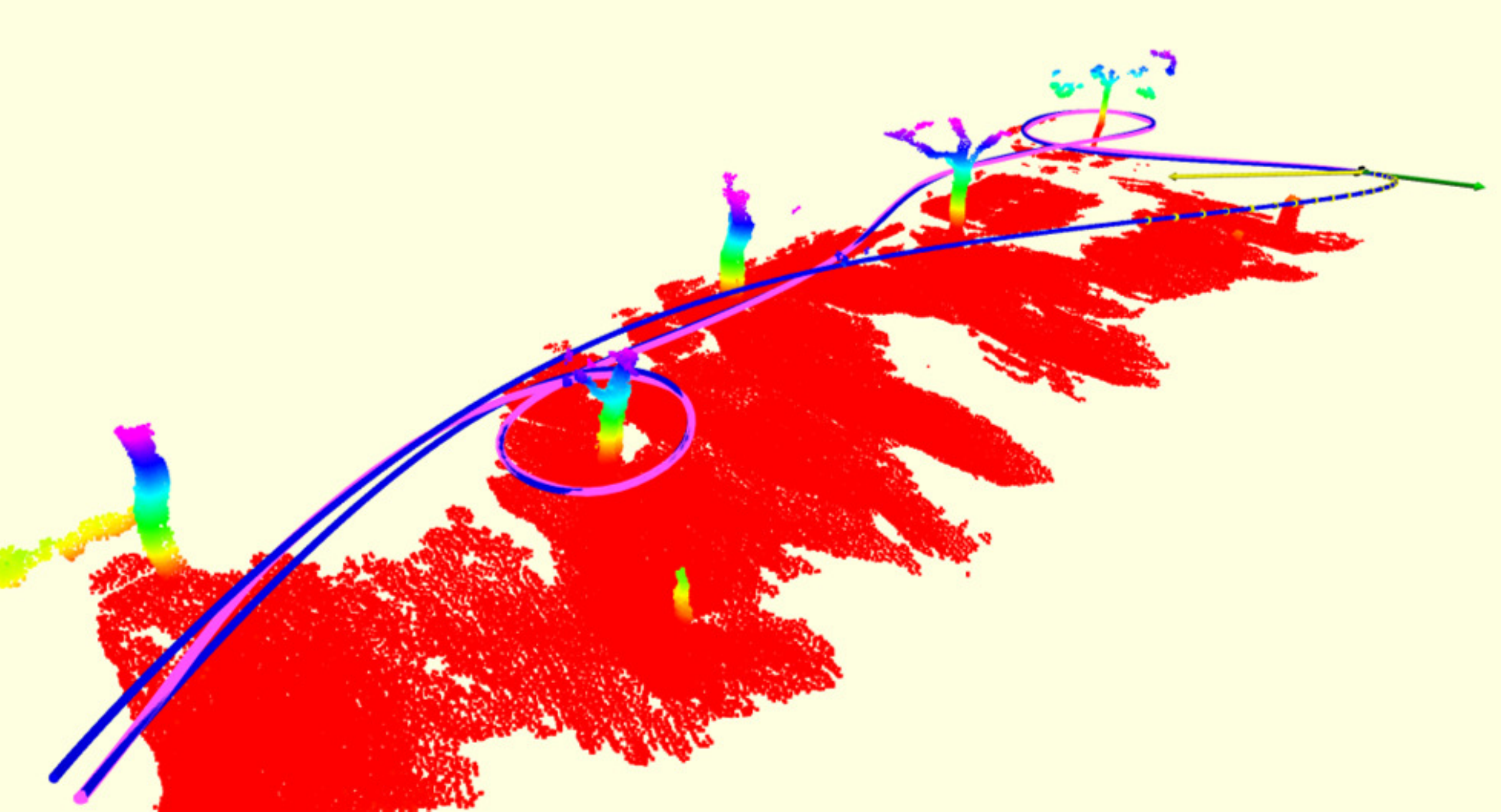}}     
\end{center}
\vspace{-0.25cm}
\caption{\label{fig:rviz_outdoor_2}
Outdoor flight, trial 2. 
Marks are interpreted as the same as in previous figures.
\vspace{-3.25cm}
}
\end{figure} 

\subsection{Outdoor Flight Test}
\label{subsec:outdoor_exp}

\begin{figure}[t]
\centering
\includegraphics[width=0.9\columnwidth]{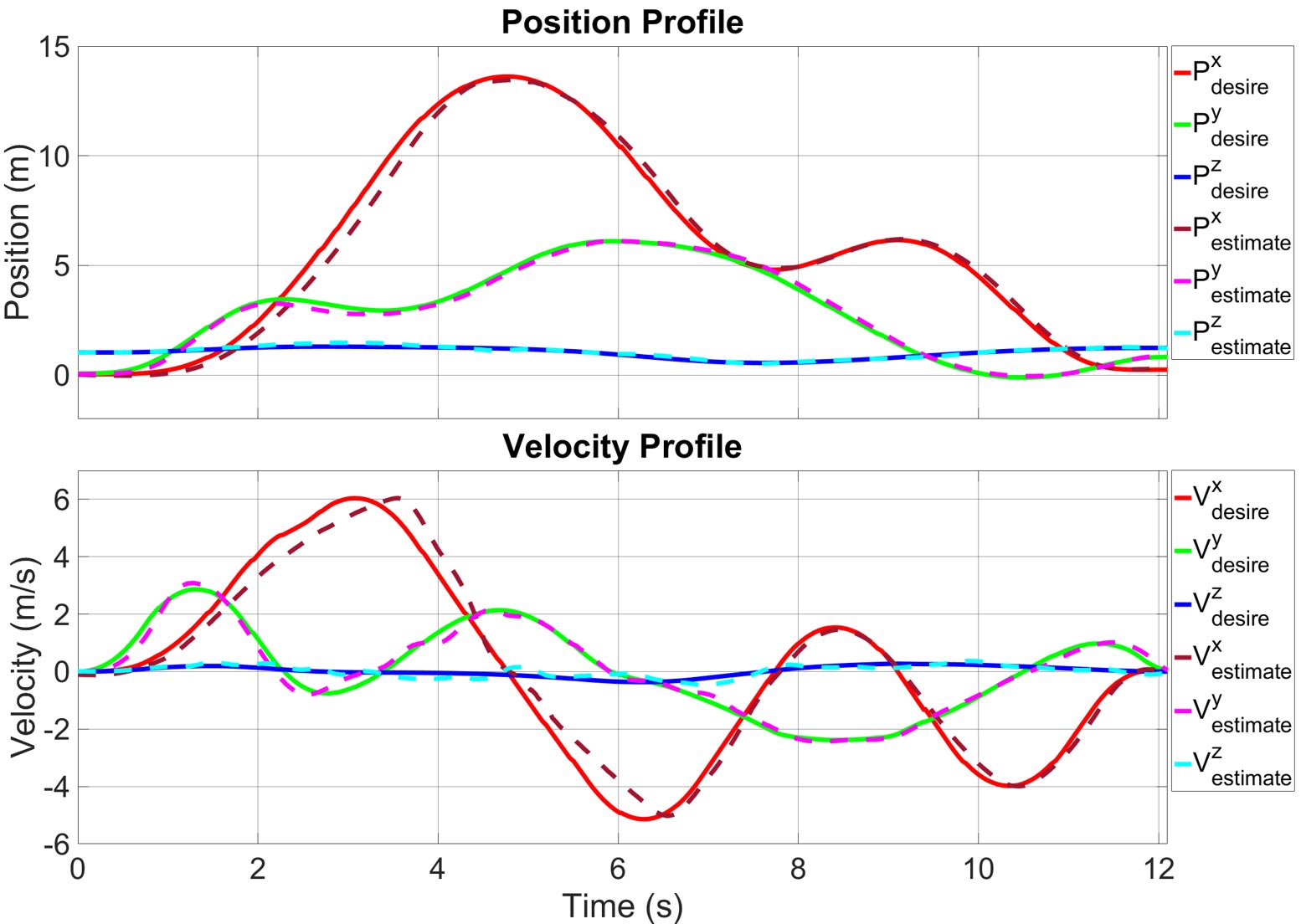}
\vspace{-0.25cm}
\caption{ 
Profiles of the desired and estimated position and velocity. The position and velocity are estimated by our localization module VINS~\cite{qin2018vins}.  
\label{fig:outdoor_pv_plot}
}
\vspace{-1.0cm}
\end{figure}

Finally, we conduct quadrotor flight experiments with a much higher aggressiveness in two different outdoor scenes, as in Fig.~\ref{fig:outdoor_exp}, to show the robustness of our system in natural environments. 
Although these experiments are conducted outdoor, GPS or other external positioning devices are not used. 
The teach-repeat-replan pipeline is as the same as before indoor experiments~\ref{subsec:indoor_exp}. 
The maximum allowed velocity and acceleration limits for these two trials are set as $5m/s$, $6m/s^2$ and $7m/s$, $6m/s^2$, respectively.
The drone's desired and estimated positions and velocities in the second trial are given in Fig.~\ref{fig:outdoor_pv_plot}, which shows acceptable tracking errors.
Since the flight speed is significantly higher than indoor experiments, we set a smaller re-planning horizon as $2.0 s$.
Results such as the global and local trajectory and the global map are visualized in Figs.~\ref{fig:rviz_outdoor_1} and~\ref{fig:rviz_outdoor_2}. 
More clearly visualizations of outdoor experiments are given in the video.

\section{Conclusion}
\label{sec:conclusion}
In this paper, we propose a framework, \textit{teach-repeat-replan} for quadrotor aggressive flights in complex environments. 
The main idea of this work is to find the topological equivalent free space of the user's teaching trajectory, use spatial-temporal trajectory optimization to obtain an energy-efficient repeating trajectory, and incorporate online perception and re-planning to ensure the safety against environmental changes and moving obstacles.
The teaching process is conducted by virtually controlling the drone in simulation.
The generated repeating trajectory captures users' intention and respect an expected flight aggressiveness, which enables autonomous flights much more aggressive than human's piloting in complex environments.
The online re-planning guarantees the safety of the flight and also respects the reference of the repeating trajectory.

To group large free space around the teaching trajectory, we propose a GPU-accelerated convex polyhedron clustering method to find a flight corridor.
The optimal global trajectory generation problem is decoupled as spatial and temporal sub-problems.
Then these two sub-problems are iteratively optimized under the coordinate descent framework. 
Moreover, we incorporate the local perception and local trajectory re-planning modules into our framework to deal with environmental changes, dynamic obstacles, and localization drifts.

The proposed system is complete and robust. 
Users of our system do not have to pilot the drone carefully to give a teaching trajectory. 
Instead, an arbitrarily jerky/poor trajectory can be converted to an efficient and safe global trajectory.
Moreover, when the environment changes or the global localization drifts, the local perception and re-planning modules guarantee the safety of the drone while tracking the global trajectory.
Our system is also flexible and easily replicable, as evidenced by various types of experiments presented in this paper, and a third-party application\footnote{Flight demonstration at the Electrical and Mechanical Services Department (EMSD), Hong Kong government. Video: \url{https://youtu.be/Ut8WT0BURrM}}.
We release all components of our system for the reference of the community.

\newlength{\bibitemsep}\setlength{\bibitemsep}{.03\baselineskip}
\newlength{\bibparskip}\setlength{\bibparskip}{0pt}
\let\oldthebibliography\thebibliography
\renewcommand\thebibliography[1]{%
  \oldthebibliography{#1}%
  \setlength{\parskip}{\bibitemsep}%
  \setlength{\itemsep}{\bibparskip}%
}
\bibliography{tro2019fei} 
\end{document}